\documentclass[11pt]{article}

\usepackage[preprint]{acl}

\usepackage{times}
\usepackage{latexsym}

\usepackage[T1]{fontenc}
\usepackage[utf8]{inputenc}

\usepackage{microtype}

\usepackage{inconsolata}

\usepackage{graphicx}

\usepackage{booktabs}
\usepackage{subcaption}
\usepackage{subfloat}
\usepackage{enumitem}

\usepackage{algorithm}
\usepackage{algpseudocode}
\usepackage[table,xcdraw]{xcolor}
\usepackage[normalem]{ulem}

\usepackage{url}
\usepackage{hyperref}

\useunder{\uline}{\ul}{}

\usepackage{amsmath,amsfonts,bm}

\def\eqref#1{equation~\ref{#1}}
\def\1{\bm{1}}

\DeclareMathAlphabet{\mathsfit}{\encodingdefault}{\sfdefault}{m}{sl}
\SetMathAlphabet{\mathsfit}{bold}{\encodingdefault}{\sfdefault}{bx}{n}

\title{LoRA on the Go: Instance-level Dynamic LoRA Selection and Merging}

\author{
 \textbf{Seungeon Lee\textsuperscript{1}},
 \textbf{Soumi Das\textsuperscript{1}},
 \textbf{Manish Gupta\textsuperscript{2}},
 \textbf{Krishna P. Gummadi\textsuperscript{1}}
\\
\\
 \textsuperscript{1}MPI-SWS\quad
 \textsuperscript{2}Microsoft, India
\\
 \small{
   \textbf{Correspondence:} \href{mailto:email@domain}{selee@mpi-sws.org}
 }
}

\newcommand{\model}{\textsc{LoGo}}

\begin{document}
\maketitle

\begin{abstract}
Low-Rank Adaptation (LoRA) has emerged as a parameter-efficient approach for fine-tuning large language models.
However, conventional LoRA adapters are typically trained for a single task, limiting their applicability in real-world settings, where inputs may span multiple, diverse task domains. 
At inference time, existing methods can combine multiple LoRAs to improve cross-task performance, but they require additional labeled data or task-specific training, which is expensive at scale.

In this work, we introduce LoRA on the Go (\model{}), a training-free framework that dynamically selects and merges adapters at the instance level without any additional requirements. \model{} leverages signals extracted from a single forward pass through LoRA adapters, to identify the most relevant adapters and determine their contributions on-the-fly.
Across 5 NLP benchmarks, 27 datasets, and 3 model families, \model{}  outperforms training-based baselines on some tasks up to a margin of $3.6\%$ while remaining competitive on other tasks and maintaining inference throughput, highlighting its effectiveness and practicality.

\end{abstract}
\section{Introduction}

\begin{figure}
\centering
\includegraphics[width=0.9\columnwidth]{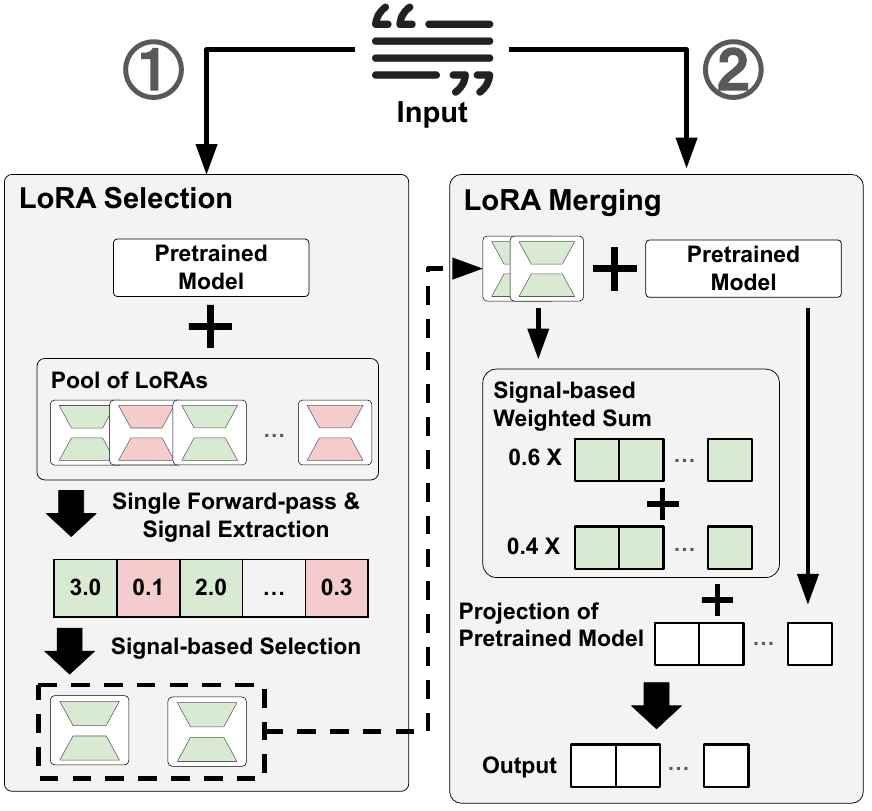}
\caption{Overall workflow of the proposed LoRA on the Go (\model{}) framework.}
\label{fig:framework}
\end{figure}

Recent advances in large language models (LLMs) such as LLaMA~\citep{dubey2024llama3} and DeepSeek~\citep{bi2024deepseek} have led to remarkable progress across diverse natural language processing (NLP) tasks.
While these models demonstrate strong generalization capabilities, achieving state-of-the-art results in specialized domains often requires task-specific fine-tuning~\citep{wei2022finetuned}.
However, the massive scale of modern LLMs makes full fine-tuning computationally prohibitive, motivating research on Parameter-Efficient Fine-Tuning (PEFT) methods that adapt models by updating only a small subset of parameters~\citep{houlsby2019parameter, li2021prefix, liu2022few}.
Among them, Low-Rank Adaptation (LoRA)~\citep{hu2022lora} is particularly effective, introducing trainable low-rank matrices while freezing pretrained weights, thus reducing trainable parameters without sacrificing performance.

Although LoRA provides an efficient adaptation mechanism, the adaptors are typically optimized for a single task domain.
In contrast, real-world applications increasingly demand generalization to unseen tasks or tasks that require specialization across multiple task domains.
Recent works~\citep{huang2024lorahub, zhao2024loraretriever} explore the possibility of simultaneously leveraging multiple LoRAs trained on diverse tasks.  

Existing multi-LoRA approaches share a key limitation: when composing LoRAs together, they need to first select relevant LoRAs from a pool and then they need to merge them. For both steps, they assume well-defined mixed-tasks and rely on labeled data.
For instance, LoRAHub~\citep{huang2024lorahub} learns fixed composition weights for each new mixed-task using labeled samples from the target input distribution, while LoRARetriever~\citep{zhao2024loraretriever} trains a retrieval model to select relevant LoRAs, but the retrieval model still depends on labeled examples to compute retrieval embeddings.
Such dependence on task homogeneity and labeled supervision restricts scalability in real-world scenarios, where LoRAs may be continually added or updated and labeled data may be unavailable.

In generic conversational systems such as AI copilots~\citep{microsoft_copilot} or multi-domain assistants~\citep{openai_chatgpt, google_gemini}, these assumptions rarely hold.
User queries are highly heterogeneous, often privacy-sensitive, and may transition across unrelated tasks (e.g., summarization, translation, coding) without explicit task boundaries.
Meanwhile, LoRA pools evolve dynamically as new adapters are introduced or deprecated, making task-specific retraining or labeled data collection expensive and impractical.
\noindent
These challenges motivate our central research question:
\textbf{\textit{How can we dynamically select suitable LoRAs for each input, given an evolving LoRA pool and heterogeneous tasks, without labeled data or retraining?}}

In this work, we introduce \textbf{Lo}RA on the \textbf{Go} (\textbf{\model{}}), a framework that operates without any pre-defined data or retraining assumptions, enabling seamless integration with a dynamic LoRA pool.
Fig.~\ref{fig:framework} illustrates the workflow of \model{}.
\model{} adopts an \textbf{instance-specific} perspective-selecting and merging LoRAs on the fly for each input.
Since selection and merging must occur over many candidates in real time, our method is entirely training-free.
The core intuition is that \textit{LoRA activations already encode signals of relevance}: when a LoRA is well-suited to an input, its updates exert stronger influence on model outputs (e.g., inference for WNLI~\citep{levesque2012wnli} benefits from LoRAs trained on SNLI~\citep{bowman2015snli} and MNLI~\citep{williams2018mnli}).

Building on this, \model{} extracts simple yet informative signals—such as the norm or entropy of LoRA activations—from a single forward pass with all LoRAs attached.
These signals are used to identify relevant adapters, which are then merged via a weighted sum of activations, where weights are determined by the extracted signals.
% (see Fig.~\ref{fig:framework}).
%

We evaluate \model{} on a diverse set of NLP benchmarks spanning 27 datasets, including BIG-Bench Hard (BBH)~\citep{suzgun2023challenging}, Translation~\citep{bojar2014findings, bojar2016findings}, Struct-to-Text~\citep{gehrmann2021gem, lin2020commongen, nan2021dart, novikova2017e2e, gardent2017creating}, Closed-Book QA~\citep{clark2018think, kwiatkowski2019natural, joshi2017triviaqa}, and Natural Language Inference~\citep{nie2020adversarial, wang2018glue}.
In our experiments, we train LoRAs for three model families over 260 FLANv2 tasks~\citep{wei2022finetuned, chung2024scaling}.

Results show that \model{}, even without retraining or data assumptions, often surpasses training-based baselines by up to $3.6\%$ on tasks like Struct-to-Text, NLI, while maintaining competitive performance on the rest.
\model{} also preserves comparable throughput during selection, merging, and inference.
Our analysis confirms that its overhead is amortized in long-output tasks such as summarization or chain-of-thought reasoning, making it highly practical.
Across different settings, \model{} shows consistent  performance.
%consistently identifies relevant LoRAs and achieves strong, robust performance.
We release our code\footnote{\url{https://github.com/archon159/LoGo}} and trained LoRAs\footnote{\url{https://huggingface.co/archon159/LoGo-loras-collection}} publicly.

\smallskip\noindent
Our main contributions are summarized as follows:
\begin{itemize}[leftmargin=*, itemsep=0pt, topsep=0pt]
    \item We identify the limitations of existing multiple LoRA-based approaches, which rely on labelled data availability and additional training, making them expensive for real-world deployment.
    \item We introduce LoRA on the Go (\model{}), a training-free, instance-specific framework that dynamically selects and merges suitable LoRAs for each input using activations extracted in a single forward pass.
    \item We conduct extensive experiments on 5 standard benchmarks encompassing 27 datasets over 3 model families, showing that \model{} not only outperforms training-based baselines but also has comparable throughput.
\end{itemize}

\section{Related Work}
Several studies have explored dynamically combining multiple LoRA adapters, each trained on different tasks, to handle new inputs.
Mixture of LoRAs (MoA) trains a router to select a single LoRA~\citep{feng2024mixture}, while LoRAHub merges multiple LoRAs via learning task-specific weights for parameter summation~\citep{huang2024lorahub}.
Mixture of LoRA Experts (MoLE) similarly learns weights, but applies them to adapter outputs rather than parameters~\citep{wu2024mixture}.
All of these methods assume access to labeled samples from the target input distribution and rely on such data to train either the router or the merging weights in a task-specific manner.  
However, this assumption rarely holds in practice: inputs usually arrive from diverse and unpredictable domains.

LoRARetriever~\citep{zhao2024loraretriever} moves toward instance-specific adaptation by retrieving relevant LoRAs using an auxiliary embedding model trained on mixed datasets.
However, it requires training a large embedding model and maintaining dataset samples.
Extending this framework to new LoRAs requires (a) samples from the corresponding datasets, and (b) re-computing an embedding point for the new LoRAs in the existing embedding space.
This embedding might not be appropriate for out-of-domain (OOD) scenarios, e.g. non-English tasks.
Additionally, it may also tamper with the in-domain performance due to entangled embedding points with the OOD LoRAs. 
Hence, the performance of the resulting model may decline when the inputs deviate significantly from the training distribution of the embedding model.

Recent contemporaneous works have also considered training-free or dynamic LoRA composition.
K-LoRA~\citep{ouyang2025k} fuses only two LoRAs with predefined roles, limiting scalability to large adapter pools.
Decouple and Orthogonalize~\citep{zheng2025decouple} is data-free but requires additional optimization and merges all LoRAs without selection.
LoRAtorio~\citep{foteinopoulou2025loratorio} computes patch-level similarity for diffusion models, incurring per-step overhead.
LoRA-Flow~\citep{wang2024lora} trains a router to compute per-token fusion weights, introducing training cost and scalability challenges.

In contrast, \model{} supports instance-specific LoRA selection without training or additional data.
By leveraging adapter activations, it identifies relevant LoRAs on the fly, eliminating reliance on auxiliary models or predefined samples.
\section{The Proposed \model{} Methodology}

The goal of \model{} is to dynamically select and merge the most relevant LoRA adapters for each input, without relying on task-specific training.
We begin by formalizing the problem setting, where a pretrained backbone LLM is equipped with a pool of LoRA adapters at the scale of hundreds, each providing low-rank updates to projection matrices (Section~\ref{sec:problem}).
Given a new input, \model{} performs a single forward pass with all adapters attached, extracts their projection outputs from a designated block, and computes signal scores (e.g., norms or inverse entropy) from these projection outputs to measure adapter relevance.  
The top-scoring adapters are then selected as candidates (Section~\ref{sec:lora_selection}). 
Next, the selected adapters are merged efficiently through a weighted sum of their outputs, where the weights are directly determined by the extracted signals (Section~\ref{sec:lora_merging}).
This design allows \model{} to adaptively combine multiple LoRAs on the fly, while maintaining real-time efficiency and avoiding any additional training overhead.
We provide an algorithm that summarizes overall procedure of \model{} in Appendix~\ref{sec:algo}.

\subsection{Problem Formulation}
\label{sec:problem}

We consider a setting where a pretrained model $f_\theta$ is paired with a set of $N$ LoRA adapters $\mathcal{L}=\{L_i\}_{i=1}^N$, each of which is fine-tuned on a distinct task $T_i$.
Given an input sequence $\mathbf{x}=(x_1,...,x_P)$ of length $P$, the model generates an output sequence $\mathbf{y}=(y_{P+1},...,y_{P+t})$ of length $t$.

The pretrained model $f_\theta$ consists of $M$ Transformer blocks $\mathcal{B}=\{B_j\}_{j=1}^M$, where each block $B_j$ contains  a self-attention mechanism with head-specific query, key, and value projections, and a feed-forward network. 
%\af{Aren't QKV part of attention? Repeating it twice seems weird}
%
We denote the query and value projection matrices of block $B_j$ (we omit head subscripts for sake of clarity) as $\mathbf{W}_j^{(Q)}$ and $\mathbf{W}_j^{(V)}$, respectively.
A LoRA adapter $L_i \in \mathcal{L}$ attaches to the projection matrices $\{(\mathbf{W}_j^{(Q)}, \mathbf{W}_j^{(V)})\}_{j=1}^M$ and introduces low-rank updates.
For example, for the query projection in block $B_j$, let $\mathbf{h}_j$ denote the latent input.
Then, the adapter $L_i$ produces an update via a low-rank projection $\Delta \mathbf{W}_{i,j}^{(Q)} \mathbf{h}_j$, where $\Delta \mathbf{W}_{i,j}^{(Q)} = \alpha_{i,j}\mathbf{A}_{i,j}\mathbf{B}_{i,j}$.
Here, $\mathbf{A}_{i,j}$ and $\mathbf{B}_{i,j}$ are the low-rank matrices of LoRA, and $\alpha_{i,j}$ is a scaling factor.

\subsection{Selection of Instance-specific LoRAs}
\label{sec:lora_selection}

\begin{figure}
    \centering
    \includegraphics[width=1.0\columnwidth,height=1.0\columnwidth,keepaspectratio]
{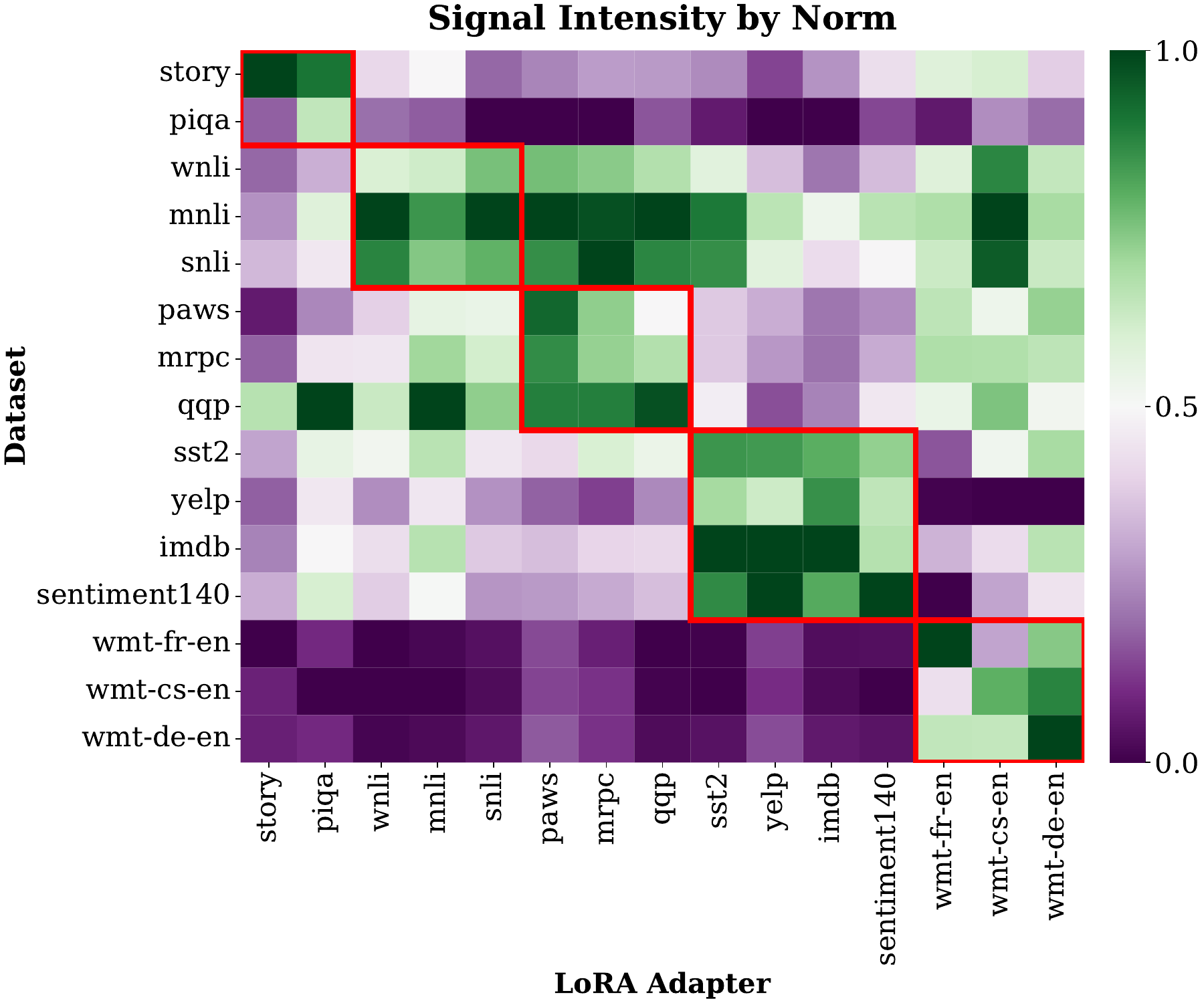}
    \caption{Heatmap illustrating signal patterns across LoRA adapters trained on top of the Qwen-2.5-7B backbone. The x-axis represents LoRAs trained on different tasks, while the y-axis corresponds to datasets from those tasks. Each cell shows the $\ell_2$ norm of the projection outputs. The norm values are min-max normalized to [0,1] across datasets for each LoRA. Related task clusters are highlighted in red boxes. More results on signal intensity are in Appendix~\ref{sec:full_signals}.}
    \label{fig:signal_intensity_norm}
\end{figure}

To identify the relevant LoRAs for a given input, \model{} relies on signals extracted during a single forward pass (probe pass) through the base model $f_\theta$ equipped with all available adapters in $\mathcal{L}$. 
Unlike prior approaches that require additional training, our method is training-free. The  procedure incurs only the cost of one token generation, making it practical even under real-time constraints. 

Formally, we define the \emph{adapter-augmented model} $f_{\theta,\mathcal{L}}$ as the base model $f_\theta$ where every adapter $L_i \in \mathcal{L}$ is attached to its designated projection matrices.
Let $B_T \in \mathcal{B}$ denote the target block from which we extract signals, and let $\mathbf{h}_T$ be the hidden representation entering block $B_T$.  
For each adapter $L_i \in \mathcal{L}$ attached to query projection matrix $\mathbf{W}_T^{(Q)}$ of $B_T$, we define projection output as
\begin{equation}
\small
\mathbf{o}_{i,T} = \Delta \mathbf{W}^{(Q)}_{i,T}\mathbf{h}_T,
\end{equation}
where $\Delta \mathbf{W}^{(Q)}_{i,T} = \alpha_{i,T}\mathbf{A}_{i,T}\mathbf{B}_{i,T}$ is the low-rank update from $L_i$ as defined in Section~\ref{sec:problem}.
From each projection, we compute a scalar \emph{signal score} $s_{i}$ that quantifies the relevance of adapter $L_i$ to the given input.
Typical examples include the $\ell_2$ norm,
\begin{equation}
\small
s_{i} = \|\mathbf{o}_{i,T}\|_2,
\end{equation}
or entropy-based measures,
\begin{equation}
\small
p_{i,T}^{(j)} = \frac{\exp(o_{i,T}^{(j)})}{\sum_{k}\exp(o_{i,T}^{(k)})}, 
s_i = \biggl({- \sum_{j} p_{i,T}^{(j)} \log p_{i,T}^{(j)}}\biggr)^{-1}
\end{equation}
derived from the projection distribution, both of which capture how strongly a LoRA adapter responds to the input.
Intuitively, a larger projection norm indicates stronger activation and thus greater influence on the model’s output, while lower-entropy projections imply more confident and focused responses.
Hence, these metrics serve as natural indicators of adapter relevance.
Here, we take the reciprocal of entropy so that scores are always positive and lower-entropy (i.e., more confident) adapters receive larger weights.
See Appendix~\ref{sec:signals_theory} for the theoretical interpretation of norm and entropy as signals.

To examine whether these projection-based signals indeed capture task relevance, Fig.~\ref{fig:signal_intensity_norm} demonstrates signal intensities across existing LoRA adapters (from the pool) and datasets.
Each column represents a LoRA trained on a specific dataset, while each row represents the dataset used for inference.
The heatmap values indicate the normalized signal (here, norm) when each LoRA is applied to samples from different datasets.
A clear block-diagonal pattern emerges, highlighted with red boxes, revealing that \textbf{\textit{similar tasks activate LoRAs in similar ways}}.
This observation provides empirical evidence that the extracted signals reflect meaningful semantic relationships among tasks and can effectively guide adapter selection without any additional training.

Finally, based on the collection of signals $\{s_i\}_{i=1}^N$, we select the \emph{top-$k$ adapter set}
\begin{equation}
\small
\mathcal{S} = \operatorname{TopK}\big(\{(L_i, s_i)\}_{i=1}^N,\, k\big),
\end{equation}
which contains the $k$ adapters with the highest scores.  
This set $\mathcal{S}$ serves as the candidate pool for merging in the next stage (Section~\ref{sec:lora_merging}).

\subsection{Merging LoRAs}
\label{sec:lora_merging}

After selecting the top-$k$ adapters $\mathcal{S}$ as described in Section~\ref{sec:lora_selection}, the next step is to merge them into the base model.  
Following prior work~\citep{zhao2024loraretriever}, we consider two types of merging strategies:
\begin{itemize}[leftmargin=*, itemsep=0pt, topsep=0pt]
    \item \textbf{Output-based Merging (Mixture).}  
    The projection outputs $\{\mathbf{p}_{i,T}\}_{i \in \mathcal{S}}$ are combined directly at the output level.  
    That is, given hidden input $\mathbf{h}_T$, the merged projection is formed as a weighted sum of the selected adapters’ projections.
    \item \textbf{Parameter-based Merging (Fusion).}  
    The low-rank parameter updates $\{\Delta \mathbf{W}_{i,T}^{(Q)}\}_{i \in \mathcal{S}}$ are merged into a single fused update, which is then re-attached to the base model.
\end{itemize}
While both strategies are possible, \model{} adopts \emph{mixture merging} for efficiency.  
Parameter-based fusion requires explicit recomputation of the merged weight matrices and re-attaching into the model at every step, which introduces significant overhead in deployment scenarios with many adapters.  
In contrast, output-based merging avoids additional overhead of parameter-level operations, since it directly discards unselected LoRAs during token generation and combines only the projections of the selected ones.

Formally, let $s_i$ be the signal score of adapter $L_i \in \mathcal{S}$ extracted from the probe pass.  
We normalize these scores into non-negative weights
\begin{equation}
\small
\tilde{w}_i = \frac{s_i}{\sum_{j \in \mathcal{S}} s_j}, \quad i \in \mathcal{S}.
\end{equation}
The merged projection is then given by
\begin{equation}
\small
\mathbf{o}_{\text{merge}} = \sum_{i \in \mathcal{S}} \tilde{w}_i \, \mathbf{o}_{i,T}.
\end{equation}
In practice, this weighted summation can be efficiently implemented by adjusting only the scaling factors of the selected adapters, without modifying or reloading their parameters.  
This design enables \model{} to adaptively merge multiple LoRAs with minimal runtime overhead, while maintaining flexibility to handle diverse inputs on the fly.  
\begin{table*}[ht]
% \scriptsize
% \tabcolsep2pt
\resizebox{\textwidth}{!}{
\begin{tabular}{@{}lccccccccccccccccccc@{}}
\toprule
 &  & \multicolumn{6}{c}{LLaMA-3.1-8B} & \multicolumn{6}{c}{Qwen-2.5-7B} & \multicolumn{6}{c}{DeepSeek-LLM-7B-Base} \\ \midrule
Task & \multicolumn{1}{c|}{Metric} & Base & \begin{tabular}[c]{@{}c@{}}Adpater-\\ Soup\end{tabular} & LoRAHub & \begin{tabular}[c]{@{}c@{}}LoRA-\\ Retriever\end{tabular} & \cellcolor[HTML]{DAE8FC}\begin{tabular}[c]{@{}c@{}}\model{}\\ (Norm)\end{tabular} & \multicolumn{1}{c|}{\cellcolor[HTML]{DAE8FC}\begin{tabular}[c]{@{}c@{}}\model{}\\ (Entropy)\end{tabular}} & Base & \begin{tabular}[c]{@{}c@{}}Adapter-\\ Soup\end{tabular} & LoRAHub & \begin{tabular}[c]{@{}c@{}}LoRA-\\ Retriever\end{tabular} & \cellcolor[HTML]{DAE8FC}\begin{tabular}[c]{@{}c@{}}\model{}\\ (Norm)\end{tabular} & \multicolumn{1}{c|}{\cellcolor[HTML]{DAE8FC}\begin{tabular}[c]{@{}c@{}}\model{}\\ (Entropy)\end{tabular}} & Base & \begin{tabular}[c]{@{}c@{}}Adapter-\\ Soup\end{tabular} & LoRAHub & \begin{tabular}[c]{@{}c@{}}LoRA-\\ Retriever\end{tabular} & \cellcolor[HTML]{DAE8FC}\begin{tabular}[c]{@{}c@{}}\model{}\\ (Norm)\end{tabular} & \cellcolor[HTML]{DAE8FC}\begin{tabular}[c]{@{}c@{}}\model{}\\ (Entropy)\end{tabular} \\ \midrule
\multicolumn{20}{l}{BBH} \\ \midrule
Boolean Expressions & \multicolumn{1}{c|}{EM} & 58.0 & 72.7 & 64.5 & \textbf{76.7} & \cellcolor[HTML]{DAE8FC}71.3 & \multicolumn{1}{c|}{\cellcolor[HTML]{DAE8FC}\textbf{76.7}} & 76.7 & \textbf{77.3} & 76.7 & \textbf{77.3} & \cellcolor[HTML]{DAE8FC}74.0 & \multicolumn{1}{c|}{\cellcolor[HTML]{DAE8FC}74.7} & 66.0 & \textbf{68.0} & 55.9 & 66.7 & \cellcolor[HTML]{DAE8FC}\textbf{68.0} & \cellcolor[HTML]{DAE8FC}66.0 \\
Causal Judgement & \multicolumn{1}{c|}{EM} & 41.4 & 48.3 & \textbf{54.0} & \textbf{54.0} & \cellcolor[HTML]{DAE8FC}48.3 & \multicolumn{1}{c|}{\cellcolor[HTML]{DAE8FC}47.1} & {\ul 59.8} & {\ul 59.8} & 51.5 & {\ul 59.8} & \cellcolor[HTML]{DAE8FC}\textbf{62.1} & \multicolumn{1}{c|}{\cellcolor[HTML]{DAE8FC}{\ul 59.8}} & 27.6 & \textbf{49.4} & 43.4 & {\ul 48.3} & \cellcolor[HTML]{DAE8FC}{\ul 48.3} & \cellcolor[HTML]{DAE8FC}14.9 \\
Formal Fallacies & \multicolumn{1}{c|}{EM} & 25.3 & \textbf{52.7} & 42.7 & 49.3 & \cellcolor[HTML]{DAE8FC}{\ul 52.0} & \multicolumn{1}{c|}{\cellcolor[HTML]{DAE8FC}50.0} & 53.3 & {\ul 55.3} & 53.2 & \textbf{56.7} & \cellcolor[HTML]{DAE8FC}54.7 & \multicolumn{1}{c|}{\cellcolor[HTML]{DAE8FC}53.3} & 0.7 & \textbf{48.7} & 31.7 & \textbf{48.7} & \cellcolor[HTML]{DAE8FC}\textbf{48.7} & \cellcolor[HTML]{DAE8FC}0.0 \\
Navigate & \multicolumn{1}{c|}{EM} & 45.3 & {\ul 54.7} & 48.5 & \textbf{56.7} & \cellcolor[HTML]{DAE8FC}50.0 & \multicolumn{1}{c|}{\cellcolor[HTML]{DAE8FC}50.0} & {\ul 50.7} & {\ul 50.7} & \textbf{52.0} & {\ul 50.7} & \cellcolor[HTML]{DAE8FC}{\ul 50.7} & \multicolumn{1}{c|}{\cellcolor[HTML]{DAE8FC}{\ul 50.7}} & 46.0 & 52.0 & 51.7 & \textbf{54.0} & \cellcolor[HTML]{DAE8FC}51.3 & \cellcolor[HTML]{DAE8FC}{\ul 53.3} \\
Object Counting & \multicolumn{1}{c|}{EM} & 35.3 & {\ul 38.0} & \textbf{39.1} & 20.7 & \cellcolor[HTML]{DAE8FC}27.3 & \multicolumn{1}{c|}{\cellcolor[HTML]{DAE8FC}24.0} & 41.3 & \textbf{44.0} & 35.2 & 40.7 & \cellcolor[HTML]{DAE8FC}\textbf{44.0} & \multicolumn{1}{c|}{\cellcolor[HTML]{DAE8FC}{\ul 43.3}} & {\ul 46.0} & \textbf{46.7} & 40.9 & 44.0 & \cellcolor[HTML]{DAE8FC}44.0 & \cellcolor[HTML]{DAE8FC}38.7 \\
Sports Understanding & \multicolumn{1}{c|}{EM} & 0.0 & {\ul 8.7} & \textbf{8.9} & 0.0 & \cellcolor[HTML]{DAE8FC}1.3 & \multicolumn{1}{c|}{\cellcolor[HTML]{DAE8FC}3.3} & 22.7 & 76.7 & 73.1 & 76.7 & \cellcolor[HTML]{DAE8FC}\textbf{80.0} & \multicolumn{1}{c|}{\cellcolor[HTML]{DAE8FC}{\ul 78.7}} & 0.0 & 0.0 & \textbf{26.8} & 0.0 & \cellcolor[HTML]{DAE8FC}0.0 & \cellcolor[HTML]{DAE8FC}{\ul 0.7} \\
Web-of-lies & \multicolumn{1}{c|}{EM} & 2.0 & {\ul 30.7} & 21.6 & \textbf{32.0} & \cellcolor[HTML]{DAE8FC}14.7 & \multicolumn{1}{c|}{\cellcolor[HTML]{DAE8FC}26.7} & 5.3 & 49.3 & 41.6 & \textbf{54.0} & \cellcolor[HTML]{DAE8FC}49.3 & \multicolumn{1}{c|}{\cellcolor[HTML]{DAE8FC}{\ul 51.3}} & 0.0 & 0.0 & \textbf{13.9} & 0.0 & \cellcolor[HTML]{DAE8FC}0.0 & \cellcolor[HTML]{DAE8FC}0.0 \\
Word Sorting & \multicolumn{1}{c|}{EM} & 11.3 & 34.7 & 16.7 & 34.0 & \cellcolor[HTML]{DAE8FC}{\ul 41.3} & \multicolumn{1}{c|}{\cellcolor[HTML]{DAE8FC}\textbf{42.0}} & 0.0 & 4.7 & 1.9 & \textbf{12.7} & \cellcolor[HTML]{DAE8FC}{\ul 12.0} & \multicolumn{1}{c|}{\cellcolor[HTML]{DAE8FC}9.3} & 3.3 & {\ul 6.0} & 4.4 & \textbf{7.3} & \cellcolor[HTML]{DAE8FC}5.3 & \cellcolor[HTML]{DAE8FC}1.3 \\
Average & \multicolumn{1}{c|}{} & 27.3 & \textbf{42.5} & 37.0 & {\ul 40.4} & \cellcolor[HTML]{DAE8FC}38.3 & \multicolumn{1}{c|}{\cellcolor[HTML]{DAE8FC}40.0} & 38.7 & 52.2 & 48.1 & \textbf{53.6} & \cellcolor[HTML]{DAE8FC}{\ul 53.3} & \multicolumn{1}{c|}{\cellcolor[HTML]{DAE8FC}52.6} & 23.7 & \textbf{33.8} & {\ul 33.6} & {\ul 33.6} & \cellcolor[HTML]{DAE8FC}33.2 & \cellcolor[HTML]{DAE8FC}21.9 \\ \midrule
\multicolumn{20}{l}{Translation} \\ \midrule
WMT'14 FR-\textgreater{}EN & \multicolumn{1}{c|}{BLEU} & 27.7 & \textbf{29.5} & 27.1 & {\ul 28.7} & \cellcolor[HTML]{DAE8FC}28.6 & \multicolumn{1}{c|}{\cellcolor[HTML]{DAE8FC}28.6} & 28.5 & \textbf{30.4} & 6.0 & 30.3 & \cellcolor[HTML]{DAE8FC}\textbf{30.4} & \multicolumn{1}{c|}{\cellcolor[HTML]{DAE8FC}30.3} & 26.0 & \textbf{27.7} & 25.4 & {\ul 27.2} & \cellcolor[HTML]{DAE8FC}27.0 & \cellcolor[HTML]{DAE8FC}26.5 \\
WMT'14 EN-\textgreater{}FR & \multicolumn{1}{c|}{BLEU} & 25.3 & {\ul 27.5} & 25.2 & \textbf{27.7} & \cellcolor[HTML]{DAE8FC}27.2 & \multicolumn{1}{c|}{\cellcolor[HTML]{DAE8FC}{\ul 27.5}} & 27.8 & 28.6 & 5.7 & \textbf{28.8} & \cellcolor[HTML]{DAE8FC}{\ul 28.7} & \multicolumn{1}{c|}{\cellcolor[HTML]{DAE8FC}{\ul 28.7}} & 19.3 & {\ul 22.3} & 21.3 & 22.2 & \cellcolor[HTML]{DAE8FC}\textbf{22.5} & \cellcolor[HTML]{DAE8FC}22.0 \\
WMT'16 DE-\textgreater{}EN & \multicolumn{1}{c|}{BLEU} & 29.8 & \textbf{31.5} & 29.8 & 30.7 & \cellcolor[HTML]{DAE8FC}{\ul 31.0} & \multicolumn{1}{c|}{\cellcolor[HTML]{DAE8FC}30.9} & 28.8 & \textbf{32.0} & 9.5 & {\ul 31.7} & \cellcolor[HTML]{DAE8FC}31.5 & \multicolumn{1}{c|}{\cellcolor[HTML]{DAE8FC}31.4} & 27.2 & \textbf{29.2} & 27.0 & 28.1 & \cellcolor[HTML]{DAE8FC}{\ul 28.6} & \cellcolor[HTML]{DAE8FC}27.8 \\
WMT'16 EN-\textgreater{}DE & \multicolumn{1}{c|}{BLEU} & 20.0 & 21.3 & 19.4 & 21.3 & \cellcolor[HTML]{DAE8FC}{\ul 21.6} & \multicolumn{1}{c|}{\cellcolor[HTML]{DAE8FC}\textbf{21.8}} & 20.0 & \textbf{20.6} & 4.2 & \textbf{20.6} & \cellcolor[HTML]{DAE8FC}\textbf{20.6} & \multicolumn{1}{c|}{\cellcolor[HTML]{DAE8FC}\textbf{20.6}} & 12.6 & \textbf{15.1} & 13.9 & 14.5 & \cellcolor[HTML]{DAE8FC}{\ul 15.0} & \cellcolor[HTML]{DAE8FC}14.3 \\
WMT'16 RO-\textgreater{}EN & \multicolumn{1}{c|}{BLEU} & 27.9 & 28.5 & 23.9 & \textbf{29.1} & \cellcolor[HTML]{DAE8FC}28.1 & \multicolumn{1}{c|}{\cellcolor[HTML]{DAE8FC}{\ul 28.7}} & 25.5 & {\ul 28.9} & 6.4 & \textbf{29.1} & \cellcolor[HTML]{DAE8FC}{\ul 28.9} & \multicolumn{1}{c|}{\cellcolor[HTML]{DAE8FC}{\ul 28.9}} & 25.9 & \textbf{27.6} & 19.5 & {\ul 27.1} & \cellcolor[HTML]{DAE8FC}26.7 & \cellcolor[HTML]{DAE8FC}26.0 \\
WMT'16 EN-\textgreater{}RO & \multicolumn{1}{c|}{BLEU} & 16.3 & 18.0 & 16.1 & {\ul 18.3} & \cellcolor[HTML]{DAE8FC}17.7 & \multicolumn{1}{c|}{\cellcolor[HTML]{DAE8FC}\textbf{18.4}} & 15.5 & \textbf{15.8} & 1.1 & 15.5 & \cellcolor[HTML]{DAE8FC}{\ul 15.7} & \multicolumn{1}{c|}{\cellcolor[HTML]{DAE8FC}{\ul 15.7}} & 11.0 & {\ul 14.1} & 12.5 & \textbf{14.2} & \cellcolor[HTML]{DAE8FC}{\ul 14.1} & \cellcolor[HTML]{DAE8FC}13.5 \\
Average & \multicolumn{1}{c|}{} & 24.5 & \textbf{26.0} & 23.6 & 25.9 & \cellcolor[HTML]{DAE8FC}25.7 & \multicolumn{1}{c|}{\cellcolor[HTML]{DAE8FC}\textbf{26.0}} & 24.4 & \textbf{26.0} & 5.5 & \textbf{26.0} & \cellcolor[HTML]{DAE8FC}25.9 & \multicolumn{1}{c|}{\cellcolor[HTML]{DAE8FC}25.9} & 20.3 & \textbf{22.7} & 19.9 & 22.2 & \cellcolor[HTML]{DAE8FC}{\ul 22.3} & \cellcolor[HTML]{DAE8FC}21.7 \\ \midrule
\multicolumn{20}{l}{Struct-to-Text} \\ \midrule
CommonGen & \multicolumn{1}{c|}{Rouge-1} & 54.8 & 55.5 & 46.8 & 55.5 & \cellcolor[HTML]{DAE8FC}\textbf{56.0} & \multicolumn{1}{c|}{\cellcolor[HTML]{DAE8FC}\textbf{56.0}} & 52.3 & 54.5 & 33.7 & \textbf{54.7} & \cellcolor[HTML]{DAE8FC}\textbf{54.7} & \multicolumn{1}{c|}{\cellcolor[HTML]{DAE8FC}54.3} & 0.0 & 52.0 & 51.9 & 50.8 & \cellcolor[HTML]{DAE8FC}{\ul 52.5} & \cellcolor[HTML]{DAE8FC}\textbf{53.5} \\
 & \multicolumn{1}{c|}{Rouge-2} & 23.6 & 24.3 & 20.4 & {\ul 24.9} & \cellcolor[HTML]{DAE8FC}24.7 & \multicolumn{1}{c|}{\cellcolor[HTML]{DAE8FC}\textbf{25.0}} & 21.5 & 22.9 & 14.0 & \textbf{23.1} & \cellcolor[HTML]{DAE8FC}\textbf{23.1} & \multicolumn{1}{c|}{\cellcolor[HTML]{DAE8FC}22.5} & 0.0 & {\ul 21.8} & 20.9 & 21.5 & \cellcolor[HTML]{DAE8FC}{\ul 21.8} & \cellcolor[HTML]{DAE8FC}\textbf{22.7} \\
 & \multicolumn{1}{c|}{Rouge-L} & 44.0 & 44.6 & 38.1 & 45.4 & \cellcolor[HTML]{DAE8FC}{\ul 45.5} & \multicolumn{1}{c|}{\cellcolor[HTML]{DAE8FC}\textbf{45.8}} & 41.6 & 44.4 & 27.6 & {\ul 44.7} & \cellcolor[HTML]{DAE8FC}\textbf{44.8} & \multicolumn{1}{c|}{\cellcolor[HTML]{DAE8FC}44.4} & 0.0 & 43.3 & 42.7 & 42.8 & \cellcolor[HTML]{DAE8FC}{\ul 44.0} & \cellcolor[HTML]{DAE8FC}\textbf{44.7} \\
DART & \multicolumn{1}{c|}{Rouge-1} & 63.9 & 66.3 & 53.8 & 63.6 & \cellcolor[HTML]{DAE8FC}{\ul 67.6} & \multicolumn{1}{c|}{\cellcolor[HTML]{DAE8FC}\textbf{68.7}} & 71.0 & 73.0 & 4.9 & 73.0 & \cellcolor[HTML]{DAE8FC}{\ul 73.2} & \multicolumn{1}{c|}{\cellcolor[HTML]{DAE8FC}\textbf{73.3}} & 0.5 & {\ul 62.2} & 61.3 & 60.4 & \cellcolor[HTML]{DAE8FC}\textbf{66.5} & \cellcolor[HTML]{DAE8FC}50.6 \\
 & \multicolumn{1}{c|}{Rouge-2} & 37.9 & 39.7 & 32.2 & 37.6 & \cellcolor[HTML]{DAE8FC}{\ul 40.9} & \multicolumn{1}{c|}{\cellcolor[HTML]{DAE8FC}\textbf{42.0}} & 44.2 & 47.2 & 1.8 & 47.1 & \cellcolor[HTML]{DAE8FC}{\ul 47.4} & \multicolumn{1}{c|}{\cellcolor[HTML]{DAE8FC}\textbf{47.5}} & 0.3 & {\ul 36.8} & 36.4 & 35.4 & \cellcolor[HTML]{DAE8FC}\textbf{40.0} & \cellcolor[HTML]{DAE8FC}27.9 \\
 & \multicolumn{1}{c|}{Rouge-L} & 47.9 & 49.6 & 40.7 & 48.1 & \cellcolor[HTML]{DAE8FC}{\ul 50.6} & \multicolumn{1}{c|}{\cellcolor[HTML]{DAE8FC}\textbf{51.4}} & 53.8 & \textbf{56.1} & 4.5 & 55.9 & \cellcolor[HTML]{DAE8FC}\textbf{56.1} & \multicolumn{1}{c|}{\cellcolor[HTML]{DAE8FC}56.0} & 0.4 & {\ul 47.9} & 47.3 & 46.8 & \cellcolor[HTML]{DAE8FC}\textbf{49.4} & \cellcolor[HTML]{DAE8FC}38.1 \\
E2ENLG & \multicolumn{1}{c|}{Rouge-1} & 65.4 & 67.2 & 53.6 & 64.7 & \cellcolor[HTML]{DAE8FC}\textbf{69.0} & \multicolumn{1}{c|}{\cellcolor[HTML]{DAE8FC}\textbf{69.0}} & 70.0 & \textbf{71.0} & 24.8 & 70.9 & \cellcolor[HTML]{DAE8FC}\textbf{71.0} & \multicolumn{1}{c|}{\cellcolor[HTML]{DAE8FC}\textbf{71.0}} & 0.0 & 60.6 & 58.6 & 57.8 & \cellcolor[HTML]{DAE8FC}\textbf{65.3} & \cellcolor[HTML]{DAE8FC}{\ul 60.7} \\
 & \multicolumn{1}{c|}{Rouge-2} & 35.8 & 37.4 & 30.0 & 35.6 & \cellcolor[HTML]{DAE8FC}{\ul 38.8} & \multicolumn{1}{c|}{\cellcolor[HTML]{DAE8FC}\textbf{39.0}} & 41.1 & 42.4 & 13.8 & 42.5 & \cellcolor[HTML]{DAE8FC}\textbf{42.7} & \multicolumn{1}{c|}{\cellcolor[HTML]{DAE8FC}{\ul 42.6}} & 0.0 & {\ul 32.8} & 31.9 & 30.6 & \cellcolor[HTML]{DAE8FC}\textbf{36.2} & \cellcolor[HTML]{DAE8FC}32.2 \\
 & \multicolumn{1}{c|}{Rouge-L} & 44.6 & 45.8 & 36.6 & 45.1 & \cellcolor[HTML]{DAE8FC}\textbf{46.9} & \multicolumn{1}{c|}{\cellcolor[HTML]{DAE8FC}{\ul 46.8}} & 48.6 & 49.6 & 18.6 & {\ul 49.7} & \cellcolor[HTML]{DAE8FC}\textbf{49.8} & \multicolumn{1}{c|}{\cellcolor[HTML]{DAE8FC}49.6} & 0.0 & {\ul 42.9} & 42.1 & 41.6 & \cellcolor[HTML]{DAE8FC}\textbf{44.9} & \cellcolor[HTML]{DAE8FC}42.3 \\
WebNLG & \multicolumn{1}{c|}{Rouge-1} & 58.4 & 66.8 & 64.2 & 63.7 & \cellcolor[HTML]{DAE8FC}{\ul 68.1} & \multicolumn{1}{c|}{\cellcolor[HTML]{DAE8FC}\textbf{69.0}} & 72.3 & \textbf{73.1} & 18.3 & {\ul 72.6} & \cellcolor[HTML]{DAE8FC}{\ul 72.6} & \multicolumn{1}{c|}{\cellcolor[HTML]{DAE8FC}{\ul 72.6}} & 0.0 & {\ul 63.4} & 54.4 & 60.8 & \cellcolor[HTML]{DAE8FC}\textbf{65.7} & \cellcolor[HTML]{DAE8FC}55.5 \\
 & \multicolumn{1}{c|}{Rouge-2} & 35.8 & 40.5 & 38.9 & 38.2 & \cellcolor[HTML]{DAE8FC}{\ul 41.8} & \multicolumn{1}{c|}{\cellcolor[HTML]{DAE8FC}\textbf{42.5}} & 46.4 & \textbf{47.7} & 10.8 & {\ul 47.1} & \cellcolor[HTML]{DAE8FC}46.5 & \multicolumn{1}{c|}{\cellcolor[HTML]{DAE8FC}46.4} & 0.0 & {\ul 39.2} & 33.2 & 37.8 & \cellcolor[HTML]{DAE8FC}\textbf{40.7} & \cellcolor[HTML]{DAE8FC}32.7 \\
 & \multicolumn{1}{c|}{Rouge-L} & 45.1 & 51.3 & 49.5 & 49.1 & \cellcolor[HTML]{DAE8FC}{\ul 52.1} & \multicolumn{1}{c|}{\cellcolor[HTML]{DAE8FC}\textbf{53.1}} & 55.8 & \textbf{57.1} & 14.7 & {\ul 56.6} & \cellcolor[HTML]{DAE8FC}56.3 & \multicolumn{1}{c|}{\cellcolor[HTML]{DAE8FC}56.3} & 0.0 & {\ul 50.9} & 44.0 & 48.8 & \cellcolor[HTML]{DAE8FC}\textbf{51.2} & \cellcolor[HTML]{DAE8FC}43.7 \\
Average & \multicolumn{1}{c|}{} & 46.4 & 49.1 & 42.1 & 47.6 & \cellcolor[HTML]{DAE8FC}{\ul 50.2} & \multicolumn{1}{c|}{\cellcolor[HTML]{DAE8FC}\textbf{50.7}} & 51.6 & \textbf{53.3} & 15.6 & {\ul 53.2} & \cellcolor[HTML]{DAE8FC}{\ul 53.2} & \multicolumn{1}{c|}{\cellcolor[HTML]{DAE8FC}53.0} & 0.1 & {\ul 46.2} & 43.7 & 44.6 & \cellcolor[HTML]{DAE8FC}\textbf{48.2} & \cellcolor[HTML]{DAE8FC}42.0 \\ \midrule
\multicolumn{20}{l}{Closed-Book QA} \\ \midrule
ARC-c & \multicolumn{1}{c|}{EM} & 64.7 & 67.2 & 64.8 & 67.9 & \cellcolor[HTML]{DAE8FC}{\ul 69.1} & \multicolumn{1}{c|}{\cellcolor[HTML]{DAE8FC}\textbf{69.9}} & 85.9 & 86.8 & 86.3 & \textbf{86.9} & \cellcolor[HTML]{DAE8FC}86.6 & \multicolumn{1}{c|}{\cellcolor[HTML]{DAE8FC}\textbf{86.9}} & 0.0 & 43.6 & \textbf{45.1} & \textbf{45.1} & \cellcolor[HTML]{DAE8FC}33.0 & \cellcolor[HTML]{DAE8FC}28.2 \\
ARC-e & \multicolumn{1}{c|}{EM} & 78.7 & 79.6 & 69.4 & 79.0 & \cellcolor[HTML]{DAE8FC}\textbf{81.3} & \multicolumn{1}{c|}{\cellcolor[HTML]{DAE8FC}{\ul 81.2}} & \textbf{91.6} & 91.4 & 22.3 & \textbf{91.6} & \cellcolor[HTML]{DAE8FC}\textbf{91.6} & \multicolumn{1}{c|}{\cellcolor[HTML]{DAE8FC}\textbf{91.6}} & 0.0 & \textbf{58.5} & {\ul 58.0} & 56.7 & \cellcolor[HTML]{DAE8FC}43.6 & \cellcolor[HTML]{DAE8FC}37.2 \\
Natural Questions & \multicolumn{1}{c|}{EM} & 5.3 & 11.0 & 7.1 & \textbf{12.6} & \cellcolor[HTML]{DAE8FC}11.1 & \multicolumn{1}{c|}{\cellcolor[HTML]{DAE8FC}{\ul 11.7}} & 4.8 & 11.8 & 6.8 & \textbf{12.8} & \cellcolor[HTML]{DAE8FC}{\ul 12.4} & \multicolumn{1}{c|}{\cellcolor[HTML]{DAE8FC}12.3} & 8.7 & \textbf{10.3} & 8.9 & {\ul 10.2} & \cellcolor[HTML]{DAE8FC}9.0 & \cellcolor[HTML]{DAE8FC}9.1 \\
Trivia QA & \multicolumn{1}{c|}{EM} & 12.7 & 14.1 & 12.2 & \textbf{15.0} & \cellcolor[HTML]{DAE8FC}{\ul 14.6} & \multicolumn{1}{c|}{\cellcolor[HTML]{DAE8FC}14.5} & 6.6 & \textbf{11.7} & 9.1 & {\ul 11.6} & \cellcolor[HTML]{DAE8FC}10.5 & \multicolumn{1}{c|}{\cellcolor[HTML]{DAE8FC}10.8} & 10.9 & {\ul 11.8} & 11.4 & \textbf{12.1} & \cellcolor[HTML]{DAE8FC}{\ul 11.8} & \cellcolor[HTML]{DAE8FC}11.3 \\
Average & \multicolumn{1}{c|}{} & 40.4 & 43.0 & 38.4 & 43.6 & \cellcolor[HTML]{DAE8FC}{\ul 44.0} & \multicolumn{1}{c|}{\cellcolor[HTML]{DAE8FC}\textbf{44.3}} & 47.2 & {\ul 50.4} & 31.1 & \textbf{50.7} & \cellcolor[HTML]{DAE8FC}50.3 & \multicolumn{1}{c|}{\cellcolor[HTML]{DAE8FC}{\ul 50.4}} & 4.9 & \textbf{31.1} & 30.8 & {\ul 31.0} & \cellcolor[HTML]{DAE8FC}24.4 & \cellcolor[HTML]{DAE8FC}21.4 \\ \midrule
\multicolumn{20}{l}{Natural Language Inference (NLI)} \\ \midrule
ANLI-R1 & \multicolumn{1}{c|}{EM} & 35.0 & 39.9 & 35.2 & 38.4 & \cellcolor[HTML]{DAE8FC}{\ul 42.0} & \multicolumn{1}{c|}{\cellcolor[HTML]{DAE8FC}\textbf{42.5}} & 55.5 & \textbf{61.5} & 58.6 & 61.2 & \cellcolor[HTML]{DAE8FC}61.3 & \multicolumn{1}{c|}{\cellcolor[HTML]{DAE8FC}\textbf{61.5}} & 7.5 & 33.4 & 12.8 & {\ul 33.5} & \cellcolor[HTML]{DAE8FC}\textbf{33.6} & \cellcolor[HTML]{DAE8FC}33.2 \\
ANLI-R2 & \multicolumn{1}{c|}{EM} & 32.6 & 39.4 & 31.4 & 40.5 & \cellcolor[HTML]{DAE8FC}{\ul 42.4} & \multicolumn{1}{c|}{\cellcolor[HTML]{DAE8FC}\textbf{42.8}} & 45.9 & 53.1 & 48.9 & {\ul 53.3} & \cellcolor[HTML]{DAE8FC}53.2 & \multicolumn{1}{c|}{\cellcolor[HTML]{DAE8FC}\textbf{53.7}} & 8.6 & {\ul 33.4} & 30.2 & \textbf{33.6} & \cellcolor[HTML]{DAE8FC}{\ul 33.4} & \cellcolor[HTML]{DAE8FC}33.1 \\
ANLI-R3 & \multicolumn{1}{c|}{EM} & 37.3 & 40.8 & 23.1 & 41.9 & \cellcolor[HTML]{DAE8FC}{\ul 43.0} & \multicolumn{1}{c|}{\cellcolor[HTML]{DAE8FC}\textbf{43.4}} & 51.2 & \textbf{54.5} & 52.4 & 54.2 & \cellcolor[HTML]{DAE8FC}{\ul 54.3} & \multicolumn{1}{c|}{\cellcolor[HTML]{DAE8FC}54.2} & 4.0 & \textbf{33.5} & 16.3 & 33.4 & \cellcolor[HTML]{DAE8FC}\textbf{33.5} & \cellcolor[HTML]{DAE8FC}33.2 \\
QNLI & \multicolumn{1}{c|}{EM} & 31.0 & {\ul 31.0} & \textbf{41.2} & 7.5 & \cellcolor[HTML]{DAE8FC}16.3 & \multicolumn{1}{c|}{\cellcolor[HTML]{DAE8FC}19.9} & 85.8 & \textbf{86.2} & 46.2 & {\ul 86.1} & \cellcolor[HTML]{DAE8FC}{\ul 86.1} & \multicolumn{1}{c|}{\cellcolor[HTML]{DAE8FC}86.0} & 0.2 & 3.4 & \textbf{4.3} & 2.4 & \cellcolor[HTML]{DAE8FC}1.4 & \cellcolor[HTML]{DAE8FC}{\ul 4.0} \\
Average & \multicolumn{1}{c|}{} & 34.0 & \textbf{37.8} & 32.7 & 32.1 & \cellcolor[HTML]{DAE8FC}35.9 & \multicolumn{1}{c|}{\cellcolor[HTML]{DAE8FC}{\ul 37.2}} & 59.6 & \textbf{63.8} & 51.5 & 63.7 & \cellcolor[HTML]{DAE8FC}63.7 & \multicolumn{1}{c|}{\cellcolor[HTML]{DAE8FC}\textbf{63.8}} & 5.1 & \textbf{25.9} & 15.9 & 25.7 & \cellcolor[HTML]{DAE8FC}25.5 & \cellcolor[HTML]{DAE8FC}\textbf{25.9} \\ \bottomrule
\end{tabular}
}
\caption{Performance of \model{} (with norm- and entropy-based signals) compared to baselines across diverse tasks on the LLaMA-3.1-8B, Qwen-2.5-7B, and DeepSeek-LLM-7B-Base backbones. The best results are in \textbf{bold}, the second-best are \underline{underlined}, and our results are highlighted with a \colorbox[HTML]{DAE8FC}{blue} background. For Struct-to-Text tasks, the reported average is computed over all ROUGE metrics (Rouge-1, Rouge-2, and Rouge-L).}
\label{tab:main}
\end{table*}

\begin{table}[t]
\resizebox{\columnwidth}{!}{
\begin{tabular}{@{}lcccc|
>{\columncolor[HTML]{DAE8FC}}c 
>{\columncolor[HTML]{DAE8FC}}c @{}}
\toprule
 & Base & \begin{tabular}[c]{@{}c@{}}Adapter-\\ Soup\end{tabular} & \begin{tabular}[c]{@{}c@{}}LoRA-\\ Hub\end{tabular} & \begin{tabular}[c]{@{}c@{}}LoRA-\\ Retriever\end{tabular} & \begin{tabular}[c]{@{}c@{}}\model{}\\ (Norm)\end{tabular} & \begin{tabular}[c]{@{}c@{}}\model{}\\ (Entropy)\end{tabular} \\ \midrule
Code Refinement & 13.8 & 41.3 & 33.5 & 29.2 & \textbf{46.3} & {\ul 41.5} \\
Code Translation: Java to C\# & 0.2 & 9.1 & 9.7 & {\ul 10.7} & \textbf{11.2} & 10.6 \\
Code Translation: C\# to Java & 1.6 & 9.2 & {\ul 10.4} & 9.4 & 7.6 & \textbf{11.7} \\
Code-to-Text: Java & \textbf{1.8} & 1.1 & {\ul 1.5} & 1.2 & 1.0 & 1.1 \\
Code-to-Text: Python & 1.5 & 1.7 & 3.2 & 1.5 & \textbf{6.1} & {\ul 3.7} \\ \midrule
Average & 3.8 & 12.5 & 11.6 & 10.4 & \textbf{14.4} & {\ul 13.7} \\ \bottomrule
\end{tabular}
}
\caption{Comparison of \model{} with baselines on unseen mixed-dataset scenarios using the LLaMA-3.1-8B backbone, evaluated on the CodeXGLUE dataset. The best results are in \textbf{bold}, the second-best are \underline{underlined}, and our results are highlighted with a \colorbox[HTML]{DAE8FC}{blue} background.}
\label{tab:mixture_scenario}
\end{table}

\section{Experiments}
We conduct extensive experiments to evaluate the performance and computational efficiency of \model{}.
Section~\ref{sec:evaluation_setup} outlines the experimental setup, including the base models, datasets, and baselines.
Section~\ref{sec:evaluation_results} reports the performance of \model{} across diverse datasets and in mixed-dataset scenarios, followed by Section~\ref{sec:computation_time}, which analyzes the inference-time throughput of our method.

\subsection{Evaluation Setup}
\label{sec:evaluation_setup}

\paragraph{Base Models and LoRA Adapters.}
We use \textbf{LLaMA-3.1-8B}, \textbf{Qwen-2.5-7B}, and \textbf{DeepSeek-LLM-7B-Base} as the base pretrained models in our evaluation.
For each model, we train \textbf{260 LoRA adapters} on distinct Flan-v2 tasks~\citep{wei2022finetuned, chung2024scaling}, a scale that covers a broad range of practical multi-task deployment settings while remaining memory-feasible for the probe pass.
Then, we evaluate \model{} as well as other baseline methods using the corresponding pretrained model with these adapters.
More details on LoRA training are in Appendix~\ref{sec:lora_training}.
\textbf{We plan to publicly release the adapters trained on all pretrained models.}

\paragraph{Datasets.}
We evaluate \model{} on a diverse set of benchmarks spanning multiple task categories.
For \textbf{BIG-Bench Hard (BBH)}~\citep{suzgun2023challenging}, we include Boolean Expressions, Causal Judgement, Formal Fallacies, Navigate, Object Counting, Sports Understanding, Web of Lies, and Word Sorting.  
For \textbf{Machine Translation}, we use datasets from the WMT benchmarks~\citep{bojar2014findings, bojar2016findings}, including WMT’14 FR$\rightarrow$EN, WMT’14 EN$\rightarrow$FR, WMT’16 DE$\rightarrow$EN, WMT’16 EN$\rightarrow$DE, WMT’16 RO$\rightarrow$EN, and WMT’16 EN$\rightarrow$RO.  
For \textbf{Struct-to-Text Generation}, we adopt datasets from the GEM benchmark~\citep{gehrmann2021gem}, including CommonGen~\citep{lin2020commongen}, DART~\citep{nan2021dart}, E2ENLG~\citep{novikova2017e2e}, and WebNLG~\citep{gardent2017creating}.  
For \textbf{Closed Book Question Answering}, we use ARC-c, ARC-e~\citep{clark2018think}, Natural Questions~\citep{kwiatkowski2019natural}, and TriviaQA~\citep{joshi2017triviaqa}.  
Finally, for \textbf{Natural Language Inference}, we evaluate on ANLI-R1, ANLI-R2, ANLI-R3~\citep{nie2020adversarial}, and QNLI~\citep{wang2018glue}.  
This collection covers reasoning, translation, structured generation, question answering, and inference tasks, providing a comprehensive evaluation of \model{} under diverse conditions.  

\paragraph{Baselines.}
We compare \model{} against four baselines:
\textbf{Base}, the base pretrained model without any LoRA adapters; 
\textbf{AdapterSoup}~\citep{chronopoulou2023adaptersoup}, which selects relevant LoRA adapters based on Sentence-BERT similarity between the input and the LoRA training datasets, and merges them via uniform averaging,
\textbf{LoRAHub}~\citep{huang2024lorahub}, which learns weights to merge the parameters of LoRA adapters via weighted summation; and \textbf{LoRARetriever}~\citep{zhao2024loraretriever}, which trains an auxiliary language model to retrieve the most relevant adapters for a given input based on embedding similarity.
For LoRARetriever, we report results with \emph{mixture} merging.
We use the implementation of all baseline methods following their guidelines.

\paragraph{Implementation Details.}
For all baselines, we fix the number of selected and merged LoRA adapters to 20 as a balanced and computationally efficient default that provides sufficient coverage of heterogeneous adapter pools, while keeping all other hyperparameters consistent with the default settings of the respective pretrained models.

We use the last Transformer block of each model as the target block for signal extraction, denoted as $B_T$, as instruction-tuned LoRAs primarily modify high-level behavioral representations in upper layers.
Similarly, the signal is extracted from the last token of each input sequence, whose hidden state directly precedes the model’s predicted response and is thus most informative for capturing adapter-induced behavior shifts.
Ablation studies on the number of selected adapters, the choice of Transformer block, the token used for signal extraction, and the merging method are presented in Appendix~\ref{sec:ablation},
which show that \model{} maintains stable performance across these configurations.
We further discuss compatibility with batched inference in Appendix~\ref{sec:batch_compatibility}.

\subsection{Evaluation Results}
\label{sec:evaluation_results}

\paragraph{Main Results}
Table~\ref{tab:main} reports the performance of \model{} with norm- and entropy-based signals compared to baseline methods on LLaMA-3.1-8B, Qwen-2.5-7B, and DeepSeek-LLM-7B-Base across multiple datasets.
\model{} consistently outperforms the baselines in many tasks and remains competitive in the rest.
This outcome is remarkable given that \model{} requires no additional training, whereas the baseline methods rely on fine-tuning.

\paragraph{Mixed-dataset Scenario}

To further assess the generalization ability of \model{} beyond the training domains of the LoRA adapters, we evaluate it on \textbf{CodeXGLUE}~\citep{lu2022codexglue}, a benchmark comprising diverse programming-language tasks unseen during adapter training.
Specifically, we consider five subtasks: Code Refinement, Code Translation (Java$\rightarrow$C\#), Code Translation (C\#$\rightarrow$Java), Code-to-Text (Java), and Code-to-Text (Python).
All tasks are evaluated using the BLEU metric.

As shown in Table~\ref{tab:mixture_scenario}, \model{} outperforms all baselines on average.
These results indicate that the signal-based selection and merging mechanism in \model{} generalizes effectively to unseen domains,
capturing cross-task relevance even when input distributions differ substantially from those of the adapters’ training data.

\subsection{Computation Time}
\label{sec:computation_time}

\begin{table}[t]
\resizebox{\columnwidth}{!}{
\begin{tabular}{@{}l|c|cccccc@{}}
\toprule
 & Training & \multicolumn{6}{c}{Inference} \\ \midrule
 & \begin{tabular}[c]{@{}c@{}}LoRA-\\ Hub\end{tabular} & Base & \begin{tabular}[c]{@{}c@{}}Adapter-\\ Soup\end{tabular} & \begin{tabular}[c]{@{}c@{}}LoRA-\\ Hub\end{tabular} & \begin{tabular}[c]{@{}c@{}}LoRA-\\ Retriever\end{tabular} & \begin{tabular}[c]{@{}c@{}}\model{}\\ (Norm)\end{tabular} & \begin{tabular}[c]{@{}c@{}}\model{}\\ (Entropy)\end{tabular} \\ \midrule
Boolean Expressions & 24.52 & 0.37 & 2.27 & 1.76 & 1.83 & 2.87 & 1.73 \\
Causal Judgement & 47.59 & 0.32 & 0.99 & 0.36 & 1.91 & 1.98 & 2.07 \\
Formal Fallacies & 25.29 & 0.44 & 0.67 & 0.39 & 1.85 & 1.77 & 1.65 \\
Navigate & 21.87 & 0.40 & 1.71 & 0.92 & 1.89 & 1.86 & 1.97 \\
Object Counting & 23.63 & 0.41 & 4.78 & 1.57 & 2.06 & 2.36 & 1.75 \\
Sports Understanding & 16.20 & 0.72 & 1.24 & 0.73 & 1.99 & 1.86 & 1.67 \\
Web-of-lies & 17.61 & 0.38 & 1.91 & 1.85 & 2.38 & 1.62 & 1.93 \\
Word Sorting & 17.51 & 0.76 & 0.83 & 1.61 & 2.29 & 2.32 & 2.22 \\ \midrule
Average & 24.28 & 0.47 & 1.80 & 1.15 & 2.03 & 2.08 & 1.87 \\ \bottomrule
\end{tabular}
}
\caption{Per-sample inference time (in seconds) of \model{} compared with baseline methods for LLaMA-3.1-8B model. For LoRAHub, we additionally report training time for learning task-specific merging weights.}
\label{tab:time_analysis}
\end{table}

We also analyze the computational cost of \model{} relative to baseline methods.
Table~\ref{tab:time_analysis} reports per-sample inference times on the LLaMA-3.1-8B model using a single NVIDIA H100 GPU.
As expected, the base pretrained model is the fastest, since no adapters are attached.
LoRAHub introduces additional overhead and requires training to learn task-specific merging weights (24.28 seconds on average), which limits its practicality when new adapters are frequently introduced.

\model{}, AdapterSoup, and LoRARetriever exhibit comparable inference times ($\sim$2 sec/sample), reflecting the cost of adapter-level operations.
Given that AdapterSoup and LoRARetriever require maintaining task datasets, and that LoRARetriever further involves training an auxiliary embedding model, these results underscore the practicality of \model{}.

\section{Analysis}

We conduct a series of analyses to gain deeper insights into the behavior and design choices of \model{}.
Section~\ref{sec:alignment_task_similarity} examines the alignment between merging weights and task similarity.
Section~\ref{sec:analysis_selected_loras} analyzes the characteristics of selected LoRAs.
Section~\ref{sec:scalability_analysis} evaluates the memory consumption and latency introduced by the probe pass in our method.
Section~\ref{sec:benefits_long_generation} highlights its benefits in long-generation scenarios.
Appendix~\ref{sec:cross_domain} demonstrates that applying domain-mismatched LoRA adapters can substantially degrade performance, Appdneix~\ref{sec:ablation} provides additional ablation studies of \model{}, Appendix~\ref{sec:robustness} examines the robustness of \model{} to adapter quality and rank variations, and Appendix~\ref{sec:lora_interference} analyzes representational interference among the selected adapters.

\subsection{Alignment with Task Similarity}
\label{sec:alignment_task_similarity}

\begin{figure}[t]
\centering
\subfloat[Norm]{
    \includegraphics[width=0.22\textwidth]{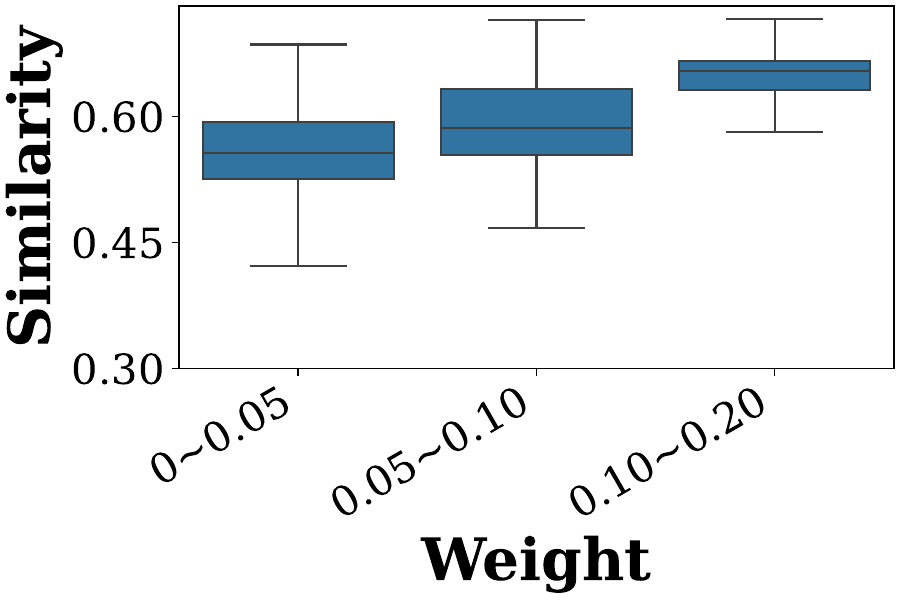}
}
\subfloat[Entropy]{
    \includegraphics[width=0.22\textwidth]{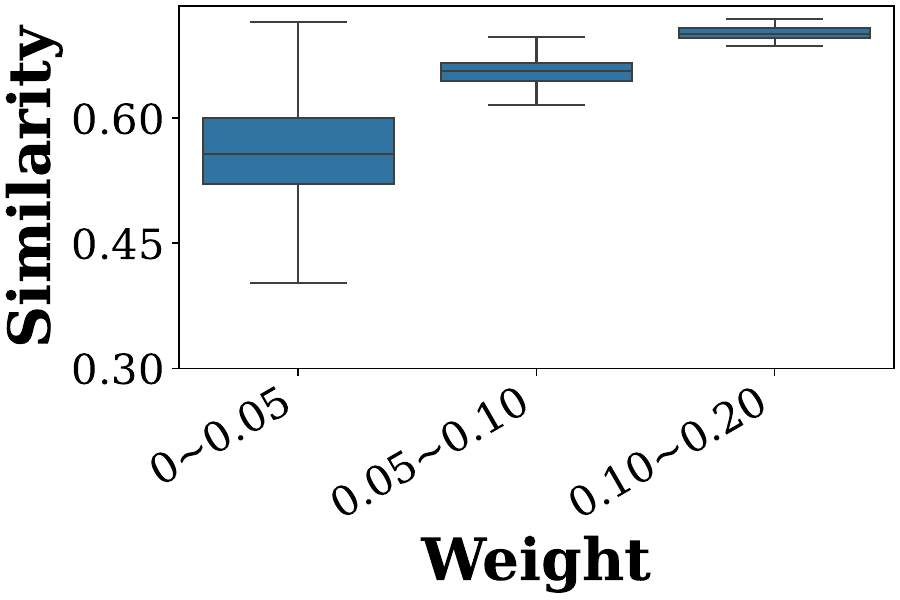}
}
\caption{Alignment between merging weights and task similarity of \model{} with (a) norm and (b) entropy as signals for BIG-Bench Hard task and DeepSeek-LLM-7B-Base.}
\label{fig:alignment_weight_similarity}
\end{figure}

We examine whether the merging weights produced by \model{} align with task similarity.
Since weights in \model{} are derived from projection-based signals, LoRAs trained on similar tasks should assign higher weights to each other’s inputs.

Task similarity is measured as the average cosine similarity, computed from the pretrained model’s embeddings between the given input and samples from the FLAN-v2 dataset.
Fig.~\ref{fig:alignment_weight_similarity} shows a box plot with merging weight on the $x$-axis and task similarity on the $y$-axis.
The results display a clear upward trend: LoRAs given larger weights correspond to more similar tasks.
%
% An exception occurs in the norm-based setting in the weight bucket 0.10–0.15, which appears anomalous because it arises from a single instance out of 22,740 cases ($1137$ samples $\times$ $20$ selected LoRAs).
% %
% This isolated case manifests as an outlier in the distribution.
%
This confirms that \model{}’s signal-based weighting not only enables efficient merging but also captures semantic relations between tasks without additional training.

% \subsection{Qualitative Analysis on Selected LoRAs}

\subsection{Analysis of Selected LoRAs}
\label{sec:analysis_selected_loras}

\begin{figure}
    \centering

\subfloat[LoRA Selection Count]{
    \includegraphics[width=0.95\columnwidth]{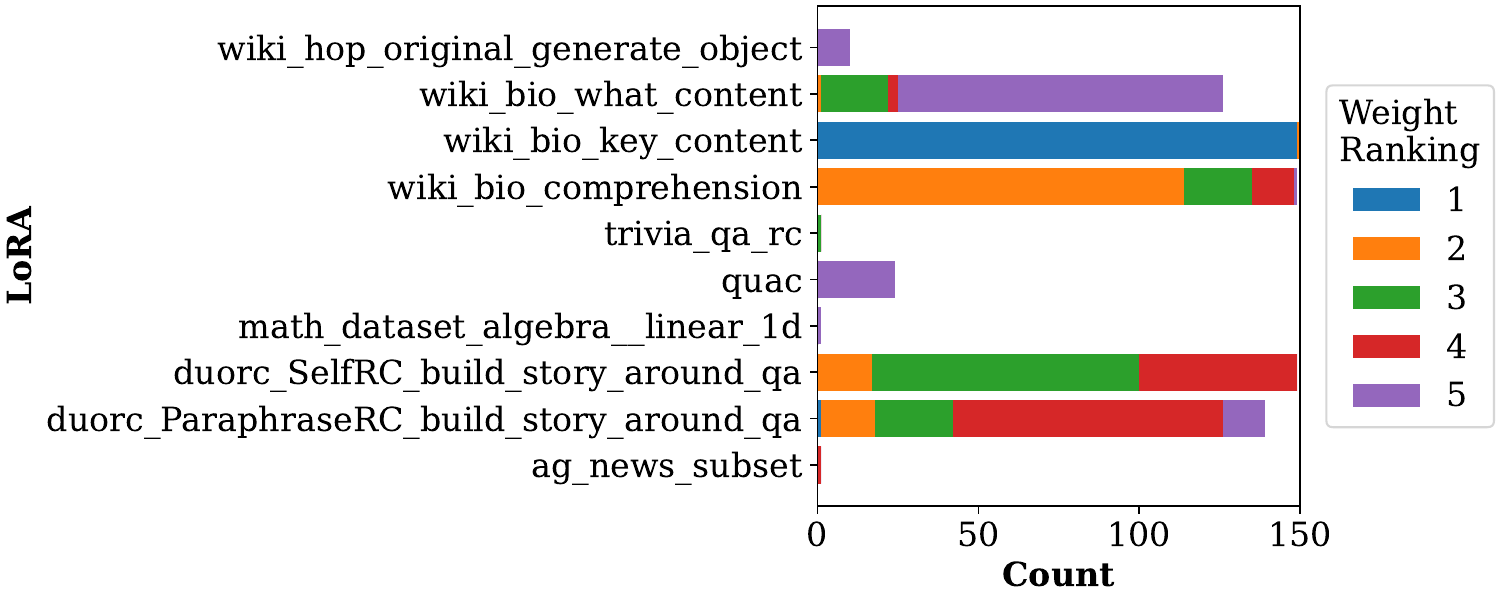}
    \label{fig:lora_selection_count}
}

\subfloat[Task Similarity]{
    \includegraphics[width=0.75\columnwidth]{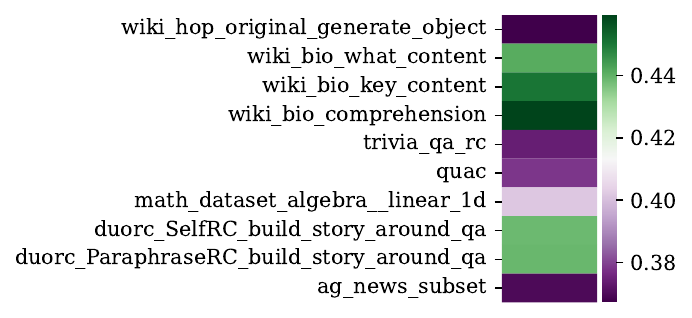}
    \label{fig:heatmap_task_similarity}
}
\caption{Comparison of (a) LoRA selection count by \model{} with Llama-3.1-8B model and (b) task similarity for BBH Word Sorting dataset. Each color in the bar present the priority of the LoRA when it was selected.}
\label{fig:selected_loras_main}
\end{figure}

\begin{table}[!t]
\centering

\begin{subtable}[t]{\columnwidth}
\resizebox{\textwidth}{!}{
\begin{tabular}{@{}l|l|l@{}}
\toprule
 & Sample A & Sample B \\ \midrule
Question & \begin{tabular}[c]{@{}l@{}}Sort the following words alphabetically:\\ List: thunderclap swab built poland\end{tabular} & \begin{tabular}[c]{@{}l@{}}Sort the following words alphabetically:\\ List: sanhedrin scratchy helical beau \\ venezuela awash bessie extricable \\ indoeuropean vice pendulum cream animism\end{tabular} \\ \midrule
Answer & built poland swab thunderclap & \begin{tabular}[c]{@{}l@{}}animism awash beau bessie cream extricable\\ helical indoeuropean pendulum sanhedrin\\  scratchy venezuela vice\end{tabular} \\ \bottomrule
\end{tabular}
}
\caption{Samples from BBH Word Sorting}
\end{subtable}

\begin{subtable}[t]{\columnwidth}
\resizebox{\textwidth}{!}{
\begin{tabular}{@{}l|l|l@{}}
\toprule
 & Sample A & Sample B \\ \midrule
Question & \begin{tabular}[c]{@{}l@{}}Given a meaning representation,\\ write a short and simple sentence \\ that contains all the information\\  in the meaning representation.\\ \\ Meaning Representation:\\ {[}eatType{[}pub{]}, food{[}Fast food{]},\\ customer rating{[}high{]}, area{[}riverside{]},\\ familyFriendly{[}no{]}, near{[}Café Rouge{]}{]}\end{tabular} & \begin{tabular}[c]{@{}l@{}}Given a meaning representation,\\ write a short and simple sentence\\ that contains all the information \\ in the meaning representation.\\ \\ Meaning Representation:\\ {[}name{[}Blue Spice{]}, eatType{[}pub{]},\\ food{[}Chinese{]}, area{[}city centre{]}, \\ near{[}Rainbow Vegetarian Café{]}{]}\end{tabular} \\ \midrule
Answer & \begin{tabular}[c]{@{}l@{}}The Mills is not kid friendly as it is a\\ riverside pub near Café Rouge. Its mid\\ priced fast food is highly rated.\end{tabular} & \begin{tabular}[c]{@{}l@{}}Blue Spice, located near Rainbow\\ Vegetarian Café in the city centre,\\ is a pub that also sells Chinese food.\\   Children should not visit.\end{tabular} \\ \bottomrule
\end{tabular}
}
\caption{Samples from E2ENLG}
\end{subtable}

\begin{subtable}[t]{\columnwidth}
\resizebox{\textwidth}{!}{
\begin{tabular}{@{}l|l|l|l@{}}
\toprule
Dataset & Common & Only in A & Only in B \\ \midrule
BBH Word Sorting & \begin{tabular}[c]{@{}l@{}}wiki\_bio\_key\_content,\\ wiki\_bio\_what\_content\end{tabular} & \begin{tabular}[c]{@{}l@{}}ag\_news\_subset,\\ math\_dataset\_algebra\\ \_\_linear\_1d,\\ trivia\_qa\_rc\end{tabular} & \begin{tabular}[c]{@{}l@{}}duorc\_SelfRC\_build\_\\ story\_around\_qa,\\ wiki\_bio\_comprehension,\\ wiki\_hop\_original\_\\ generate\_object\end{tabular} \\ \midrule
E2ENLG & \begin{tabular}[c]{@{}l@{}}duorc\_ParaphraseRC\_\\ build\_story\_around\_qa,\\ duorc\_SelfRC\_build\_\\ story\_around\_qa,\\ wiki\_bio\_comprehension,\\ wiki\_bio\_key\_content,\\ wiki\_bio\_what\_content\end{tabular} & \multicolumn{1}{c|}{N/A} & \multicolumn{1}{c}{N/A} \\ \bottomrule
\end{tabular}
}
\caption{LoRA adapters selected by \model{}}
\end{subtable}

\caption{Comparison of top-5 selected LoRA adapters for two samples from the BBH Word Sorting and E2ENLG dataset. For the E2ENLG dataset, selected LoRA adapters are common across both samples.}
\label{tab:qualitative}
\end{table}

In this section, we analyze the LoRA adapters selected by \model{}.
We employ \model{} with the LLaMA-3.1-8B model on the BBH Word Sorting task and report the selection counts of LoRA adapters in Figure~\ref{fig:lora_selection_count}.
Each color within a bar represents the ranking of a LoRA adapter when it was selected.
Results show that certain LoRA adapters are consistently selected across samples, although their relative priorities vary considerably.
We further compare these selection frequencies with each adapter’s task similarity to the BBH Word Sorting dataset, as shown in Figure~\ref{fig:heatmap_task_similarity}.
This confirms that LoRAs trained on tasks more similar to the target dataset tend to be selected more frequently.

\noindent\textbf{Anecdotal Examples:} 
We compare the LoRA selections of \model{} for two samples from the BBH Word Sorting and E2ENLG dataset in Table~\ref{tab:qualitative}.
For BBH Word Sorting, a small set of general-purpose adapters (e.g., wiki\_bio family) are consistently activated across both samples.
These modules are typically associated with reasoning and factual comprehension, providing a general foundation across tasks.
Beyond these common adapters, Sample A relies on adapters related to summarization (e.g., ag\_news), which facilitate short and structured outputs.
In contrast, Sample B engages modules emphasizing long story generation ability (e.g., duorc family).
Meanwhile, E2ENLG commonly selects general-purpose comprehension modules, as its story-generation task involves casual content that does not require domain-specific expertise.

\subsection{Ablation Study}

\begin{table}[t]
\resizebox{\columnwidth}{!}{
\begin{tabular}{@{}rr|rrrr
>{\columncolor[HTML]{DAE8FC}}r @{}}
\toprule
Task & Metric & Base & \begin{tabular}[c]{@{}r@{}}Top1-\\ Adapter\end{tabular} & \begin{tabular}[c]{@{}r@{}}Uniform\\ Weight\end{tabular} & \begin{tabular}[c]{@{}r@{}}Similarity-\\ based\end{tabular} & \model{} \\ \midrule
BBH & EM & 27.3 & 43.3 & 33.9 & \textbf{43.7} & 38.3 \\
Translation & BLEU & 24.5 & 23.3 & 25.1 & \textbf{25.7} & \textbf{25.7} \\
Struct-to-Text & Rouge & 46.4 & 40.7 & 48.4 & 48.9 & \textbf{50.2} \\
Closed-Book QA & EM & 40.4 & 40.2 & 42.6 & 43.7 & \textbf{44.0} \\
NLI & EM & 34.0 & 35.8 & 33.0 & \textbf{38.4} & 35.9 \\ \bottomrule
\end{tabular}
}
\caption{Comparison between \model{} and its variants.}
\label{tab:ablation}
\end{table}

We conduct an ablation study to evaluate adapter selection and merging strategies.
We compare \model{} with its three variants: (1) top-1 adapter selection, (2) merging adapters with uniform weights, and (3) similarity-based adapter selection and merging.
All experiments use LLaMA-3.1-8B as backbone, with \model{} using the norm-based weighting.
Except for Base and Top-1, all methods select and merge 20 LoRAs.

Table~\ref{tab:ablation} shows that merging multiple adapters consistently outperforms top-1 selection, highlighting the advantage of combining diverse adaptations.
Uniform-weight merging is consistently weaker than \model{}’s signal-weighted merging, indicating that the weighting mechanism contributes meaningfully to performance.
Compared to similarity-based selection, \model{} achieves competitive or better results across task categories while avoiding the need to maintain per-task datasets.

\subsection{Scalability Analysis}
\label{sec:scalability_analysis}

\begin{figure}[t]
\centering
\subfloat[Model Parameters]{
    \includegraphics[width=0.2\textwidth]{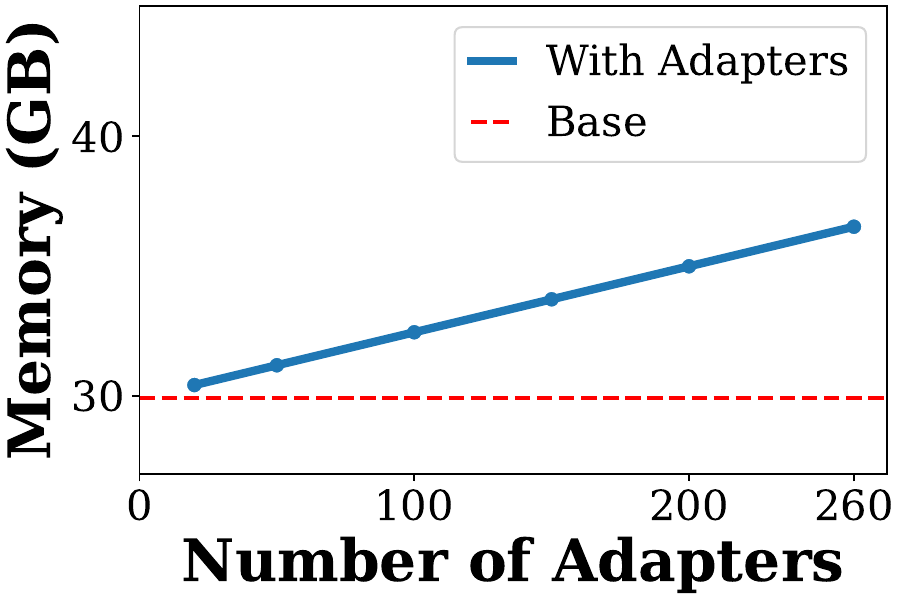}
    \label{fig:scalability_memory_parameter}
}
\hspace{1mm}
\subfloat[Activation in Probe Pass]{
    \includegraphics[width=0.2\textwidth]{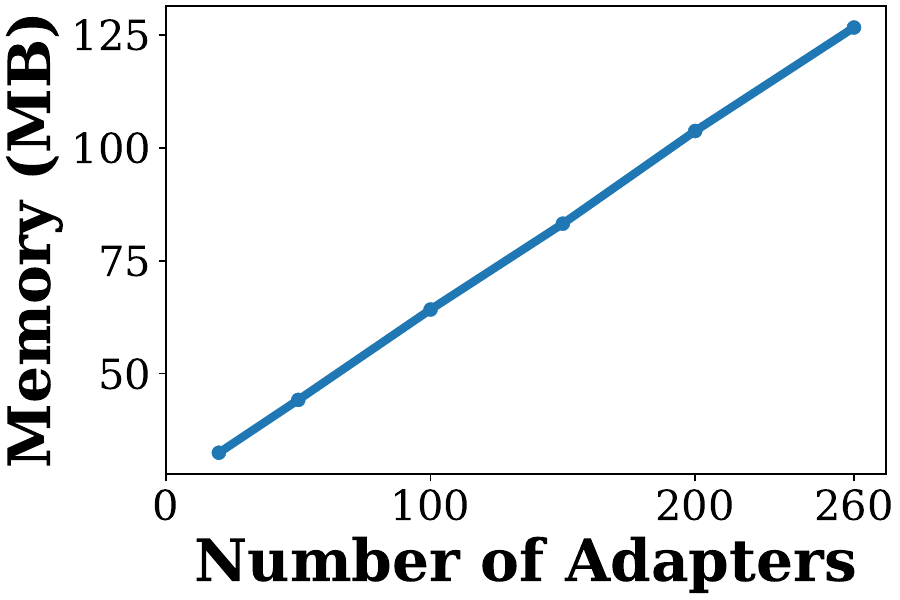}
    \label{fig:scalability_memory_activation}
}
\caption{Memory usage of \model{} for (a) model parameters and (b) activations during the probe pass.} 
\label{fig:scalability_memory}
\end{figure}

In this section, we analyze the computational and memory costs of our probe pass based selection procedure with respect to the total adapter pool size $N$.
Note that after selection, \model{} reduces the set to the top-$k$ adapters, where $k \ll N$. All subsequent merging and inference therefore incur the same cost as using only $k$ adapters concurrently.

\noindent\textbf{Memory.}
The overall additional memory overhead consists of two components: model parameters and activations during the probe pass.
Both parameter memory and activation memory scale linearly with the number of adapters (i.e., $O(N)$), as shown in Fig.~\ref{fig:scalability_memory}.
With LLaMA-3.1-8B and 260 LoRA adapters, the total parameter memory was 6.8 GB, which is $\sim$22\% of the pretrained model size (30.6 GB).
During the probe pass, we store only a single forward activation per probed layer; hence, the activation overhead is 126.7 MB (i.e., less than 1\% of the pretrained model size) even with 260 adapters.
We disable KV caching during probing to avoid unnecessary memory usage.

\noindent\textbf{Latency.}
The probe pass incurs a fixed overhead of $\sim$0.005s per adapter, roughly $8.3\%$ of the forward-pass time without an adapter ($\sim$0.06s).

\subsection{Benefits in Long Generation Task}
\label{sec:benefits_long_generation}

\begin{figure}
    \centering
\includegraphics[width=0.7\columnwidth]{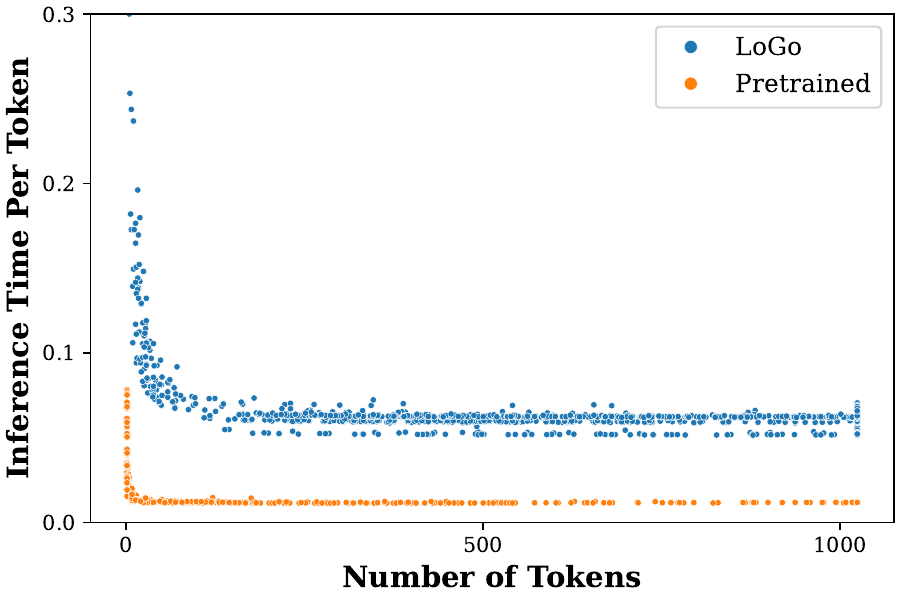}
\caption{Inference time per token with varying numbers of tokens in CNN-DailyMail dataset samples.}
\label{fig:long_generation}
\vspace{-2mm}
\end{figure}

We analyze how the computational overhead of \model{} behaves in long text generations.
Experiments are conducted on the \textbf{CNN-DailyMail}~\citep{see2017get} dataset using the LLaMA-3.1-8B model with an NVIDIA H200 GPU.
We measure the inference time per token as a function of the number of generated tokens, and compare \model{} against the pretrained model.

As shown in Figure~\ref{fig:long_generation}, the per-token inference time of \model{} decreases rapidly as the number of generated tokens increases, stabilizing after approximately 100 tokens.
While a gap remains between \model{} and the pretrained model, this difference reflects the intrinsic overhead of employing LoRA adapters.
These results indicate that the cost of signal extraction at the beginning of generation is effectively amortized over longer sequences.
\section{Conclusion}

We present LoRA on the Go (\model{}), a training-free framework that dynamically selects and merges LoRA adapters in deployment scenarios without any task-specific training.
By extracting lightweight relevance signals from a single forward pass, \model{} identifies the most suitable adapters for each input and merges them through signal-weighted summation.
Experiments across diverse NLP benchmarks show that \model{} achieves comparable or better performance than training-based baselines, while maintaining inference throughput.
These results highlight the potential of training-free, instance-specific adaptation as a promising direction for deploying large language models in real-world, heterogeneous environments.
\section*{Limitations}

While \model{} demonstrates strong performance without any task-specific training, several limitations remain.
First, our approach relies on projection-based signals extracted from a single forward pass. Although effective in practice, this mechanism does not guarantee that the selected adapters always align with task relevance, particularly in highly out-of-distribution scenarios.

Second, while the probe-based signals implicitly integrate multiple dimensions of relevance through forward activations, they do not explicitly disentangle factors such as topic, style, or skill. Exploring richer signals (e.g., multi-layer probes or attention-based signals) for finer-grained relevance decomposition is an interesting direction for future work.

Third, our experiments primarily use adapters fine-tuned on the FLAN-v2 dataset. Extending evaluation to LoRAs trained on diverse domains (e.g., multimodal, or low-resource data) would help assess generality.
%

% Fourth, the current framework assumes access to a shared pretrained model and a large pool of LoRA adapters. Attaching many adapters simultaneously can increase memory usage and slow inference, making optimization of adapter management (e.g., pruning or selective loading) an important direction for future work.

Fourth, the current framework assumes access to a shared pretrained model and a large pool of LoRA adapters. Attaching many adapters simultaneously can increase memory usage and slow inference. Furthermore, the probe pass introduces an additional forward computation, which increases the time to first token—particularly in short-text generation scenarios where the overhead cannot be amortized over longer outputs. Since the additional cost stems from low-rank projection operations, it is substantially cheaper than the base model's attention and MLP layers, and can be further reduced through practical optimizations such as caching probe results for repeated inputs, top-k preselection, or early-exit probing.

Fifth, the current framework does not account for redundancy within the adapter pool, which may lead to inefficient memory usage and suboptimal selection when multiple adapters encode similar representations. Addressing this through pool compression strategies, such as filtering adapters with redundant activation signals, pruning, or selective loading, remains an important direction for future work.

% \section*{Acknowledgments}
% We need to write Acknowledgement here.

\bibliography{reference}

\clearpage
\newpage

\appendix
\begin{figure*}[t]
\centering
\subfloat[LLaMA-3.1-8B signal intensity by norm]{
    \includegraphics[width=0.45\textwidth]{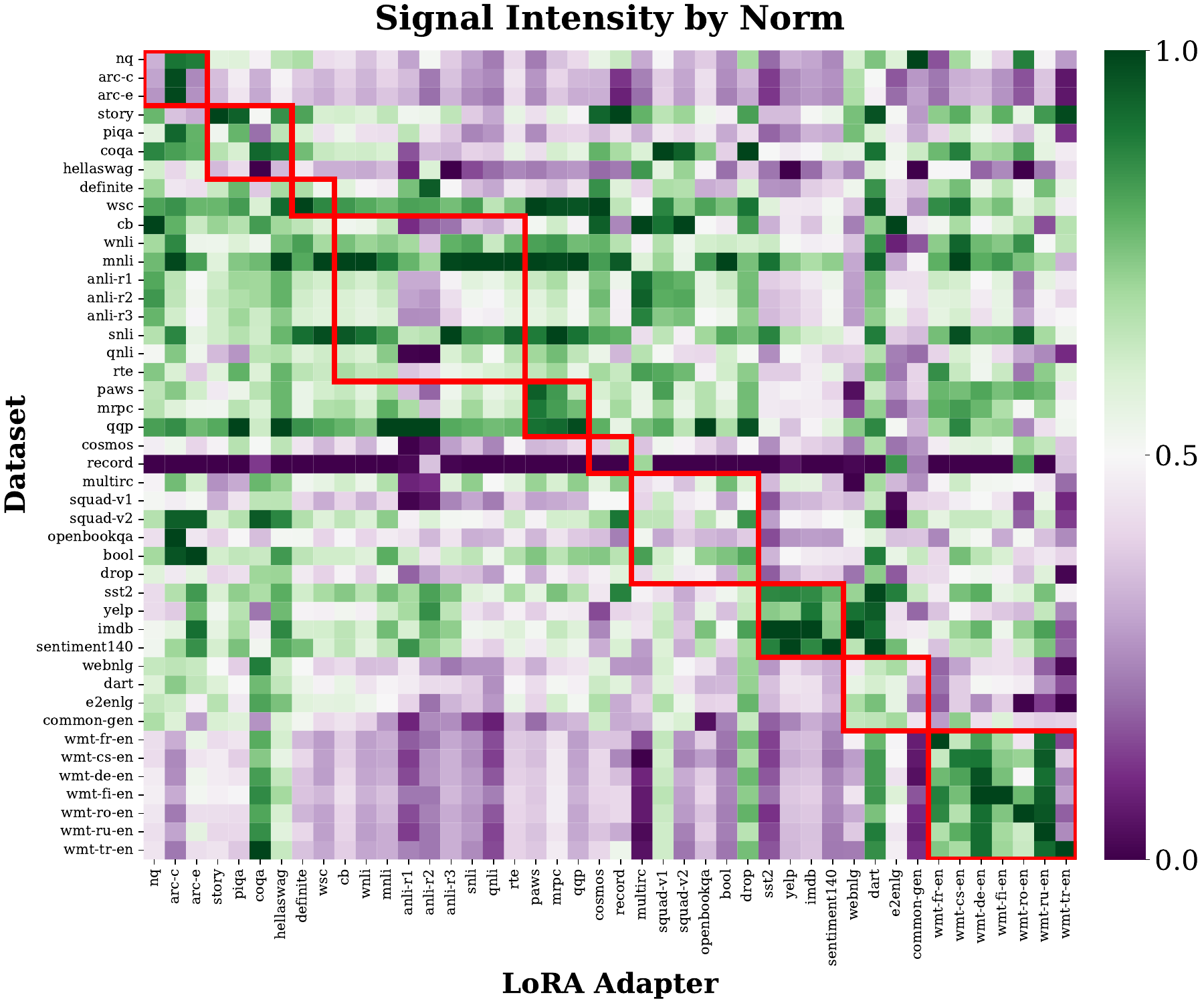}
}
\subfloat[LLaMA-3.1-8B signal Intensity by entropy]{
    \includegraphics[width=0.45\textwidth]{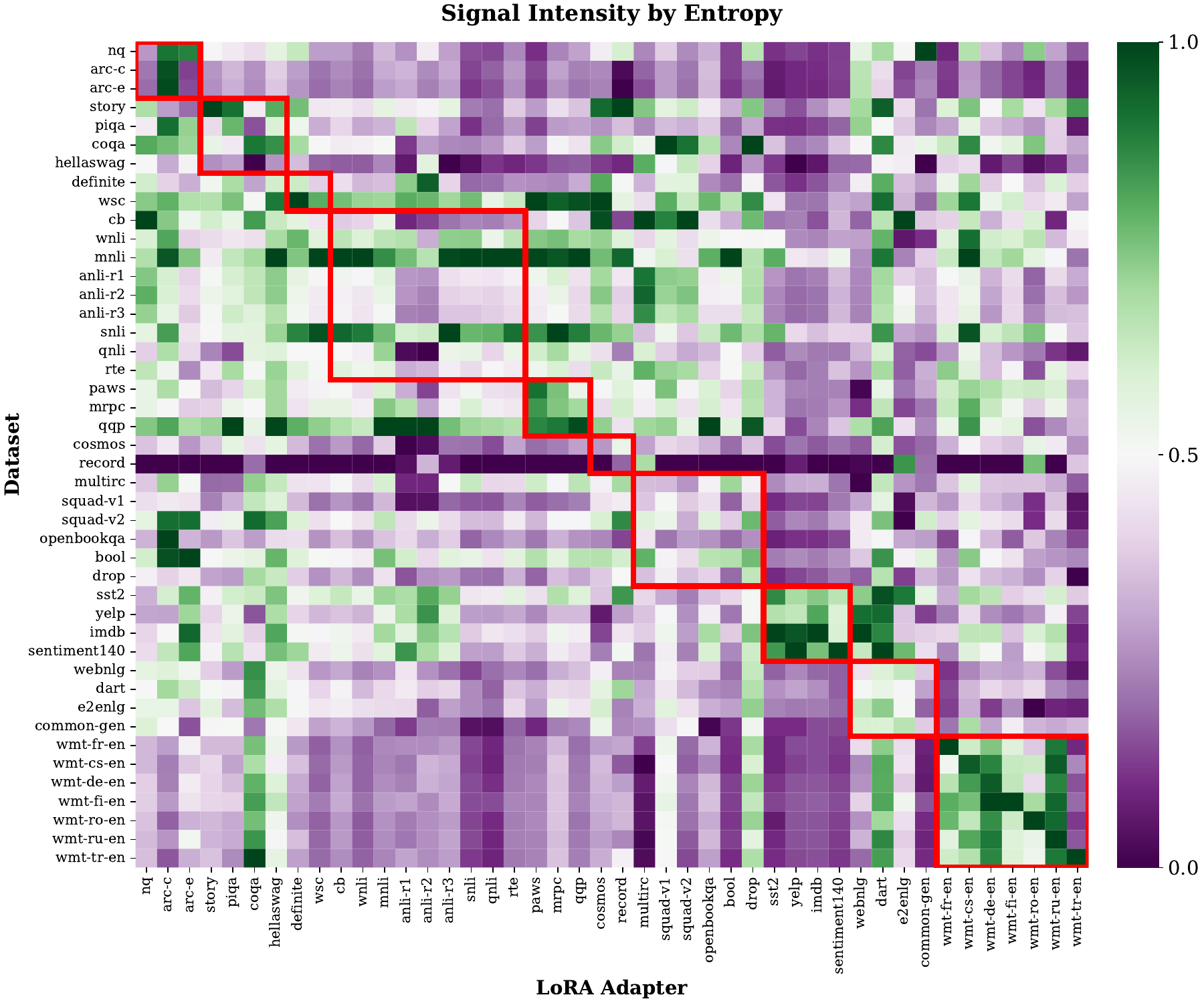}
}

\subfloat[Qwen-2.5-7B signal intensity by norm]{
    \includegraphics[width=0.45\textwidth]{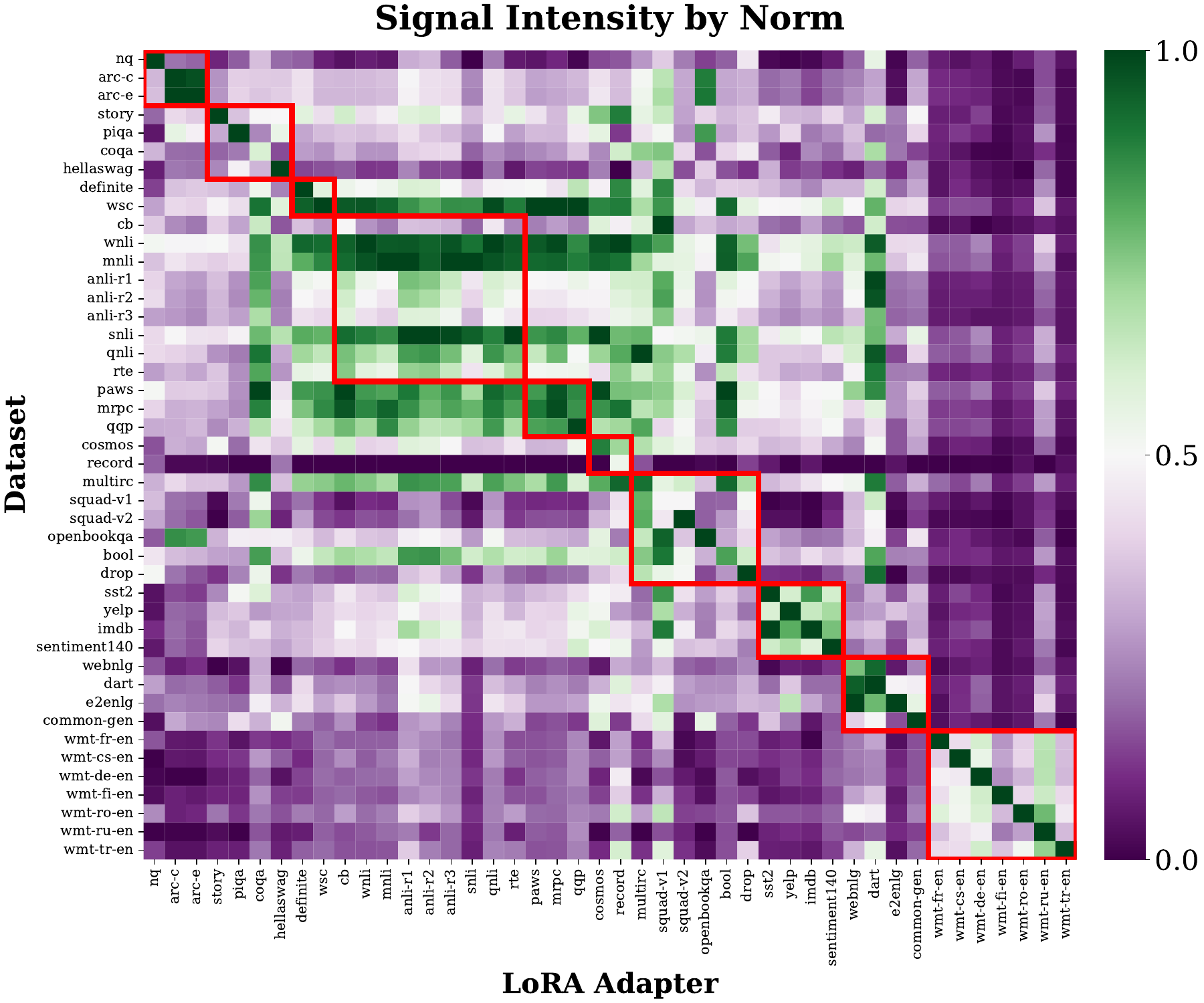}
}
\subfloat[Qwen-2.5-7B signal Intensity by entropy]{
    \includegraphics[width=0.45\textwidth]{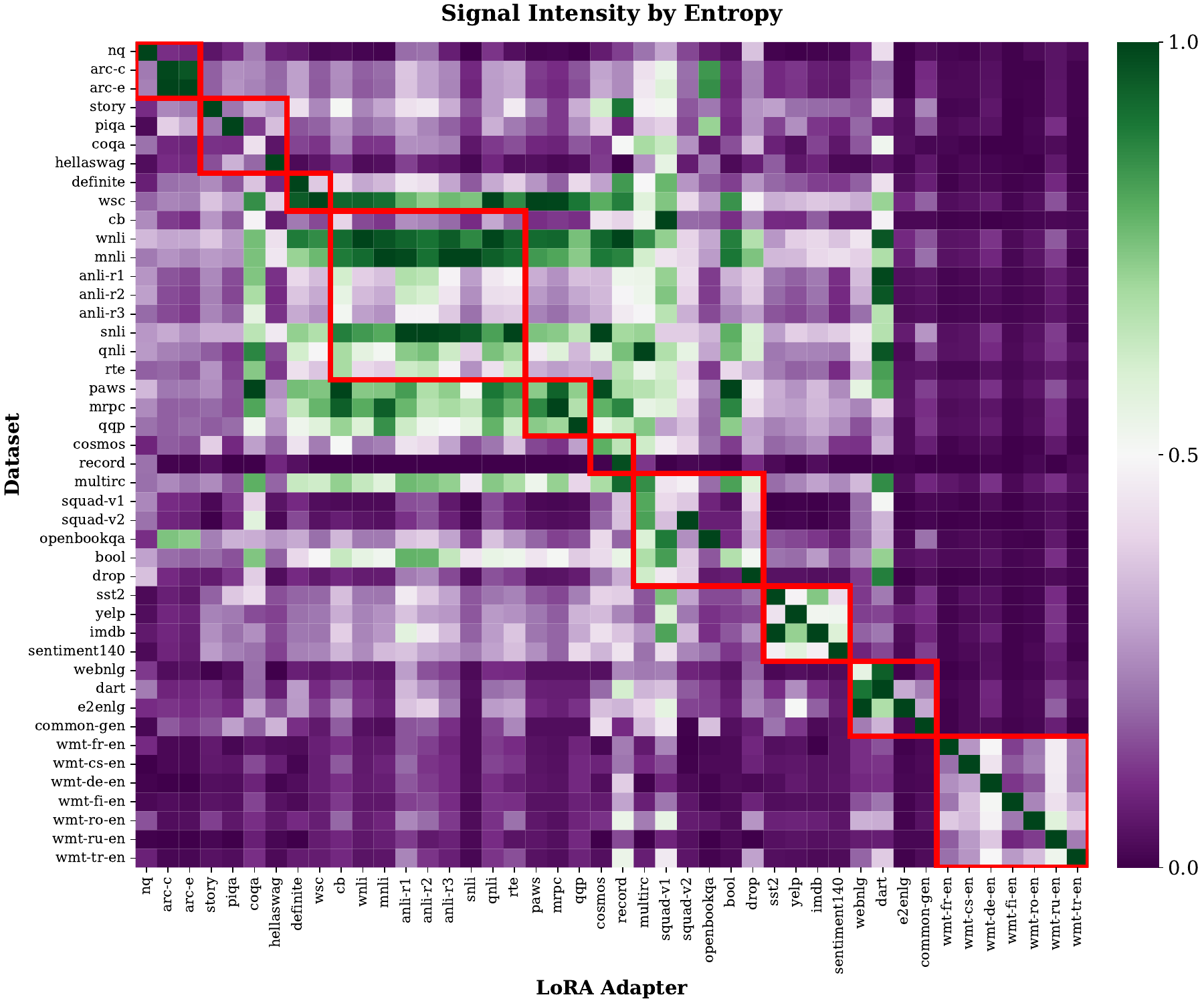}
}
\caption{Heatmaps illustrating signal intensity patterns across LoRA adapters trained on top of the LLaMA-3.1-8B and Qwen-2.5-7B backbone. The x-axis represents LoRAs trained on different tasks, while the y-axis corresponds to datasets from those tasks. Each cell shows the norm or (1 / entropy) of the projection outputs. The cell values are min-max normalized across datasets for each LoRA. The related tasks are highlighted in red boxes.}
\label{fig:signal_intensity_all}
\end{figure*}

\noindent {\Large \textbf{Overview of Appendix Sections}}

\begin{itemize}
\item Appendix~\ref{sec:full_signals}: Signals from LoRA Projections
\item Appendix~\ref{sec:signals_theory}: Theoretical Background of Projection-based Signals
\item Appendix~\ref{sec:algo}: The \model{} Algorithm
\item Appendix~\ref{app:additionalDetails}: Additional Details on Experiments
\item Appendix~\ref{app:additionalResults}: Additional Results
\end{itemize}

\section{Signals from LoRA Projections}
\label{sec:full_signals}

We analyze the signal intensity of LoRA projections across datasets for the LLaMA-3.1-8B and Qwen-2.5-7B models, using both norm and entropy as metrics.
Fig.~\ref{fig:signal_intensity_all} presents a heatmap where columns correspond to LoRAs trained on different tasks and rows correspond to datasets from those tasks.
Clear block structures emerge, with related tasks highlighted by red boxes, suggesting that similar tasks activate LoRAs in similar ways.

\section{Theoretical Background of Signals}
\label{sec:signals_theory}

Our choice of signals is grounded in empirical findings from the LoRA projection, which we complement with a theoretical background of their distinct roles below.

\paragraph{Norm.}
Let $z$ denote the activation at the last Transformer block of the base model, $\Delta z$ the LoRA-induced activation update at that layer, and $\mathcal{L}$ the cross-entropy loss.
A first-order Taylor expansion gives,
\begin{equation}
\mathcal{L}(z + \Delta z)
\approx
\mathcal{L}(z) + \nabla_z \mathcal{L}^\top \Delta z.
\end{equation}
Thus, the potential loss change depends on both the magnitude of $\Delta z$ and its alignment with the loss gradient.
By the Cauchy--Schwarz inequality,
\begin{equation}
\left| \nabla_z \mathcal{L}^\top \Delta z \right|
\le
\|\nabla_z \mathcal{L}\| \, \|\Delta z\|.
\end{equation}
This provides a first-order justification for using $\|\Delta z\|$, the norm of Lora update, as a proxy for the potential influence of an adapter on the loss.

\paragraph{Entropy.} Unlike the norm, entropy does not provide a direct bound on the loss change nor guarantee alignment with the loss gradient. Instead, it reflects the structural concentration of the activation update. Low-entropy updates correspond to concentrated, decisive modifications along a small subset of activation dimensions, whereas high-entropy updates introduce noisy perturbations.

\section{The \model{} Algorithm}
\label{sec:algo}

Algorithm~\ref{alg:model} summarizes the overall procedure of \model{}.

\begin{algorithm}[h!]
\footnotesize
\caption{\model{}: Dynamic Selection and Merging of LoRA Adapters}
\label{alg:model}
\begin{algorithmic}[1]
\State \textbf{Input:} Hidden input $\mathbf{h}_T$ at target block $B_T$, N tasks LoRAs $\mathcal{L}=\{L_i\}_{i=1}^N$, the respective low rank Q-projection update $\Delta W_{i,T}^{(Q)}$ , top-$k$ parameter $k$, scoring method
\State \textbf{Output:} Merged adapter projection $\mathbf{o}_{\text{merge}}$
\State \textbf{Probe pass:} Compute each LoRA's projection output
\For{$i \gets 1 \ldots N$}
    \State $\mathbf{o}_{i,T} \gets \Delta \mathbf{W}_{i,T}^{(Q)} \mathbf{h}_T$
    \If{scoring method = $\ell_2$}
        \State $s_i \gets \|\mathbf{o}_{i,T}\|_2$
    \ElsIf{scoring method = entropy}
        \State $p \gets \text{softmax}(\mathbf{o}_{i,T})$
        \State $s_i \gets 1 / \Big(-\sum_j p_j \log p_j \Big)$
    \EndIf
\EndFor

\State \textbf{Adapter selection:} 
\Statex $\mathcal{S} \gets \operatorname{TopK}(\{(L_i,s_i)\},k)$
\State Normalize weights $\tilde{w}_i = s_i / \sum_{j \in \mathcal{S}} s_j$ for each $L_i \in \mathcal{S}$

\State \textbf{Output-based merging:}
\State $\mathbf{o}_{\text{merge}} \gets \sum_{i \in \mathcal{S}} \tilde{w}_i \mathbf{o}_{i,T}$

\State \Return $\mathbf{o}_{\text{merge}}$

\end{algorithmic}
\end{algorithm}

\section{Additional Details on Experiments}
\label{app:additionalDetails}
\subsection{Details on LoRA Training}
\label{sec:lora_training}

\begin{table*}[h!]
\centering
\scriptsize
\begin{tabular}{p{3cm} p{13cm}}
\toprule
\textbf{Category} & \textbf{Datasets (Flan-v2 subsets used)} \\
\midrule
\textbf{Question Answering} &
adversarial\_qa\_dbert\_*, adversarial\_qa\_dbidaf\_*, adversarial\_qa\_droberta\_*, 
ai2\_arc\_ARC-Challenge, ai2\_arc\_ARC-Easy, bool\_q, coqa, cosmos\_qa, drop, 
duorc\_ParaphraseRC\_*, duorc\_SelfRC\_*, hotpotqa (kilt\_tasks\_hotpotqa\_*), 
natural\_questions\_open, openbookqa, qasc\_*, quac, quail\_*, quarel\_*, quartz\_*, 
quoref\_*, race\_high\_*, race\_middle\_*, ropes\_*, sciq\_*, squad\_v1.1, squad\_v2.0, 
trivia\_qa\_rc, unified\_qa\_science\_inst, web\_questions\_*, wiki\_hop\_original\_*, wiki\_qa\_* \\
\midrule
\textbf{Natural Language Inference} &
anli\_r1, anli\_r2, anli\_r3, glue\_mnli, glue\_rte, glue\_wnli, snli, super\_glue\_cb, 
super\_glue\_copa, super\_glue\_multirc, super\_glue\_record, super\_glue\_wic, 
super\_glue\_wsc.fixed \\
\midrule
\textbf{Classification / Sentiment} &
ag\_news\_subset, amazon\_polarity\_*, app\_reviews\_*, dbpedia\_14\_*, glue\_cola, glue\_mrpc, 
glue\_qqp, glue\_sst2, imdb\_reviews\_plain\_text, opinion\_abstracts\_idebate, 
opinion\_abstracts\_rotten\_tomatoes, sentiment140, trec, yelp\_polarity\_reviews \\
\midrule
\textbf{Commonsense Reasoning} &
cos\_e\_v1.11\_*, hellaswag, lambada, piqa, social\_i\_qa\_*, story\_cloze\_2016, winogrande, 
wiqa\_* \\
\midrule
\textbf{Summarization / Dialogue} &
aeslc, dream\_*, gem\_wiki\_lingua\_english\_en, samsum, gigaword \\
\midrule
\textbf{Data-to-Text / Structured Generation} &
gem\_common\_gen, gem\_dart, gem\_e2e\_nlg, gem\_web\_nlg\_en, wiki\_bio\_* \\
\midrule
\textbf{Translation} &
para\_crawl\_enes, wmt14\_translate\_fr-en, wmt16\_translate\_cs-en, wmt16\_translate\_de-en, 
wmt16\_translate\_fi-en, wmt16\_translate\_ro-en, wmt16\_translate\_ru-en, 
wmt16\_translate\_tr-en \\
\midrule
\textbf{Miscellaneous / Preprocessing} &
definite\_pronoun\_resolution, fix\_punct, huggingface, math\_dataset\_algebra\_\_linear\_1d, 
opinion\_abstracts\_*, true\_case, word\_segment \\
\bottomrule
\end{tabular}
\caption{Full list of Flan-v2 datasets used for LoRA training, grouped by task category. 
For brevity, ``*'' denotes multiple variants included in the collection.}
\label{tab:flanv2_datasets}
\end{table*}

We split each Flan-v2 dataset into training, validation, and test sets with an 8:1:1 ratio.
We trained the LoRA adapter with a per-device batch size of 4 and gradient accumulation of 16, resulting in an effective batch size of 64.
The learning rate was set to $2 \times 10^{-4}$, and training was conducted for 20 epochs.
The best model checkpoint was selected based on validation loss.
A full list of the Flan-v2 datasets used for training is provided in Table~\ref{tab:flanv2_datasets}.

\subsection{Implementation Details}
We implement \model{} based on Pytorch~\citep{paszke2019pytorch}, Huggingface~\citep{wolf2020transformers}, and PeFT library~\citep{peft}.
Specifically we utilize \textit{PeftMixedModel} class, which allows multiple adapters simultaneously and control their scales.
We fix the number of selected and merged LoRA adapters to 20, while keeping all other hyperparameters consistent with the default settings of the respective pretrained models used in our experiments.
We use the last Transformer block of each model as the target block for signal extraction.
Similarly, the signal is extracted from the last token of each input sequence.
Since the LoRA adapters are fixed and their respective signals (norm and entropy) are deterministic, we conduct our experiment over a single run.

\subsection{Details on Evaluation Datasets}
In our evaluation, we used the designated test split of each dataset. When test labels were not available, we relied on the validation or development split for evaluation. For training LoRAHub, we sampled five instances from the training split. For the BIG-Bench Hard datasets, we followed the same train–test split configuration as LoRAHub.

\subsection{Compatibility with Batched Inference}
\label{sec:batch_compatibility}
Our primary target scenario is heterogeneous or streaming inference, where inputs arrive individually and task identity is not known a priori. In such settings, instance-level selection is the more natural choice, but it does not preclude efficient batched inference.

Specifically, selection is implemented as a lightweight scoring step over adapter activations from a single shared forward pass. Because all LoRA adapters can be attached simultaneously, their activations and relevance scores can be computed in parallel using standard batched tensor operations. Merging is likewise implemented as a weighted sum of adapter activations, so instance-specific adapter weights do not require separate forward passes.

Thus, the overhead scales linearly with the number of adapters but introduces no sequential dependency across batch elements. In practice, it is negligible relative to the base model forward pass and is further amortized in long-generation tasks. In addition, \model{} requires no retraining, retrieval model, or labeled data, simplifying deployment in dynamic adapter pools.

\section{Additional Results}
\label{app:additionalResults}

\subsection{Cross-Domain Evaluation of LoRAs}
\label{sec:cross_domain}

\begin{table}[t]
\centering

\resizebox{\columnwidth}{!}{
\begin{tabular}{llccc}
\toprule
& & \multicolumn{3}{c}{\textbf{LoRA Trained on}} \\
\cmidrule(lr){3-5}
\textbf{Task} & \textbf{Base} & \textbf{GSM8K} & \textbf{MedQA} & \textbf{WritingPrompts} \\
\midrule
GSM8K          & 3.8 & \textbf{14.6} & 2.4  & 1.4  \\
MedQA          & 0.5 & 0.5  & \textbf{63.1} & 0.6  \\
WritingPrompts & 20.1  & 20.6 & 20.7 & \textbf{22.2} \\
\bottomrule
\end{tabular}
}
\caption{Performance of domain-specific LoRA adapters evaluated across conflicting domains. Accuracy is reported for GSM8K and MedQA; ROUGE-1 is reported for WritingPrompts. The best results are in \textbf{bold}.}
\label{tab:cross_domain}
\end{table}

It is known that applying a LoRA adapter to out-of-domain inputs can degrade model performance.
To empirically verify this, we conduct a cross-domain evaluation across three conflicting domains: creative writing, mathematical reasoning, and medical QA.
Results in Table~\ref{tab:cross_domain} exhibit a clear diagonal pattern -- each domain-specific adapter performs best on its matched domain, while activation on a mismatched domain leads to substantial performance degradation, often falling below the base model.
This underscores the importance of accurate instance-level adapter selection to prevent harmful interference from domain-mismatched adapters.

\subsection{Additional Ablation Studies}
\label{sec:ablation}
In this section, we conduct ablation studies on \model{} to examine the effects of the token used for signal extraction, the number of selected modules, the specific block used for signal extraction, and merging method.
All experiments are performed using \model{} with LLaMA-3.1-8B on the BIG-Bench Hard, Translation, Struct-to-Text, Closed-Book QA, and Natural Language Inference tasks.
Evaluation metrics are BLEU for translation, ROUGE for Struct-to-Text, and Exact Match for the other datasets.
The reported results are averaged over each dataset category.

\noindent
\paragraph{Token for Signal Extraction}
\begin{figure}[t]
\centering

\subfloat{
    \includegraphics[width=0.3\textwidth]{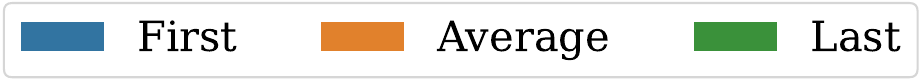}
}
\addtocounter{subfigure}{-1}
\subfloat[Norm]{
    \includegraphics[width=0.46\columnwidth]{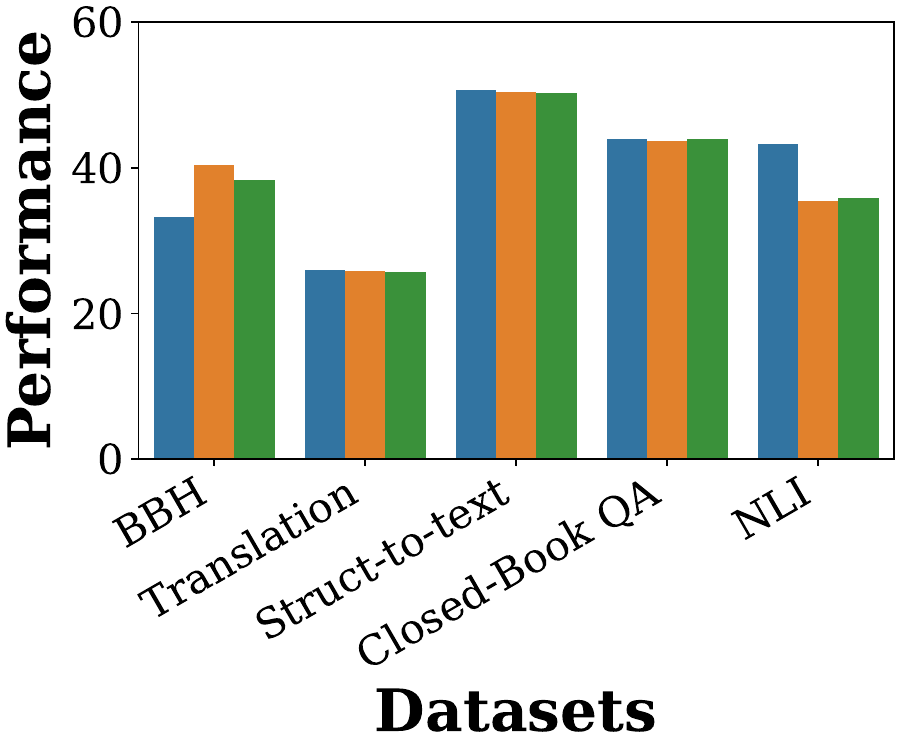}
}
\hspace{1mm}
\subfloat[Entropy]{
    \includegraphics[width=0.46\columnwidth]{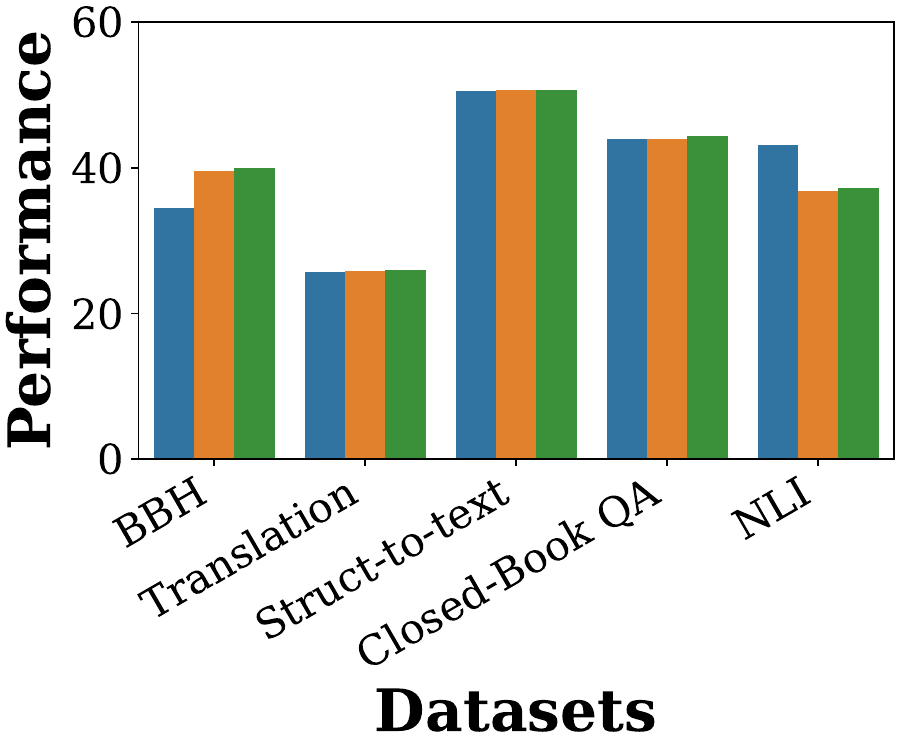}
}
\caption{Performance of \model{} using (a) norm and (b) entropy across datasets with different tokens for signal extraction, first, average, and last. Here, average denotes the mean signal values across all tokens. The last token serves as the default setting of our method.}
\label{fig:token_for_signal}
\end{figure}

In \model{}, signals are extracted from the projection outputs corresponding to the last token of the input, which serves as our default setting.
To investigate the effect of this choice, we compare three alternatives: using the first token, the last token, and the average across all tokens.
Results are shown in Fig.~\ref{fig:token_for_signal}, which presents bar plots comparing all alternatives.
Across both norm- and entropy-based scoring, we observe that the performance differences between token choices are small, indicating that \model{} is robust to this design decision.
Among the three options, however, the last token consistently achieves slightly higher performance across most datasets, supporting its use as the default configuration in our method.

\noindent
\paragraph{Number of Selected  Modules}

\begin{figure}[t]
\centering

\subfloat{
    \includegraphics[width=0.97\columnwidth]{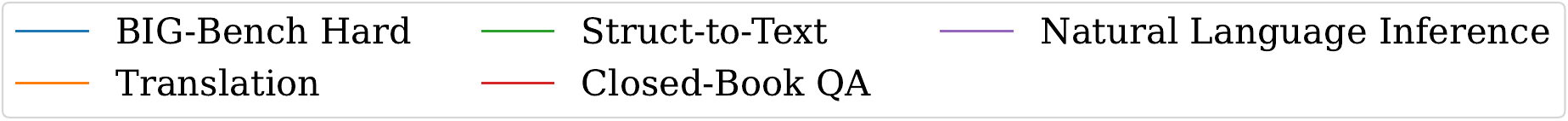}
}
\addtocounter{subfigure}{-1}

\subfloat[Norm]{
    \includegraphics[width=0.46\columnwidth]{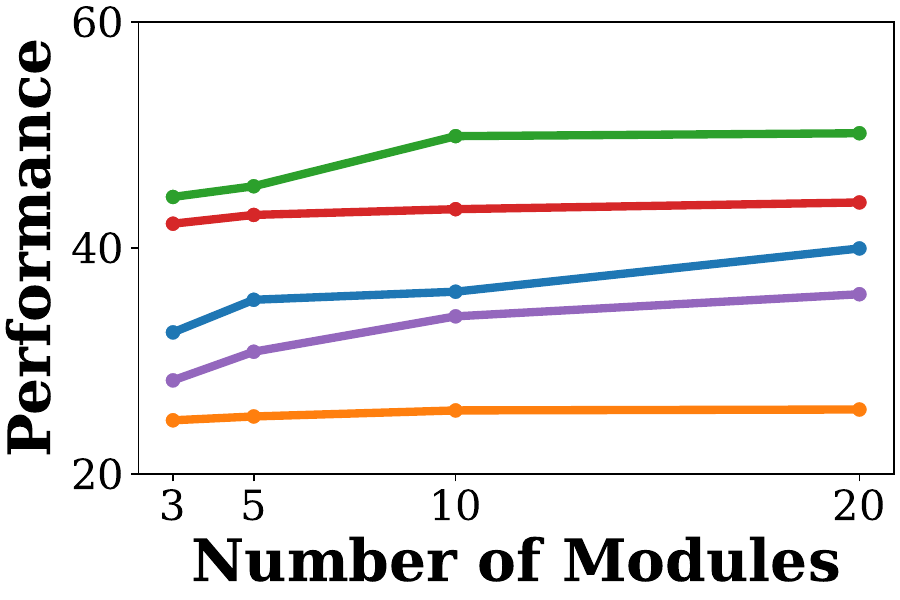}
}
\hspace{1mm}
\subfloat[Entropy]{
    \includegraphics[width=0.46\columnwidth]{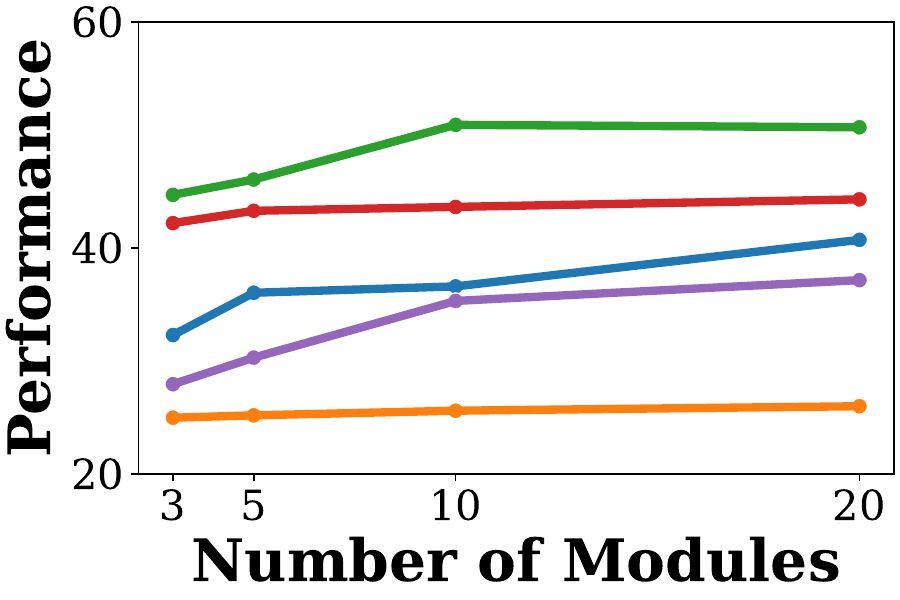}
}
\caption{Performance of \model{} using (a) norm and (b) entropy across datasets with different numbers of selected modules. }
\label{fig:n_selected_modules}
\end{figure}

We analyze the effect of varying the number of selected modules $k$ in \model{} by evaluating performance with $k \in {3, 5, 10, 20}$.
Fig.~\ref{fig:n_selected_modules} presents the results for both norm- and entropy-based scoring.
Overall, performance improves as the number of selected modules increases, but the gains are relatively modest.
This suggests that \model{} is not highly sensitive to the exact choice of $k$: even with only a few modules, it achieves performance close to the larger settings.
Such robustness further highlights the practicality of \model{}, as it enables efficient operation with a small number of modules while retaining strong performance.

\noindent
\paragraph{Block for Signal Extraction}

\begin{figure}[t]
\centering

\subfloat{
    \includegraphics[width=0.97\columnwidth]{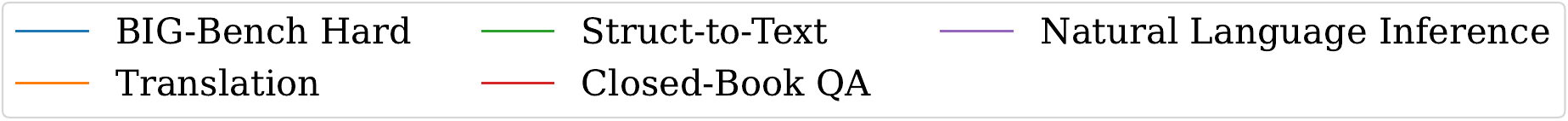}
}
\addtocounter{subfigure}{-1}

\subfloat[Norm]{
    \includegraphics[width=0.46\columnwidth]{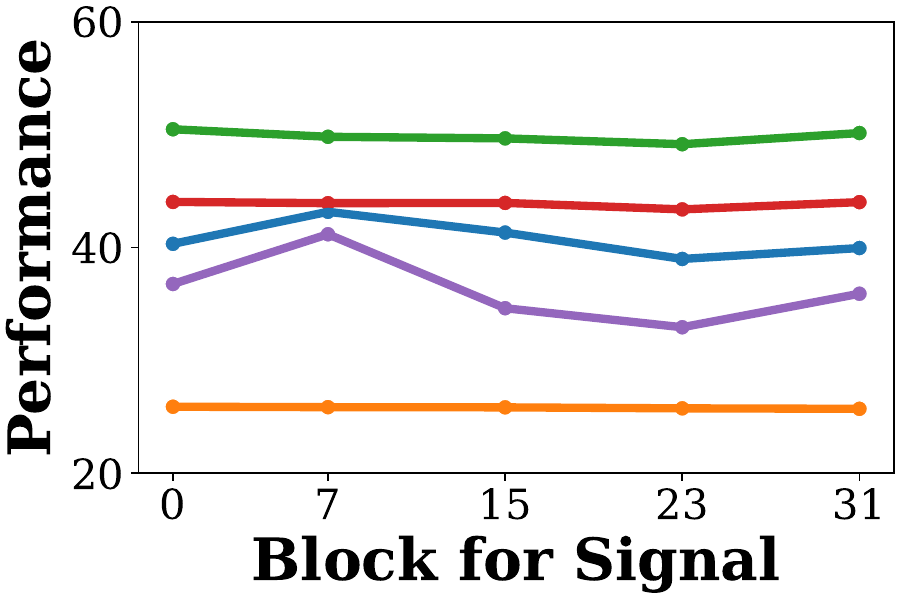}
}
\hspace{1mm}
\subfloat[Entropy]{
    \includegraphics[width=0.46\columnwidth]{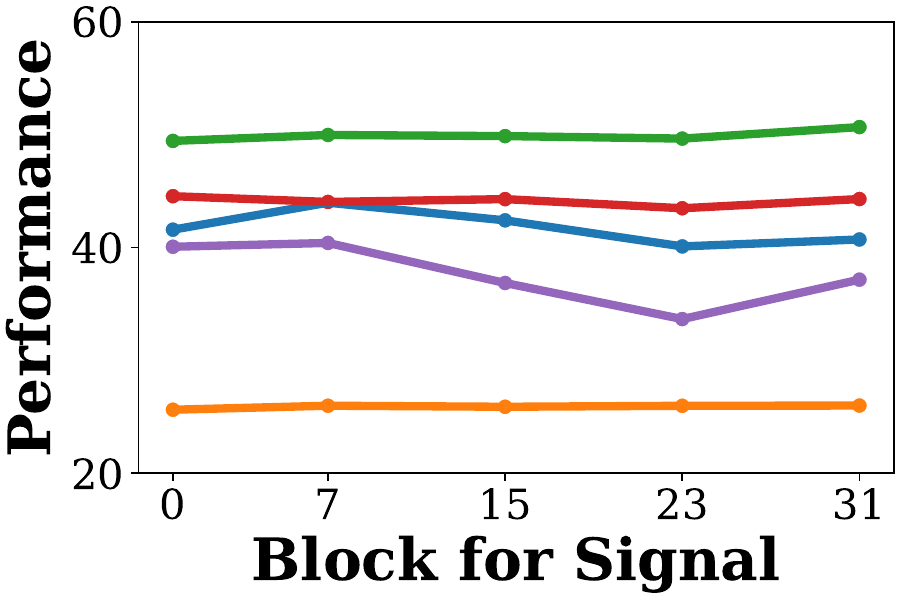}
}
\caption{Performance of \model{} using (a) norm and (b) entropy across datasets with different target block for signal extraction.}
\label{fig:n_target_block}
\end{figure}

To assess the sensitivity of \model{} to the layer from which signals are extracted, we vary the target Transformer block used for signal computation.
Specifically, we extract projection-based signals from the $0$-th, $7$-th, $15$-th, $23$-rd, and $31$-st blocks.
Results in Fig.~\ref{fig:n_target_block} show minor variations in performance across layers, indicating that \model{} is not sensitive to the specific block chosen for signal extraction.
This suggests that task-relevant activation patterns are distributed across multiple layers, and that \model{} can robustly estimate adapter relevance from various depths without requiring careful layer tuning.

\begin{figure}[t]
\centering

\subfloat{
    \includegraphics[width=0.23\textwidth]{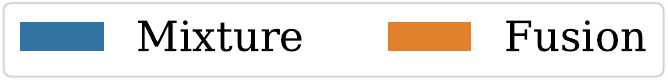}
}
\addtocounter{subfigure}{-1}

\subfloat[Norm]{
    \includegraphics[width=0.46\columnwidth]{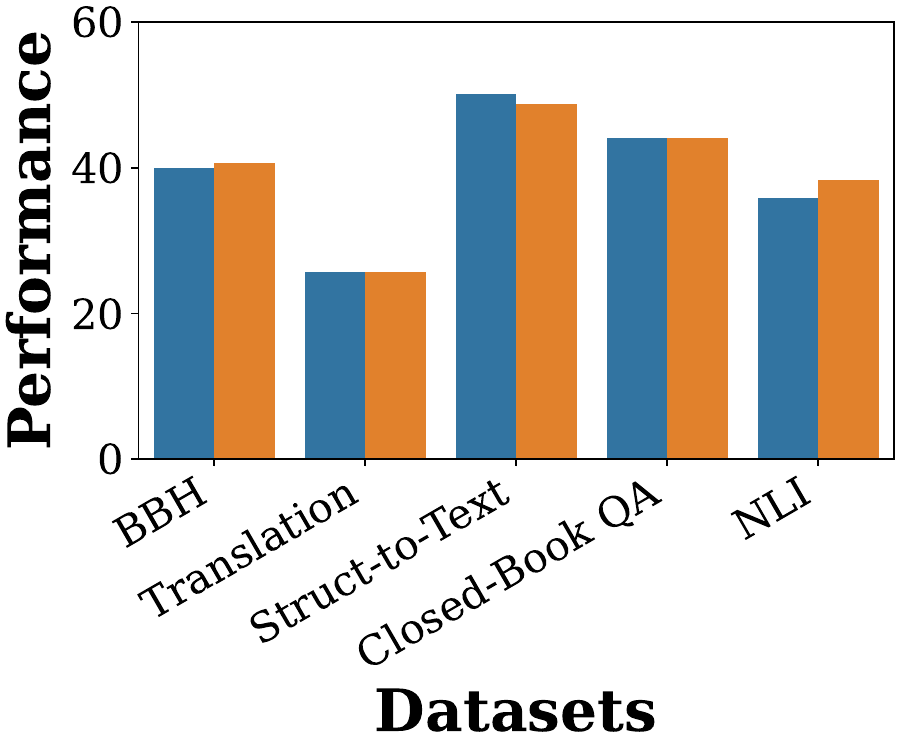}
}
\hspace{1mm}
\subfloat[Entropy]{
    \includegraphics[width=0.46\columnwidth]{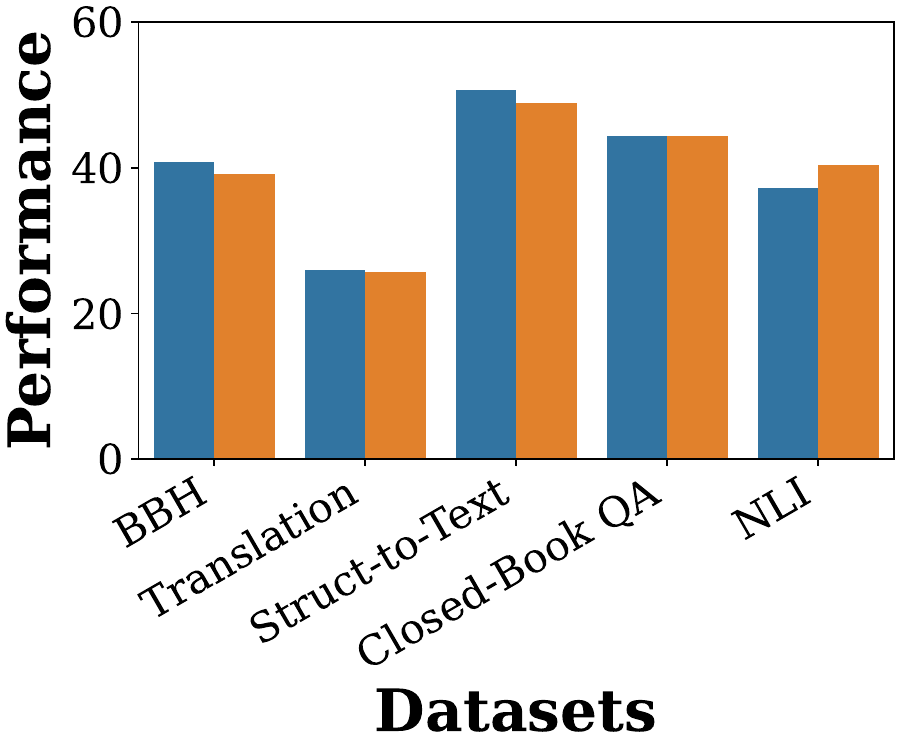}
}
\caption{Performance of \model{} using (a) norm and (b) entropy across datasets with different merging methods -- mixture and fusion.}
\label{fig:merging_method}
\end{figure}

\noindent
\paragraph{Analysis on Merging Method}
\model{} adopts the mixture as merging strategy, where the projection outputs of selected adapters are combined.
To assess this choice, we conduct an ablation study by replacing mixture with fusion, which merges the parameters of the selected adapters into a single set of weights before re-attaching them to the model.
Results are summarized in Fig.~\ref{fig:merging_method}, which compares the two strategies on the LLaMA-3.1-8B model across five task categories: BBH, Translation, Struct-to-Text, Closed-Book QA, and NLI.  The reported results are averaged over each dataset category.

Overall, we find that mixture and fusion achieve comparable performance across all categories.
However, fusion requires heavy parameter recomputation and re-attaching at the instance level, which introduces a substantial computational burden in deployment scenarios with large adapter pools.
In contrast, mixture achieves similar accuracy while incurring much lower computational overhead, making it a more practical choice.

\subsection{Robustness to Adapter Quality and Rank Variations}
\label{sec:robustness}

We study the robustness of \model{} to variations in adapter quality and rank.
Throughout the paper, we use adapters trained under a standardized instruction-tuning setup to enable controlled comparisons and isolate the behavior of probe-based signals.

\paragraph{Adapter Quality.}
\model{}’s probe-based signals are inherently agnostic to the cause of poor performance.
Adapters that are poorly trained or misaligned with the input tend to exhibit lower activation norms or higher-entropy responses, which naturally results in lower selection frequencies or reduced merging weights.
As a consequence, \model{} deprioritizes low-quality adapters without supervision or task-specific signals.

To validate this behavior, we construct a mixed adapter pool containing both well-trained adapters and intentionally under-trained checkpoints.
When all adapters are well trained, \model{} selects adapters from both groups at comparable rates (e.g., 57\% vs. 43\% on BBH tasks).
In contrast, when under-trained adapters are introduced, \model{} shifts its preference strongly toward well-trained adapters (e.g., 80\% vs. 20\%), demonstrating robustness to noisy or uneven-quality LoRAs.
While under-trained adapters are not completely suppressed, they are consistently deprioritized by the signals.

\paragraph{Adapter Rank.}
Higher-rank adapters may, in principle, produce larger projection norms due to increased representational capacity.
Since selection is performed comparatively at the instance level, such rank-based inflation would only matter if it consistently overrides task-alignment signals.
To empirically assess this, we construct a mixed adapter pool of 45 adapters using 15 FLAN-v2 training datasets, each trained at three different ranks ($r \in \{8, 16, 32\}$), and evaluate norm-based top-10 selection on BBH.
We find that approximately 80\% of selected adapters belong to training-dataset groups with multiple rank variants, indicating that selection is primarily driven by training-dataset alignment rather than adapter rank.

\subsection{Representational Interference Between LoRAs}
\label{sec:lora_interference}

A potential concern when merging multiple LoRA adapters is representational interference arising from conflicting parameter updates.
We argue that our setting is less susceptible to this issue for the following reasons.
First, LoRA adapters are low-rank updates applied on top of a shared frozen backbone, constraining perturbations to low-dimensional subspaces and thereby reducing the risk of large-scale representational conflict.
Second, our method performs instance-level, input-conditioned weighting rather than static parameter averaging: adapters with stronger relevance signals dominate the merged update for a given input, while unrelated adapters receive near-zero weights.
Third, we merge adapter activations during the forward pass rather than permanently combining parameters in weight space, avoiding commitment to a single merged parameter state.

To directly assess potential interference, we analyze the adapters that are actually merged, i.e., the selected top-20 set.
Specifically, we measure the proportion of negatively aligned adapter pairs (cosine similarity < 0) within the selected sets at the activation level.
We find that approximately 16\% of selected adapter pairs exhibit negative cosine similarity, indicating that the majority (84\%) are non-negatively aligned.
Together with the stable downstream performance observed across benchmarks, this suggests that destructive interference is not a dominant effect under the proposed method.

\subsection{Selected LoRAs}
\begin{figure*}[t]
\centering

\subfloat{
    \includegraphics[width=0.35\textwidth]{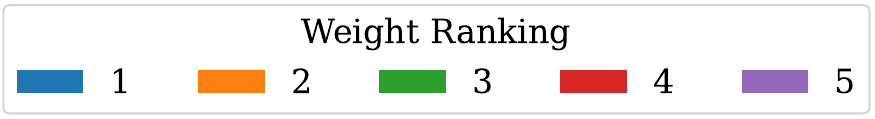}
}
\addtocounter{subfigure}{-1}
\vspace{2mm}

\subfloat[BBH Boolean Expressions]{
    \includegraphics[width=0.48\textwidth]{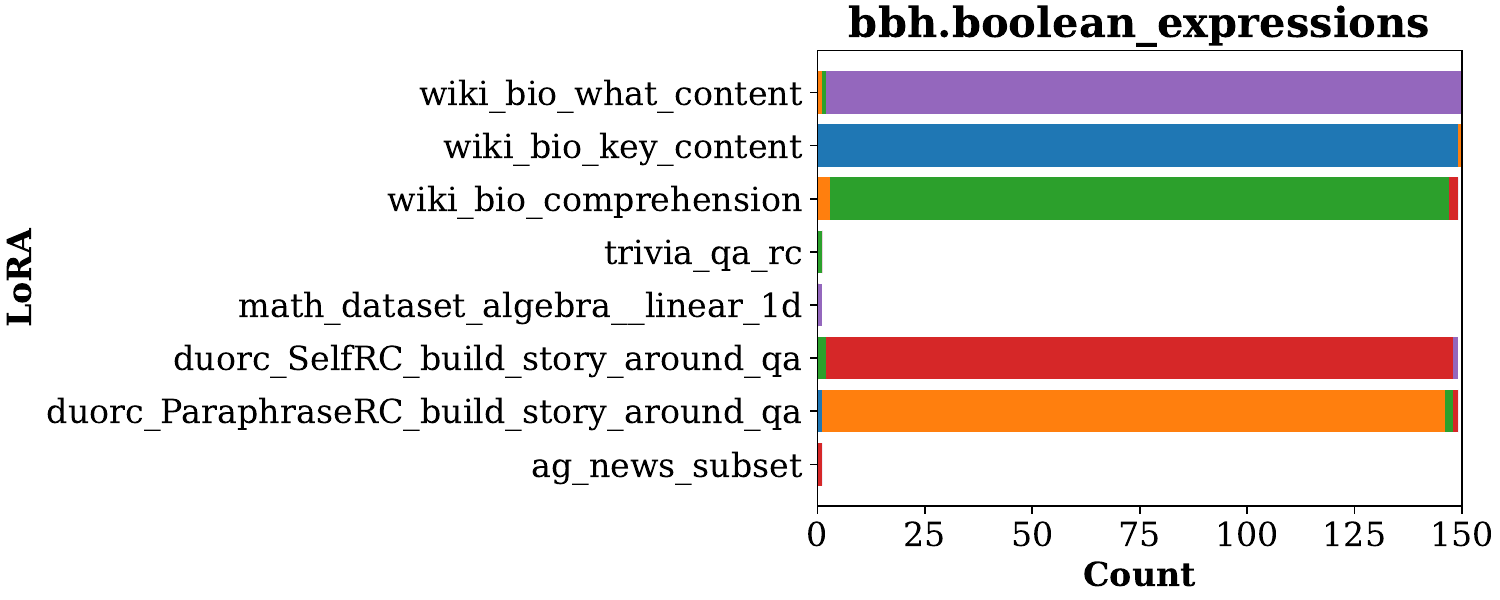}
}
\subfloat[BBH Causal Judgement]{
    \includegraphics[width=0.48\textwidth]{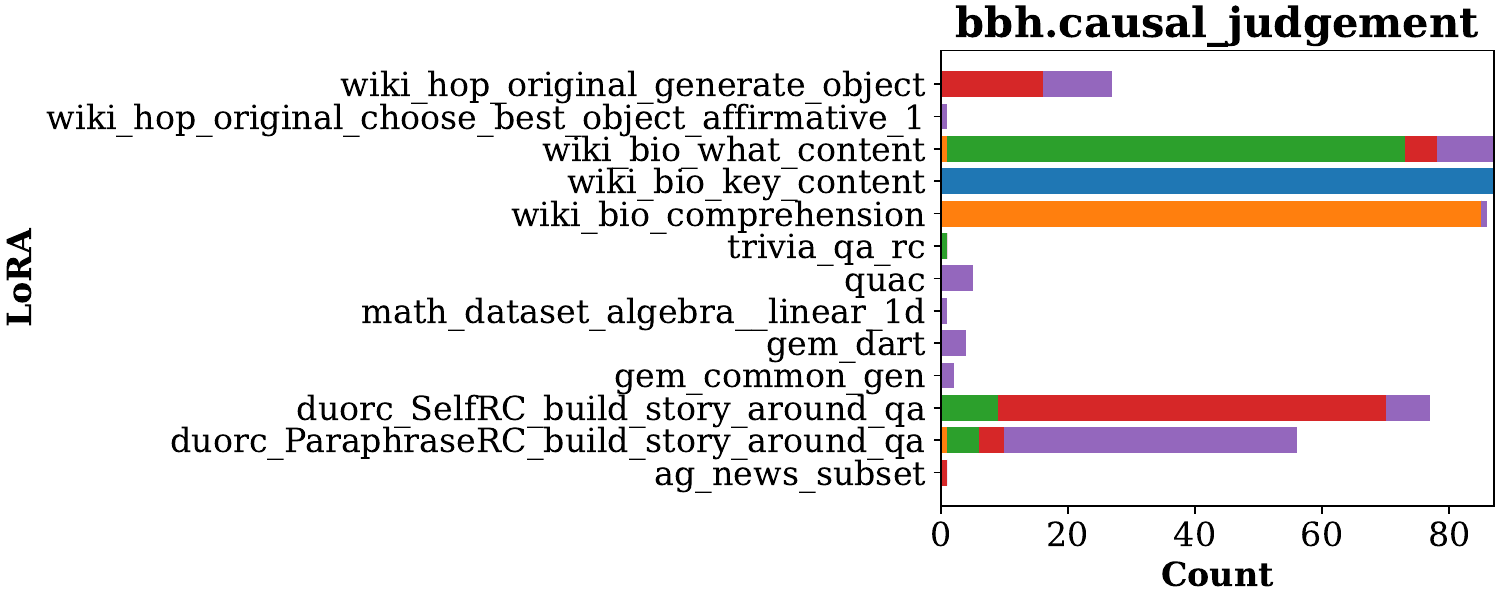}
}

\subfloat[BBH Formal Fallacies]{
    \includegraphics[width=0.48\textwidth]{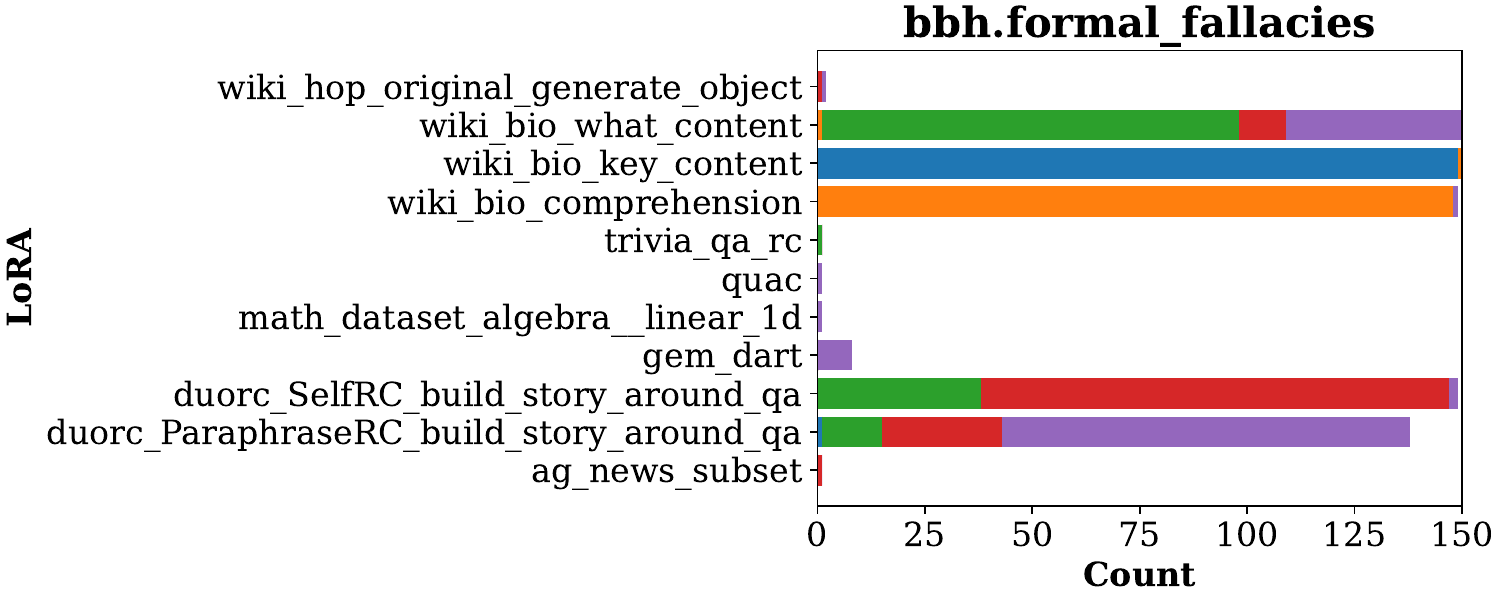}
}
\subfloat[BBH Navigate]{
    \includegraphics[width=0.48\textwidth]{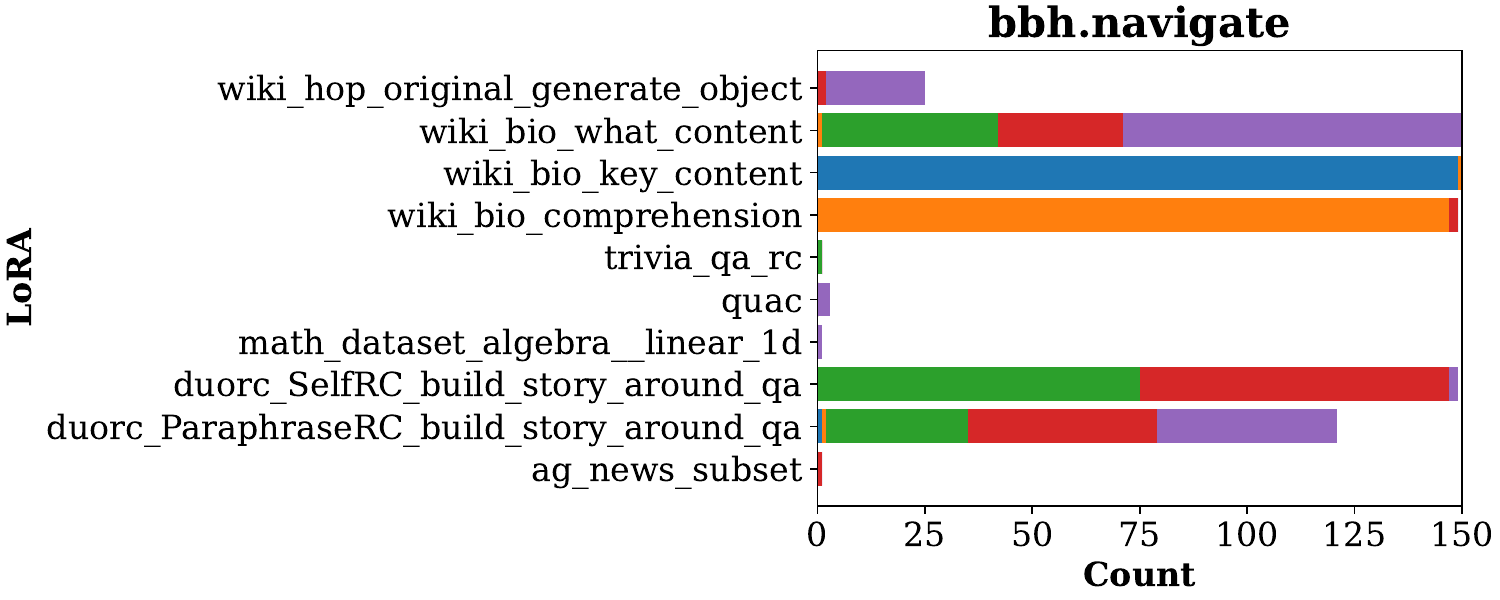}
}

\subfloat[BBH Object Counting]{
    \includegraphics[width=0.48\textwidth]{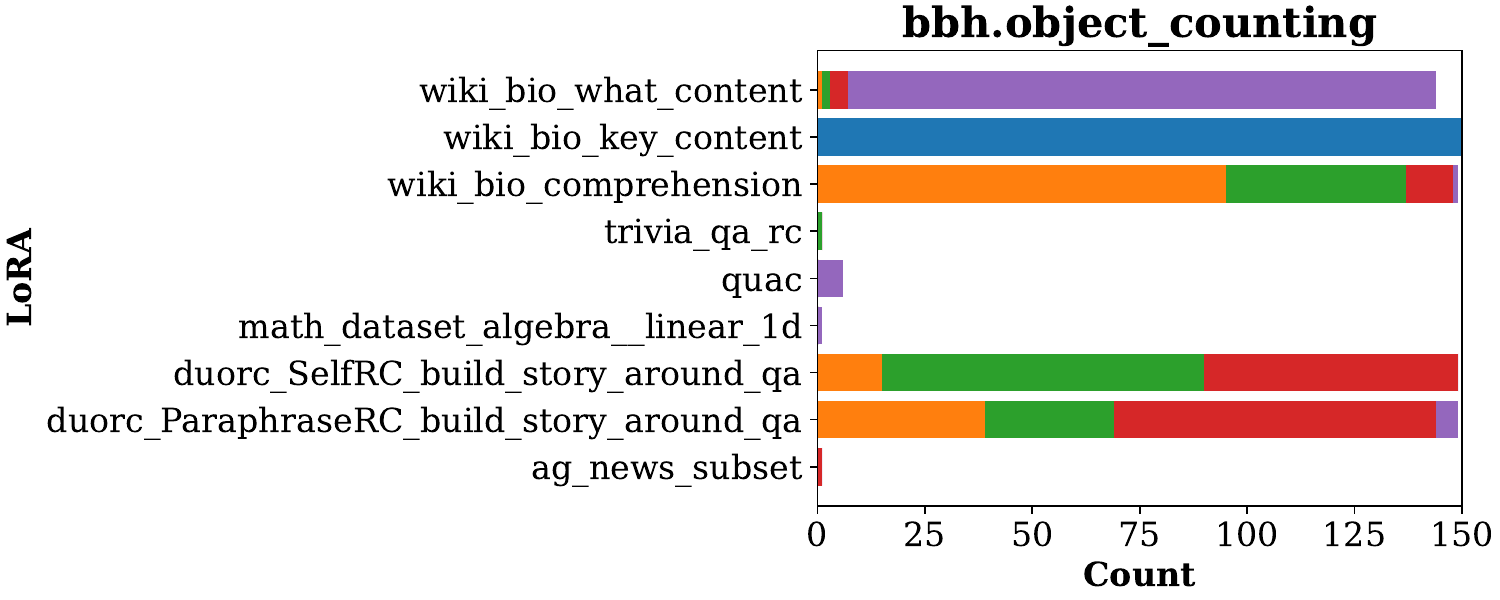}
}
\subfloat[BBH Sports Understanding]{
    \includegraphics[width=0.48\textwidth]{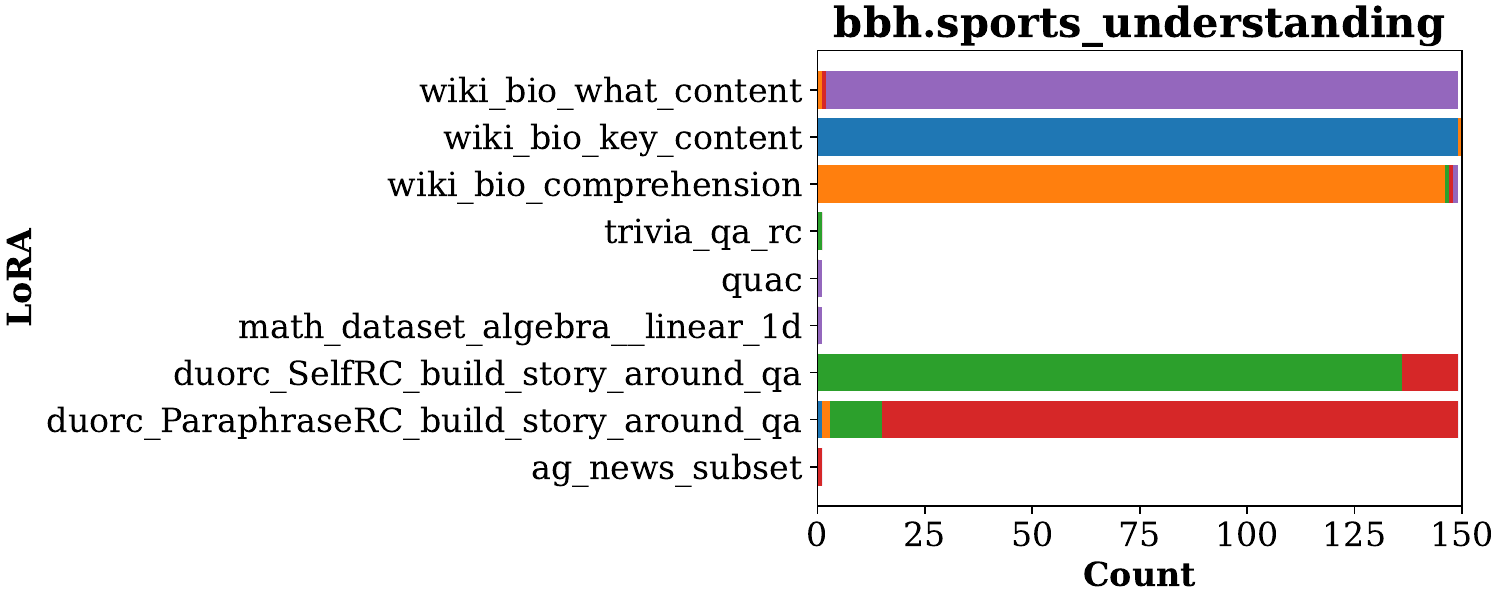}
}

\subfloat[BBH Web-of-lies]{
    \includegraphics[width=0.48\textwidth]{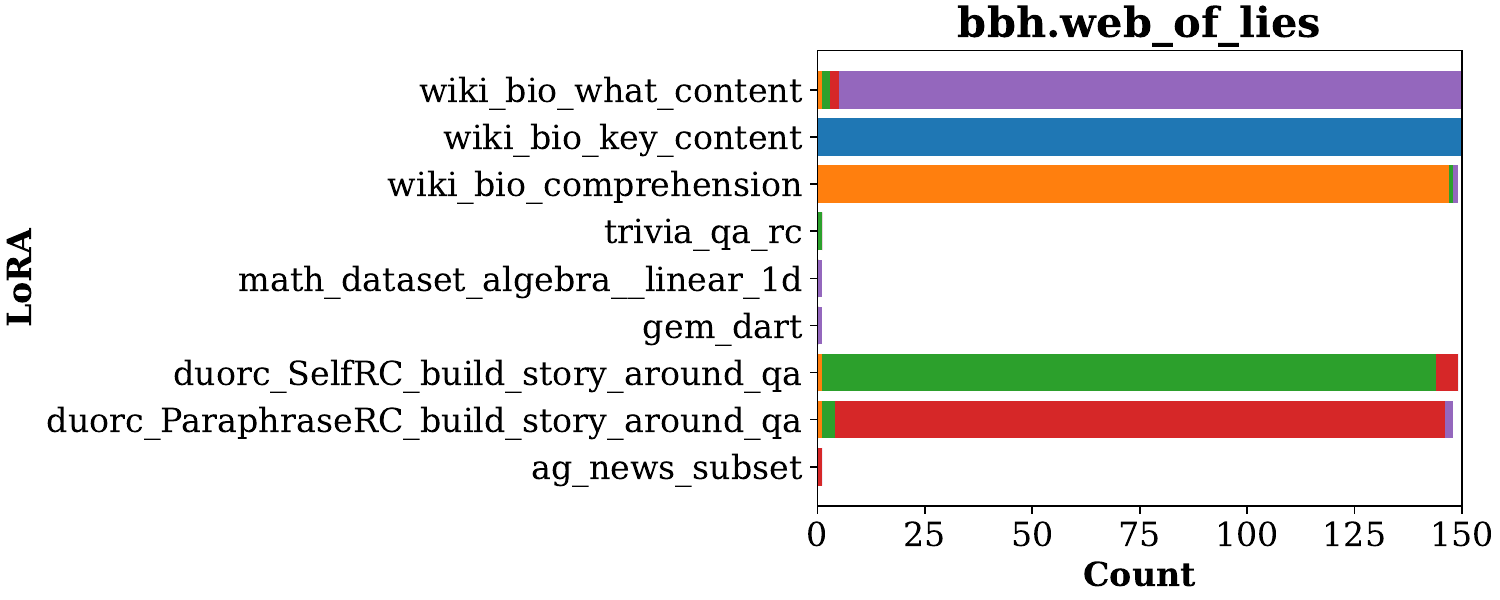}
}
\subfloat[BBH Word Sorting]{
    \includegraphics[width=0.48\textwidth]{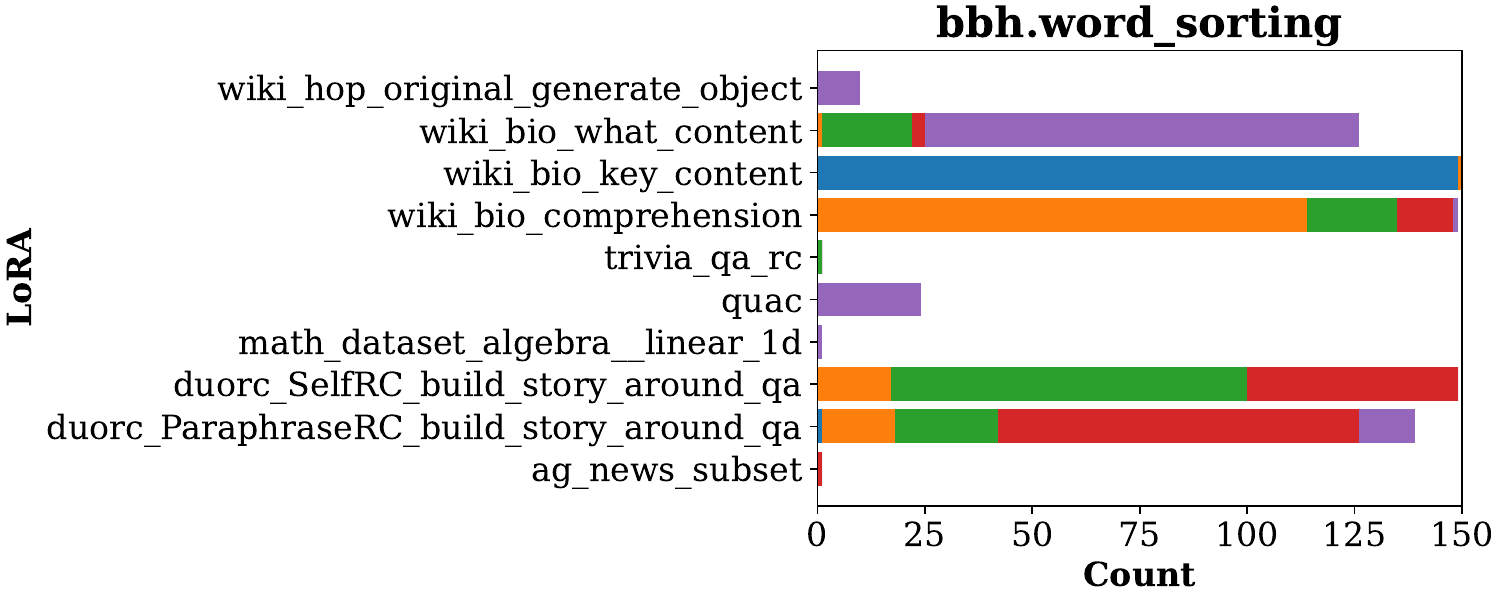}
}

\caption{The LoRA selection count by \model{} with LLaMA-3.1-8B model for BIG Bench Hard datasets. }
\label{fig:lora1}
\end{figure*}

\begin{figure*}[t]
\centering

\subfloat{
    \includegraphics[width=0.35\textwidth]{graphics/selected_loras_v2/lora_counts_horizontal_legend.pdf}
}
\addtocounter{subfigure}{-1}
\vspace{2mm}

\subfloat[WMT'14 FR->EN]{
    \includegraphics[width=0.48\textwidth]{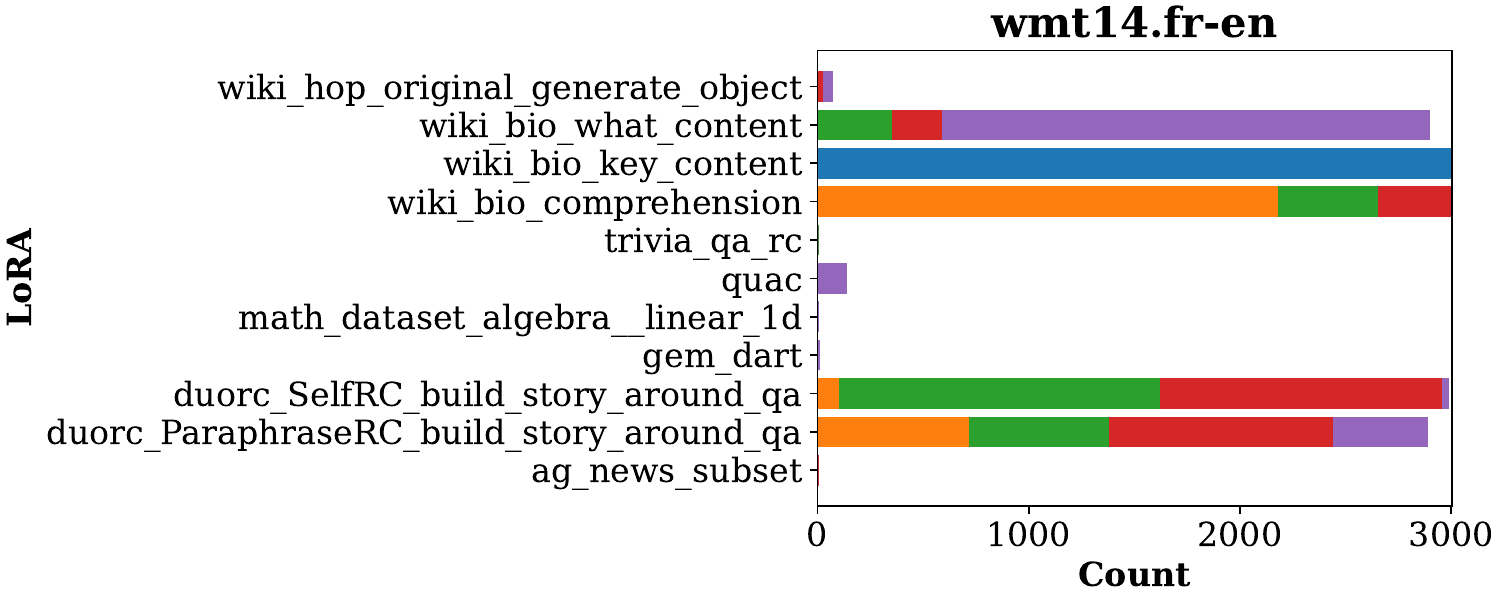}
}
\subfloat[WMT'14 EN->FR]{
    \includegraphics[width=0.48\textwidth]{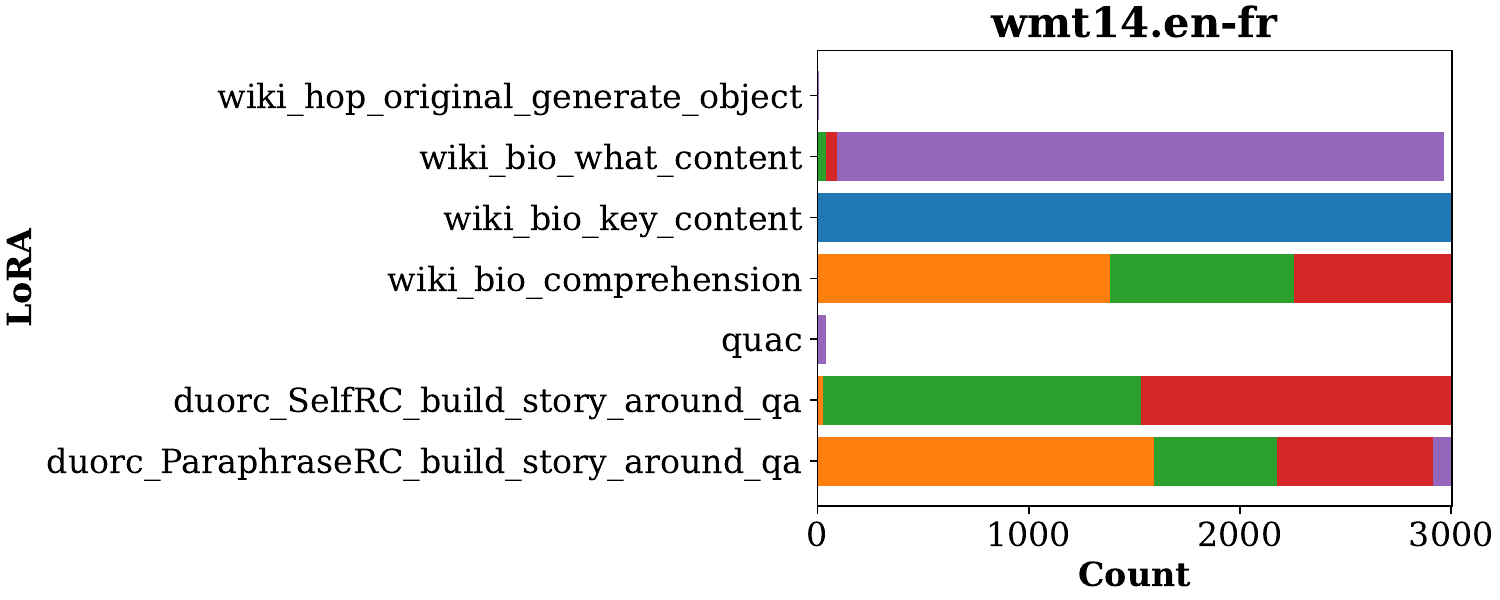}
}

\subfloat[WMT'16 DE->EN]{
    \includegraphics[width=0.48\textwidth]{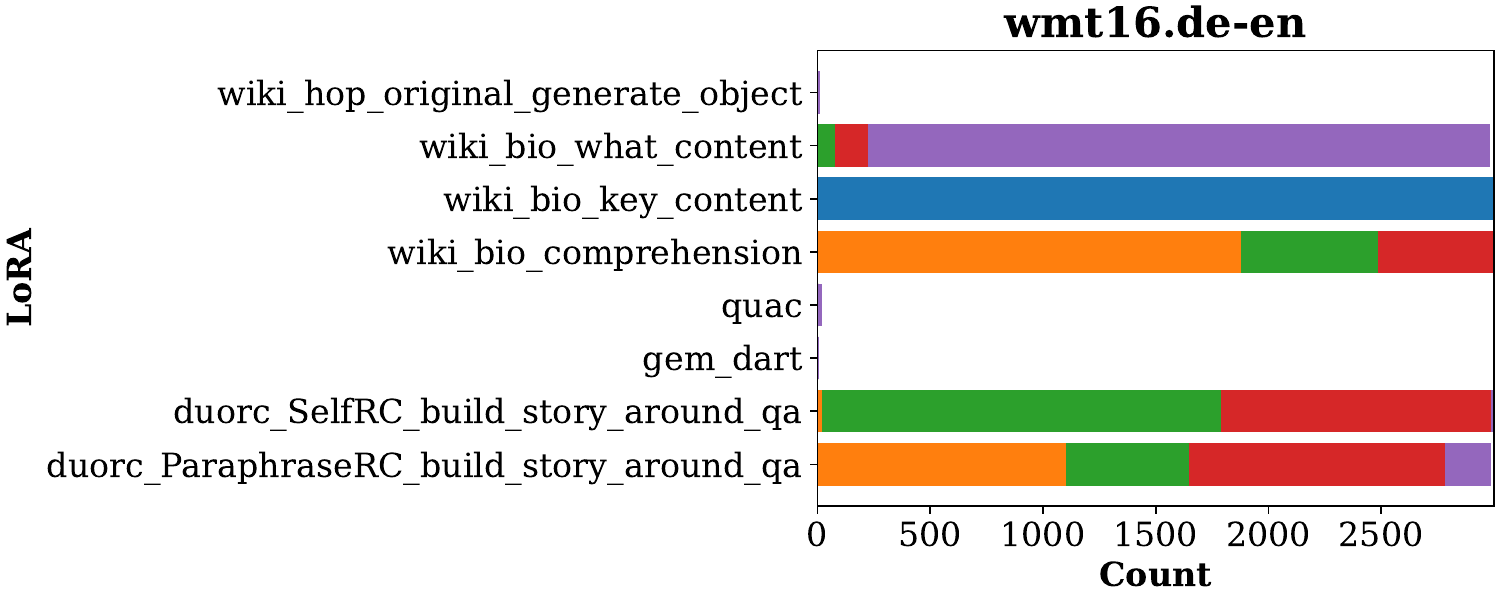}
}
\subfloat[WMT'16 EN->DE]{
    \includegraphics[width=0.48\textwidth]{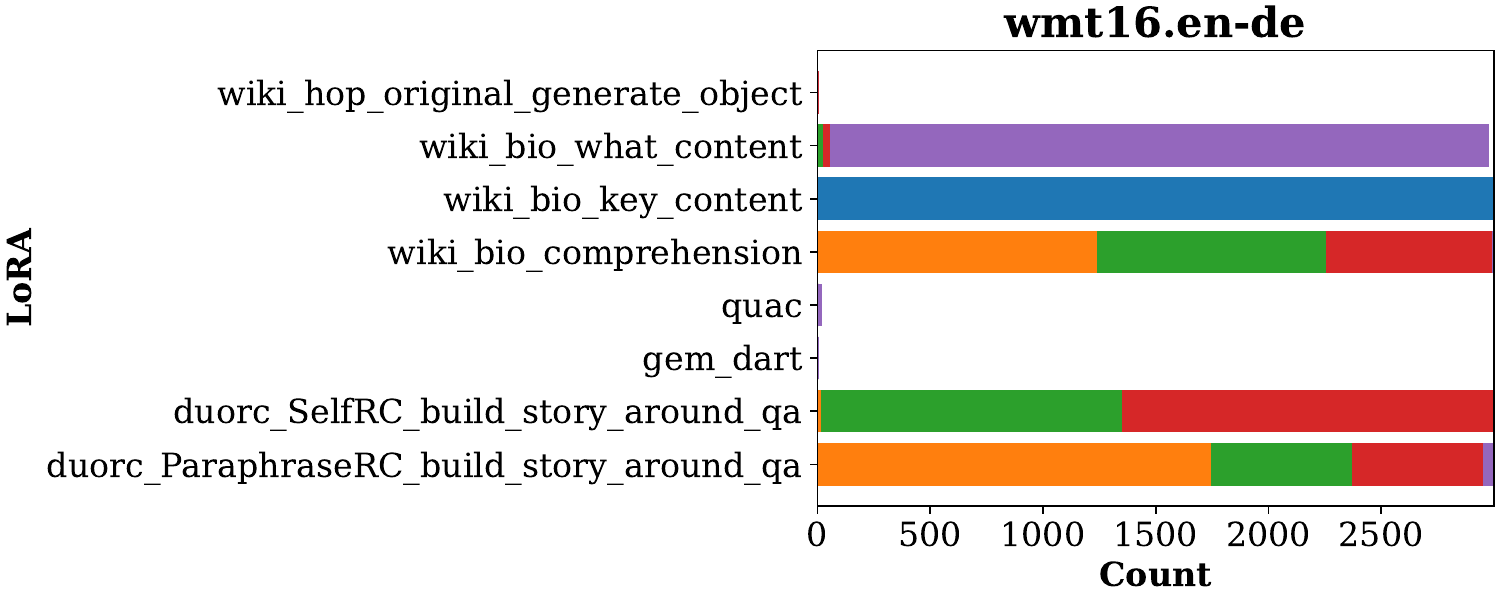}
}

\subfloat[WMT'16 RO->EN]{
    \includegraphics[width=0.48\textwidth]{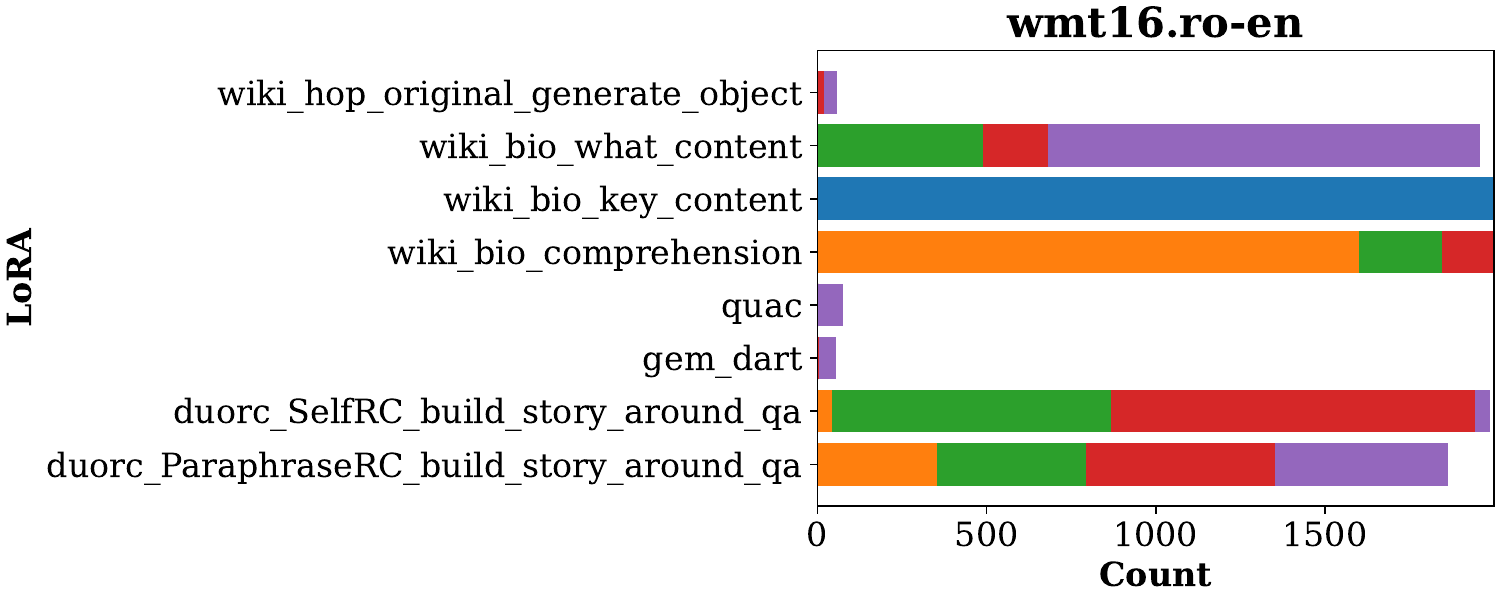}
}
\subfloat[WMT'16 EN->RO]{
    \includegraphics[width=0.48\textwidth]{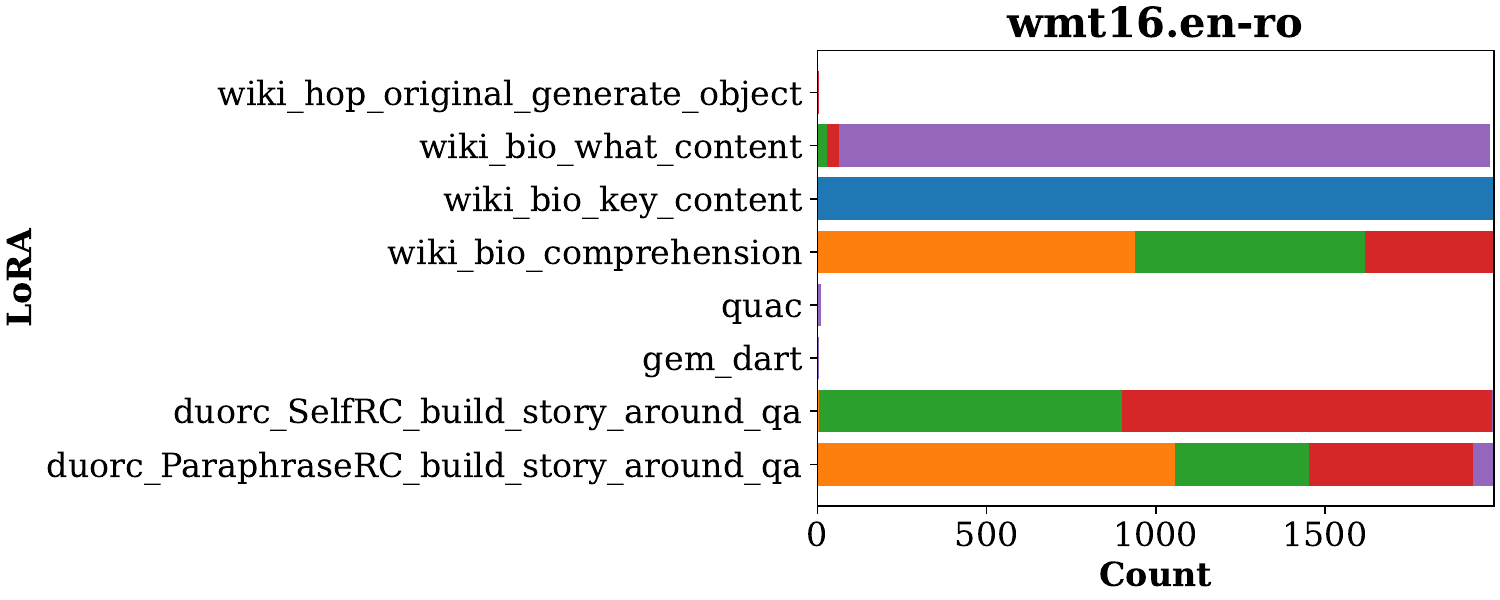}
}

\caption{The LoRA selection count by \model{} with LLaMA-3.1-8B model for translation datasets. }
\label{fig:lora2}
\end{figure*}

\begin{figure*}[t]
\centering

\subfloat{
    \includegraphics[width=0.35\textwidth]{graphics/selected_loras_v2/lora_counts_horizontal_legend.pdf}
}
\addtocounter{subfigure}{-1}
\vspace{2mm}

\subfloat[CommonGen]{
    \includegraphics[width=0.48\textwidth]{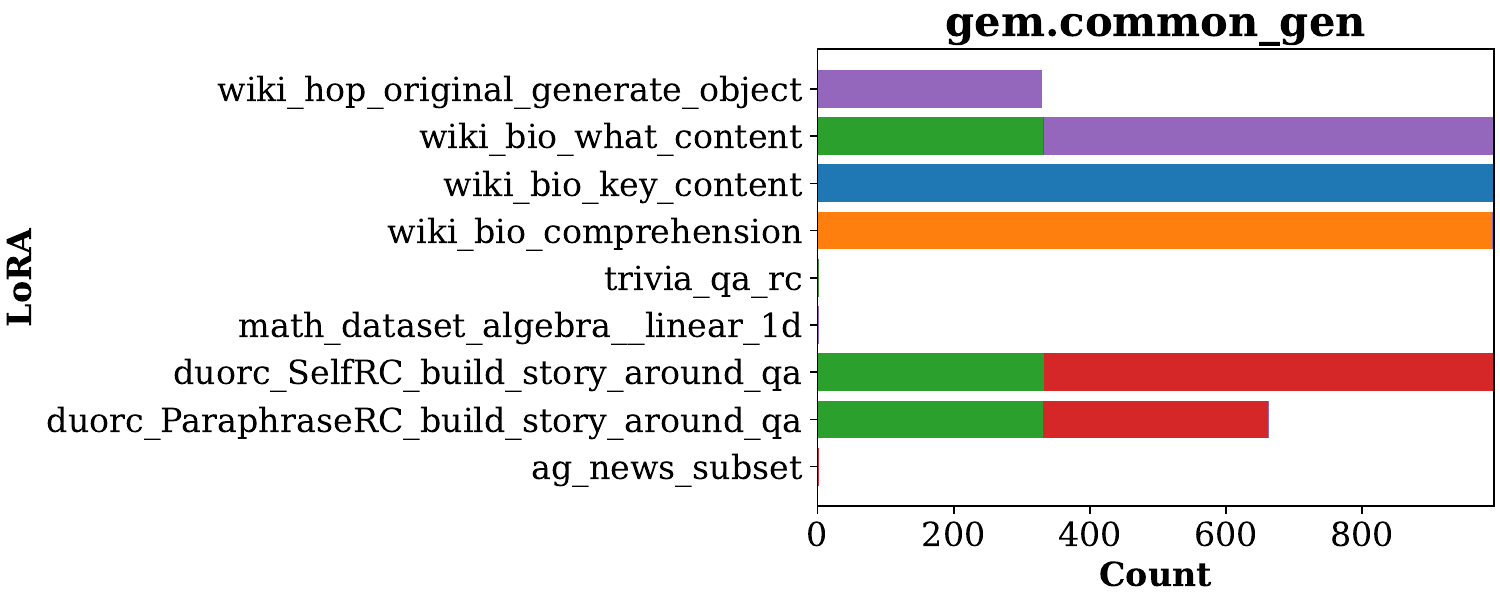}
}
\subfloat[DART]{
    \includegraphics[width=0.48\textwidth]{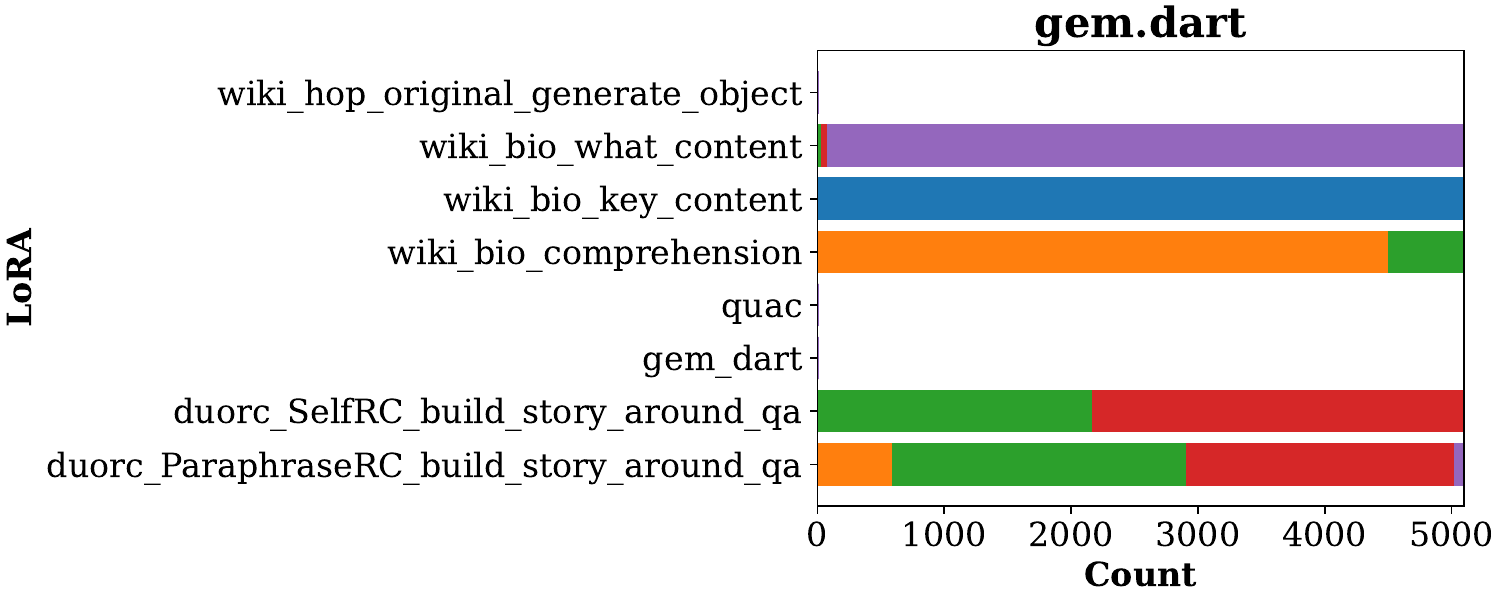}
}

\subfloat[E2ENLG]{
    \includegraphics[width=0.48\textwidth]{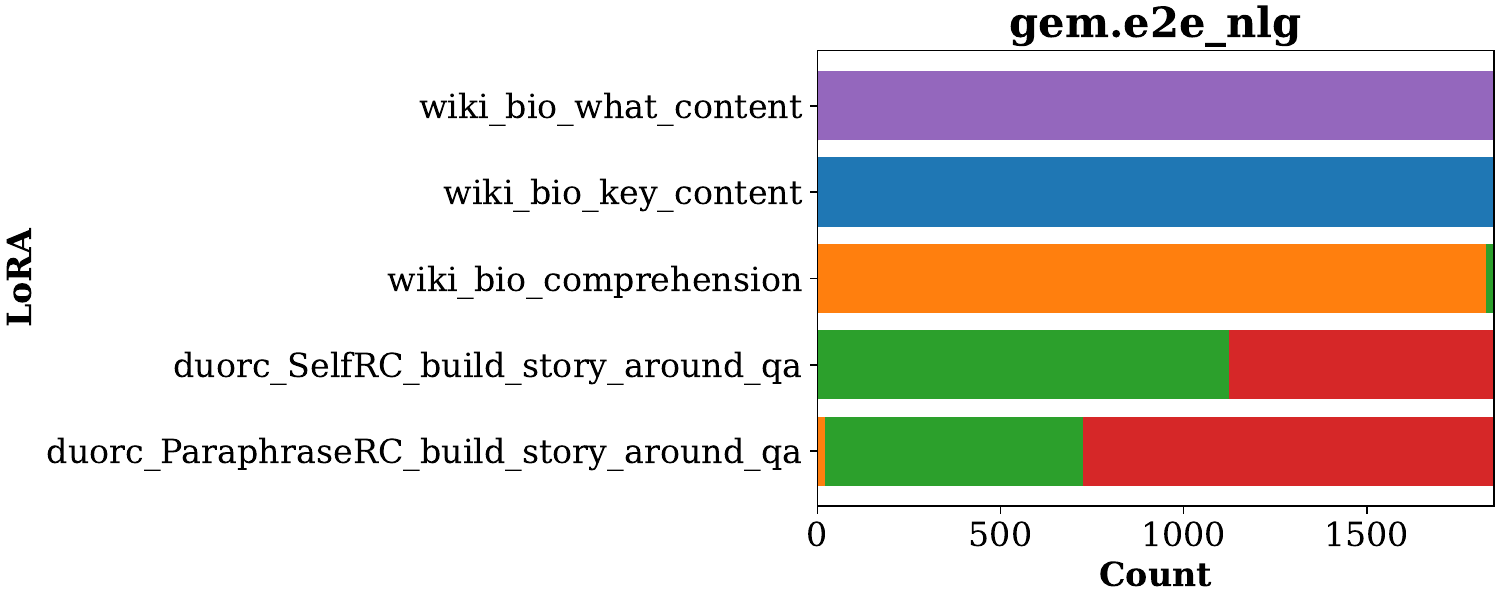}
}
\subfloat[WebNLG]{
    \includegraphics[width=0.48\textwidth]{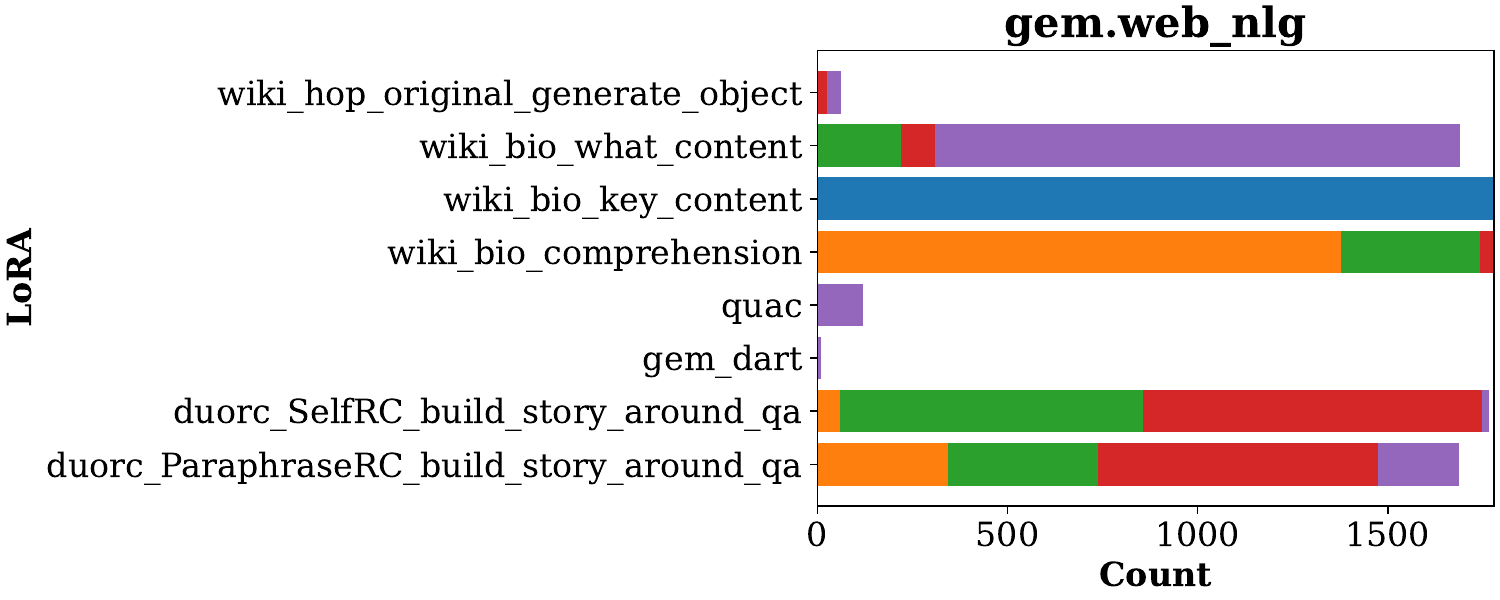}
}

\caption{The LoRA selection count by \model{} with LLaMA-3.1-8B model for struct-to-text datasets. }
\label{fig:lora3}
\end{figure*}

\begin{figure*}[t]
\centering

\subfloat{
    \includegraphics[width=0.35\textwidth]{graphics/selected_loras_v2/lora_counts_horizontal_legend.pdf}
}
\addtocounter{subfigure}{-1}
\vspace{2mm}

\subfloat[ARC-c]{
    \includegraphics[width=0.48\textwidth]{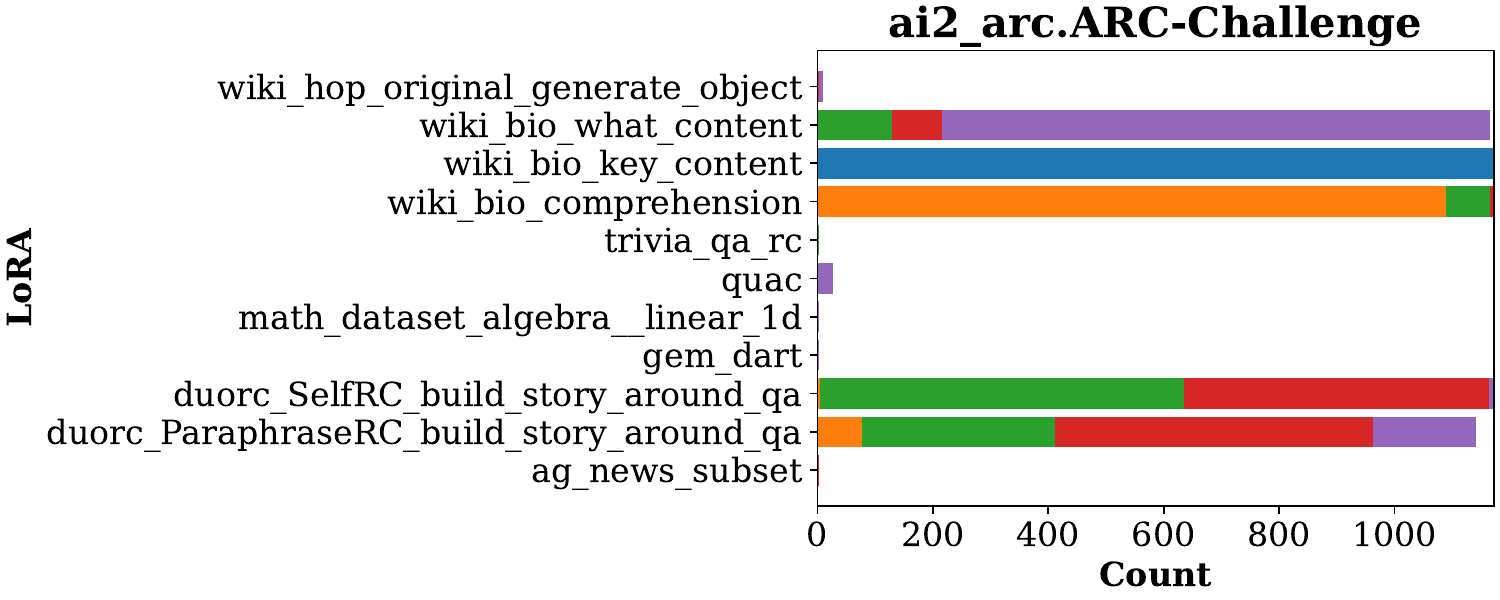}
}
\subfloat[ARC-e]{
    \includegraphics[width=0.48\textwidth]{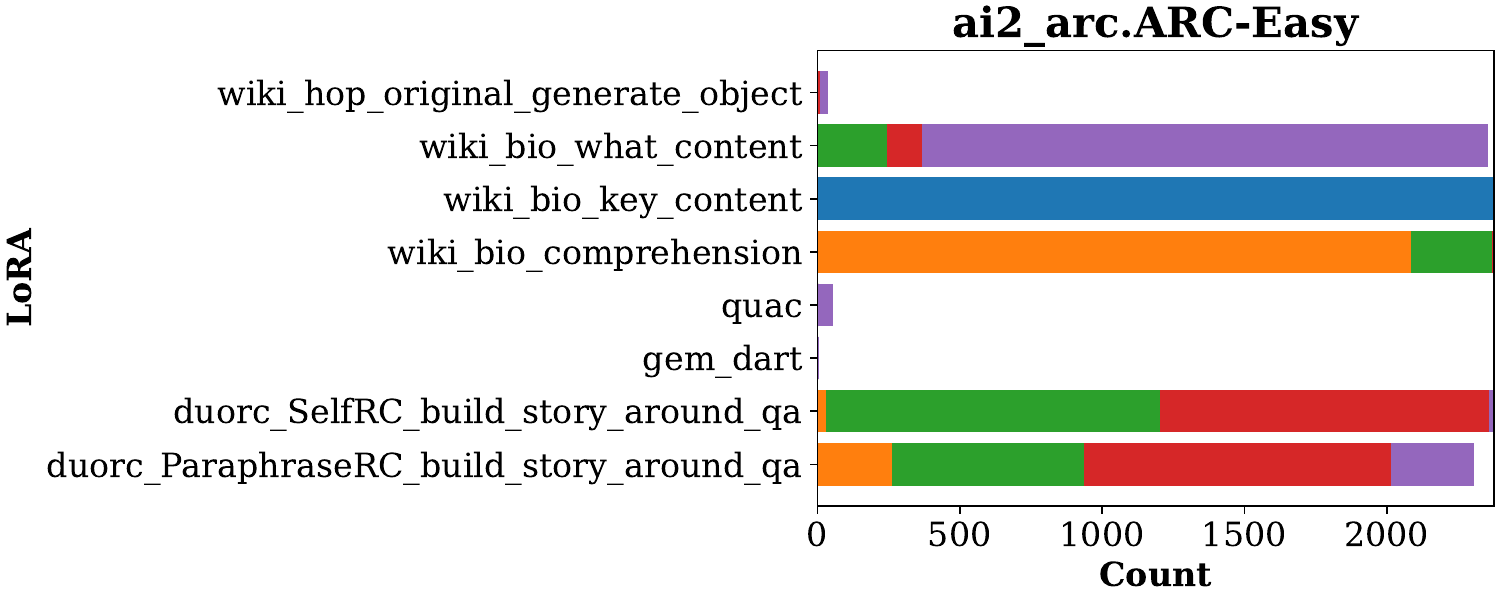}
}

\subfloat[Natural Questions]{
    \includegraphics[width=0.48\textwidth]{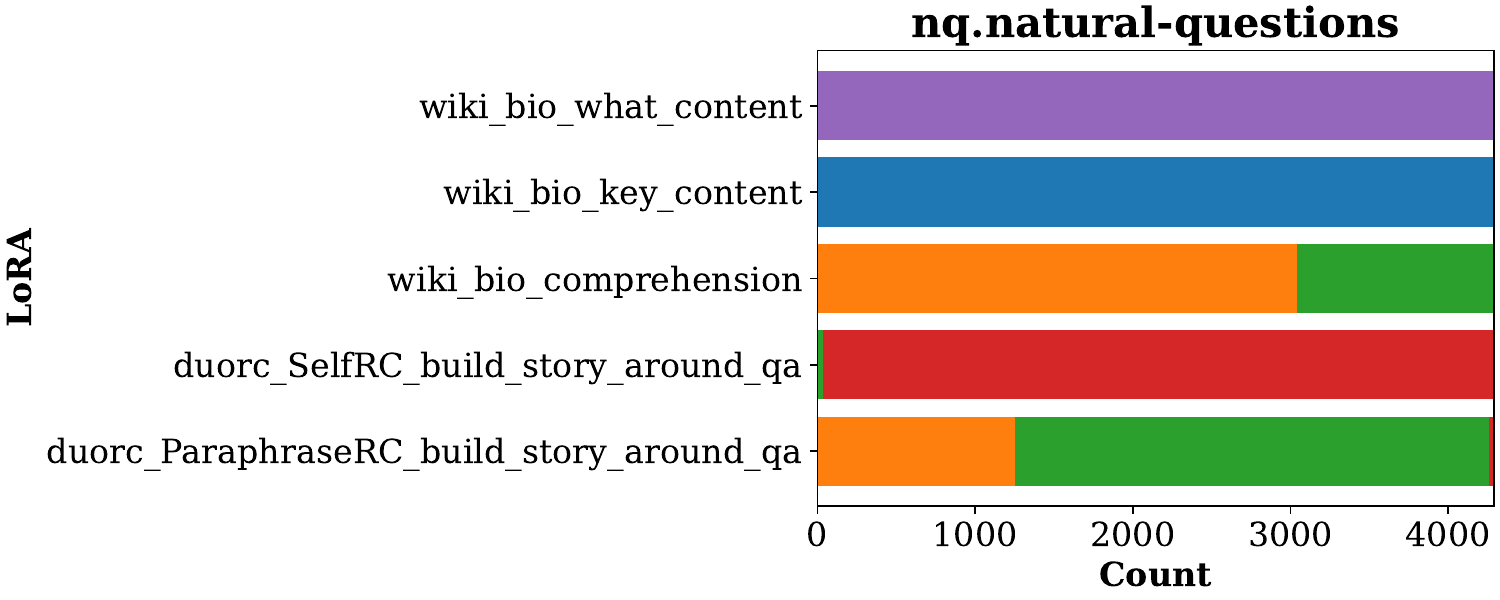}
}
\subfloat[Trivia QA]{
    \includegraphics[width=0.48\textwidth]{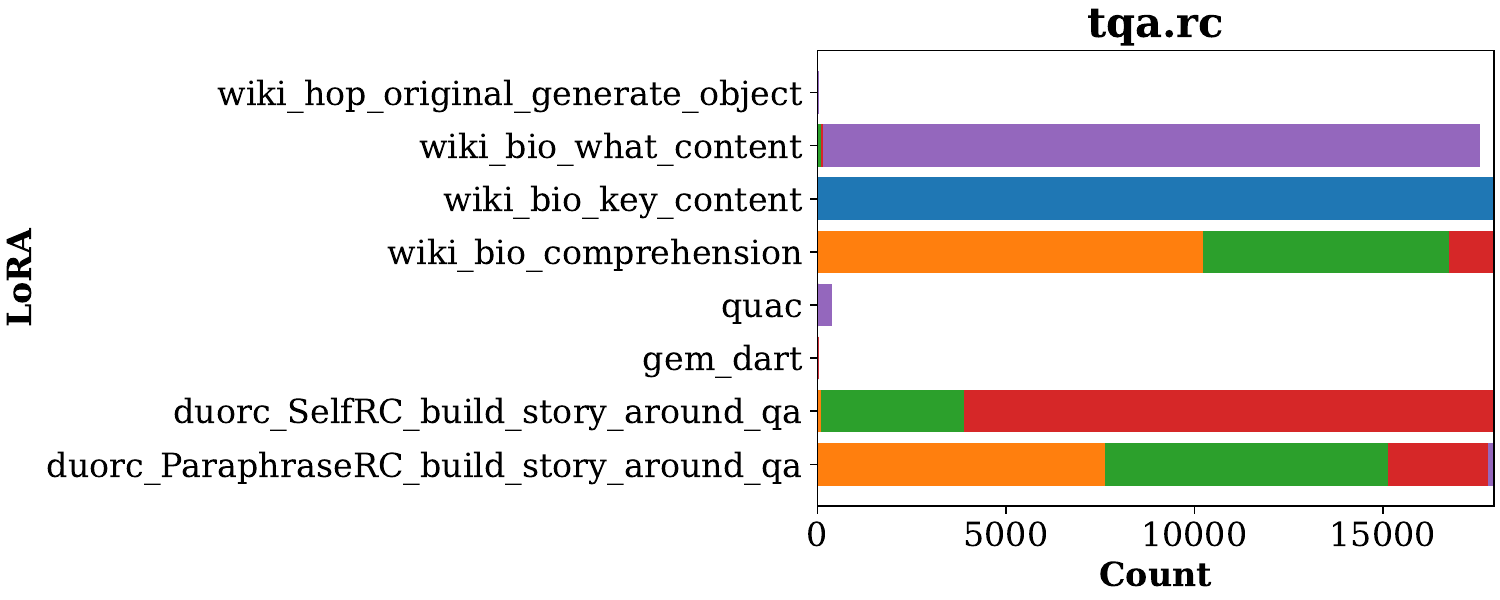}
}

\caption{The LoRA selection counts by \model{} with LLaMA-3.1-8B model for closed-book QA datasets. }
\label{fig:lora4}
\end{figure*}

\begin{figure*}[t]
\centering

\subfloat{
    \includegraphics[width=0.35\textwidth]{graphics/selected_loras_v2/lora_counts_horizontal_legend.pdf}
}
\addtocounter{subfigure}{-1}
\vspace{2mm}

\subfloat[ANLI-R1]{
    \includegraphics[width=0.48\textwidth]{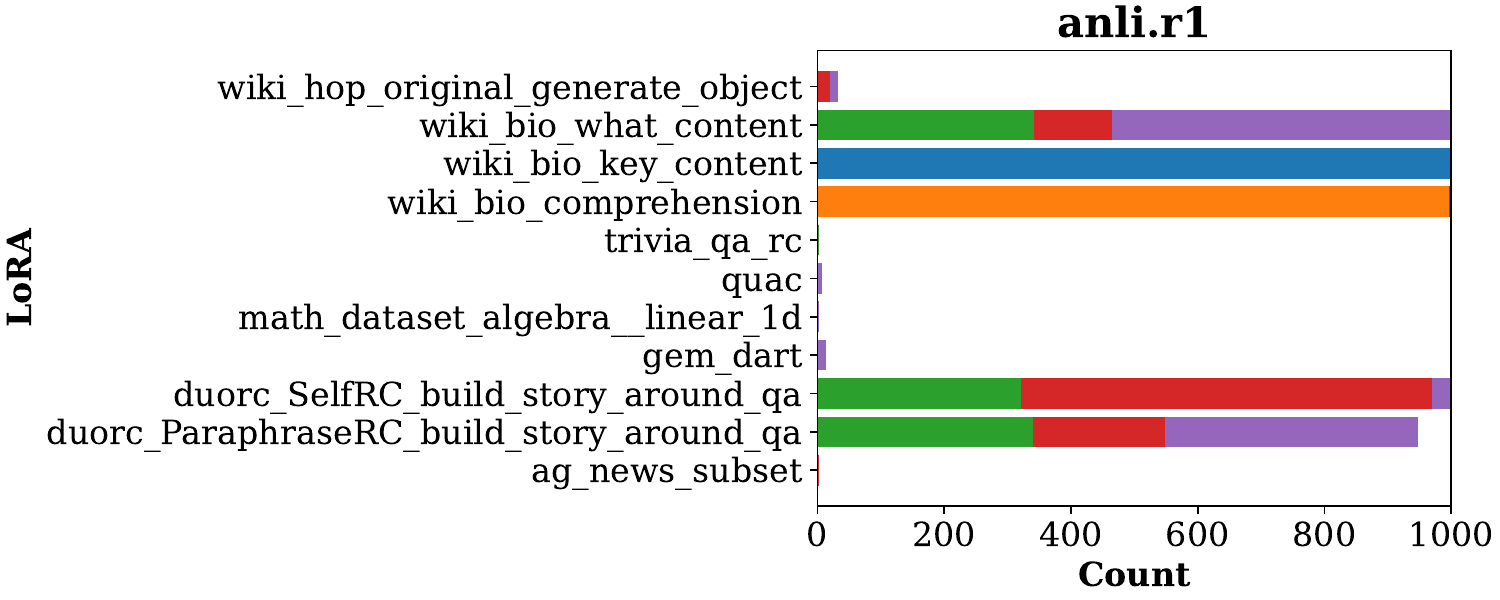}
}
\subfloat[ANLI-R2]{
    \includegraphics[width=0.48\textwidth]{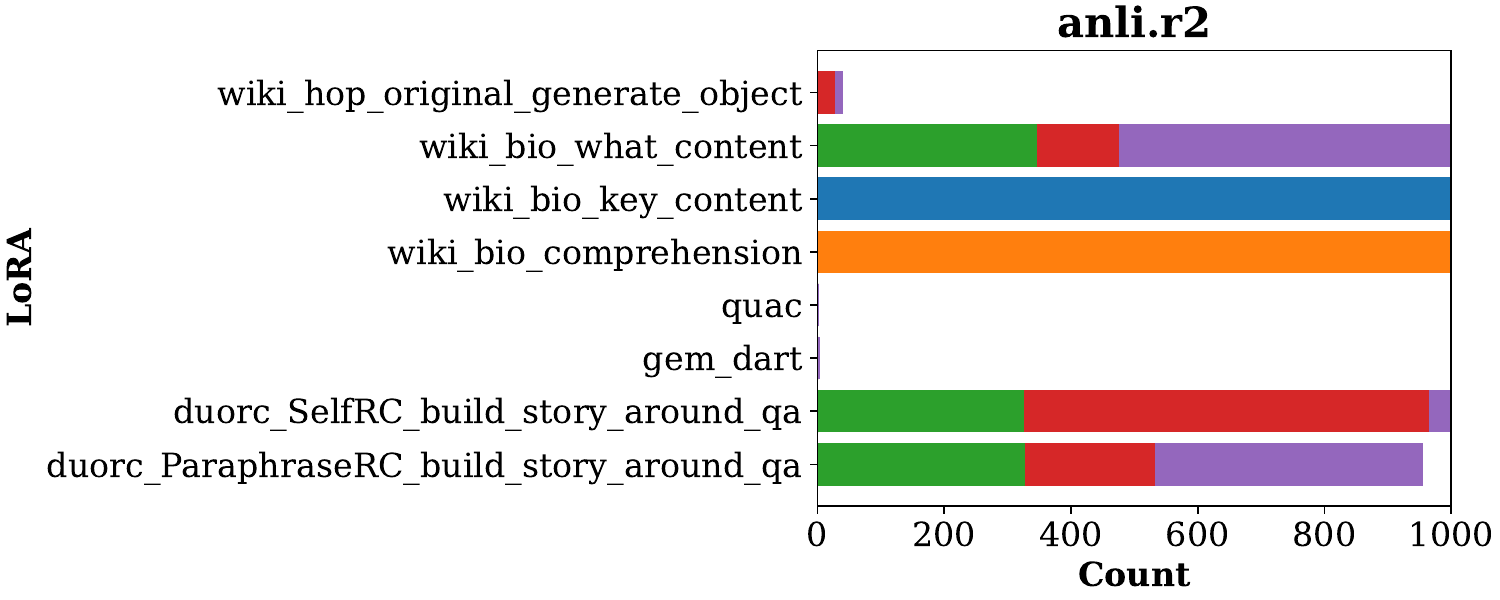}
}

\subfloat[ANLI-R3]{
    \includegraphics[width=0.48\textwidth]{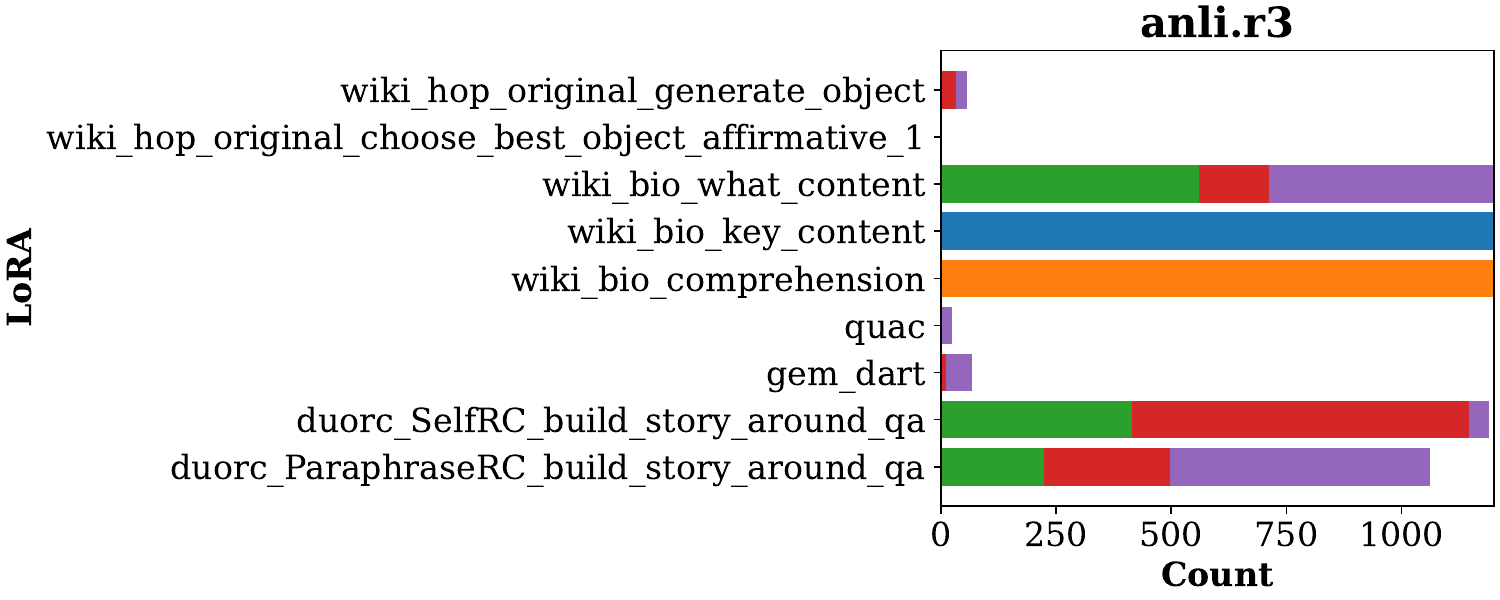}
}
\subfloat[QNLI]{
    \includegraphics[width=0.48\textwidth]{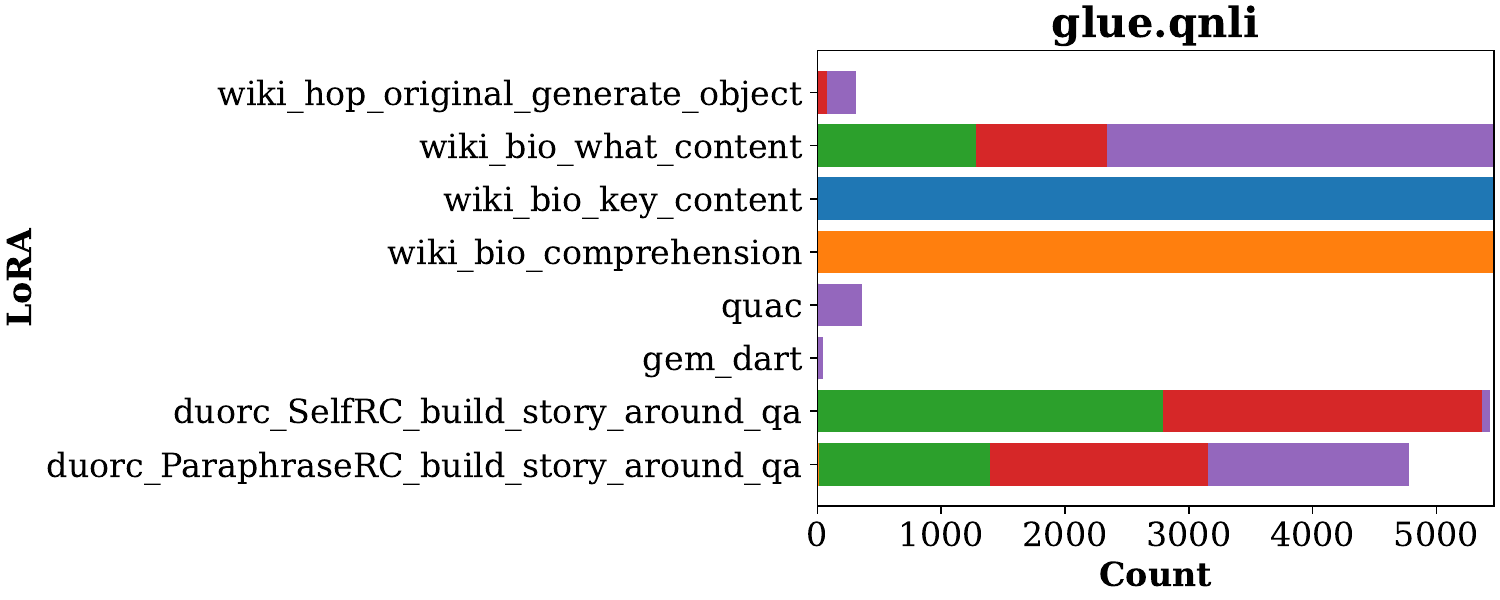}
}

\caption{The LoRA selection counts by \model{} with LLaMA-3.1-8B model for NLI datasets. }
\label{fig:lora5}
\end{figure*}

Figures~\ref{fig:lora1} to~\ref{fig:lora5} show the LoRA adapter selection counts by our proposed method, \model{}, with LLaMA-3.1-8B model for various datasets.

\end{document}